%% file: bgplvm.tex
\documentclass[twoside,11pt]{article}

\usepackage{jmlr2e}
\usepackage{times}

\usepackage[margin=3cm]{geometry}
\usepackage[small]{caption}
\usepackage[ansinew]{inputenc}
\usepackage{amsmath}
\usepackage{bm}
\usepackage{bbm}
\usepackage{verbatim}
\usepackage{nccmath}
\usepackage{subfigure}
\usepackage{cancel}
\usepackage{verbatim}
\usepackage{color}
\usepackage{algorithm}
\usepackage[noend]{algorithmic}
\usepackage{eqparbox} 
\newlength\myindent
\newcommand\SSTATE{\STATE}

\usepackage{todonotes}
\usepackage{todonotes}
\usepackage[hyphens]{url}

\expandafter\def\expandafter\UrlBreaks\expandafter{\UrlBigBreaks
  \do\a\do\b\do\c\do\d\do\e\do\f\do\g\do\h\do\i\do\j%
  \do\k\do\l\do\m\do\n\do\o\do\p\do\q\do\r\do\s\do\t%
  \do\u\do\v\do\w\do\x\do\y\do\z\do\A\do\B\do\C\do\D%
  \do\E\do\F\do\G\do\H\do\I\do\J\do\K\do\L\do\M\do\N%
  \do\O\do\P\do\Q\do\R\do\S\do\T\do\U\do\V\do\W\do\X%
  \do\Y\do\Z\do\*\do\-\do\~\do\'\do\"\do\-}

\usepackage{mathtools}
\mathtoolsset{showonlyrefs}

\input{notationDef.tex}

\input{definitions.tex}

\input{graphicalModels.tex}

\newcommand{\observedSet}{o}
\newcommand{\unobservedSet}{u}
\newcommand{\dataIndexOne}{i}

\newcommand{\footremember}[2]{%
   \thanks{#2}
    \newcounter{#1}
    \setcounter{#1}{\value{footnote}}%
}
\newcommand{\footrecall}[1]{%
    \footnotemark[\value{#1}]%
}

\ShortHeadings{Variational GP-LVM}{Damianou, Titsias and Lawrence}
\firstpageno{1}

\begin{document}

\title{Variational Inference for Uncertainty on the Inputs of Gaussian Process Models}


\author{\name Andreas C. Damianou\footremember{equal}{These authors contributed equally to this work.} \email andreas.damianou@sheffield.ac.uk \\
       \addr Dept. of Computer Science and Sheffield Institute for Translational Neuroscience\\
       University of Sheffield\\
       UK
  \AND
\name Michalis K. Titsias\footrecall{equal} \email mtitsias@aueb.gr \\
       \addr Department of Informatics\\
       Athens University of Economics and Business\\
       Greece
  \AND       
\name Neil D. Lawrence \email N.Lawrence@dcs.sheffield.ac.uk \\
       \addr Dept. of Computer Science and Sheffield Institute for Translational Neuroscience\\
       University of Sheffield\\
       UK
}

\editor{}

\maketitle

\newcommand{\highlight}[1]{\colorbox{yellow}{#1}}

\newcommand{\bfzi}{\mathbf{z}}
\newcommand{\bfF}{\mathbf{F}}

\newcommand{\ie}{i.e.\ }
\newcommand{\eg}{e.g.\ }

\renewcommand{\la}{\left\langle}
\newcommand{\ra}{\right\rangle}
 \newcommand{\vv}{\vartheta}

\newcommand{\intd}{\text{d}}
\newcommand{\F}{\mathcal{F}}

\renewcommand{\Kff}{\kernelMatrix_{\mappingFunction\mappingFunction}}
\renewcommand{\Kfu}{\kernelMatrix_{\mappingFunction\inducingScalar}}
\renewcommand{\Kuf}{\kernelMatrix_{\inducingScalar\mappingFunction}}
\renewcommand{\Kuu}{\kernelMatrix_{\inducingScalar\inducingScalar}}
\newcommand{\Kstaru}{\kernelMatrix_{* \inducingScalar}}
\newcommand{\Kustar}{\kernelMatrix_{\inducingScalar *}}
\newcommand{\Kfstar}{\kernelMatrix_{\mappingFunction *}}
\newcommand{\Kstarf}{\kernelMatrix_{* \mappingFunction}}
\newcommand{\Kstarstar}{\kernelMatrix_{* *}}
\newcommand{\Kx}{\kernelMatrix_\latentScalar}



\input{abstract.tex}
\input{introduction.tex}
\input{background.tex}
\input{vargplvm.tex}
\input{predictions.tex}
\input{matlab.tex} 
\input{experiments.tex}

\input{extensions.tex}
\input{conclusions.tex}

\acks{This research was partially funded by the European research project EU FP7-ICT (Project
Ref 612139 ``WYSIWYD''), the Greek State
 Scholarships Foundation (IKY) and the University of Sheffield Moody endowment fund.
We also thank Colin Litster and ``Fit Fur Life'' for allowing us to use their video files as datasets.
}


\newpage

\input{supplementary.tex}

\clearpage 

\vskip 0.2in

\bibliography{bgplvm,other,lawrence,zbooks}

\end{document}

%% file: notationDef.tex
\global\long\def\outputIndex{j}
\global\long\def\dataIndex{i}

\global\long\def\latentIndex{j}

\global\long\def\dataDim{p}
\global\long\def\latentDim{q}

\global\long\def\numData{n}

\global\long\def\numSequences{s}
\global\long\def\numInducing{m}

\global\long\def\rbfWidth{\ell}
\global\long\def\lengthScale{\ell}

\global\long\def\dataStd{\sigma}

\global\long\def\dataScalar{y}
\global\long\def\dataVector{\mathbf{\dataScalar}}
\global\long\def\dataMatrix{\mathbf{\MakeUppercase{\dataScalar}}}

\global\long\def\latentScalar{x}
\global\long\def\latentMatrix{\mathbf{\MakeUppercase{\latentScalar}}}
\global\long\def\latentVector{\mathbf{\latentScalar}}

\global\long\def\noiseVector{\boldsymbol{\epsilon}}
\global\long\def\noiseScalar{\epsilon}

\global\long\def\kernelScalar{k}

\global\long\def\kernelMatrix{\mathbf{\MakeUppercase{\kernelScalar}}}

\global\long\def\lengthScale{\ell}

\global\long\def\zerosVector{{\bf 0}}

\global\long\def\numData{n}

\global\long\def\dataDim{p}

\global\long\def\inputScalar{x}

\global\long\def\inputVector{{\bf \inputScalar}}

\global\long\def\mappingMatrix{\mathbf{W}}

\global\long\def\mappingFunction{f}
\global\long\def\mappingFunctionVector{\mathbf{\mappingFunction}}
\global\long\def\mappingFunctionMatrix{\mathbf{\MakeUppercase{\mappingFunction}}}
\global\long\def\mappingFunctionTwo{g}
\global\long\def\mappingFunctionTwoVector{\mathbf{\mappingFunctionTwo}}

\global\long\def\inducingScalar{u}
\global\long\def\inducingVector{\mathbf{\inducingScalar}}
\global\long\def\inducingMatrix{\mathbf{\MakeUppercase{\inducingScalar}}}

\global\long\def\eye{\mathbf{I}}

\global\long\def\onesVector{\mathbf{1}}
\global\long\def\zerosVector{\mathbf{0}}

\global\long\def\gaussianSamp#1#2{\mathcal{N}\left(#1,#2\right)}
\global\long\def\gaussianDist#1#2#3{\mathcal{N}\left(#1|#2,#3\right)}

\global\long\def\KL#1#2{\text{KL}\left( #1\,\|\,#2 \right)}

\global\long\def\Kff{\kernelMatrix_{\mappingFunctionVector \mappingFunctionVector}}
\global\long\def\Kuu{\kernelMatrix_{\inducingVector \inducingVector}}

\global\long\def\Kfu{\kernelMatrix_{\mappingFunctionVector \inducingVector}}
\global\long\def\Kuf{\kernelMatrix_{\inducingVector \mappingFunctionVector}}

\global\long\def\det#1{\left|#1\right|}

\global\long\def\tr#1{\text{tr}\left(#1\right)}


%% file: definitions.tex
\usepackage{color}
\usepackage{verbatim}

\definecolor{brown}{rgb}{0.9,0.59,0.078}
\definecolor{ironsulf}{rgb}{0,0.7,.5}
\definecolor{lightpurple}{rgb}{0.156,0,0.245}

\newenvironment{matlab}{\comment}{\endcomment}

\ifdefined\blackBackground
\definecolor{colorOne}{rgb}{0, 1, 1}
\definecolor{colorTwo}{rgb}{1, 0, 1}
\definecolor{colorThree}{rgb}{1, 1, 0}
\definecolor{colorTwoThree}{rgb}{1, 0, 0}
\definecolor{colorOneThree}{rgb}{0, 1, 0}
\definecolor{colorOneTwo}{rgb}{0, 0, 1}
\else
\definecolor{colorOne}{rgb}{1, 0, 0}
\definecolor{colorTwo}{rgb}{0, 1, 0}
\definecolor{colorThree}{rgb}{0, 0, 1}
\definecolor{colorTwoThree}{rgb}{0, 1, 1}
\definecolor{colorOneThree}{rgb}{1, 0, 1}
\definecolor{colorOneTwo}{rgb}{1, 1, 0}
\fi

\ifdefined\blackBackground

\global\long\def\blackColor{white}
\global\long\def\whiteColor{black}
\else

\global\long\def\blackColor{black}
\global\long\def\whiteColor{white}
\fi

\global\long\def\det#1{\left|#1\right|}

\global\long\def\bfmu{\boldsymbol{\mu}}

\global\long\def\bfSigma{\boldsymbol{\Sigma}}

\global\long\def\bflambda{\boldsymbol{\lambda}}
\global\long\def\bfLambda{\boldsymbol{\Lambda}}

\global\long\def\bfPsi{\boldsymbol{\Psi}}

\global\long\def\bftheta{\boldsymbol{\theta}}

\global\long\def\bfa{\mathbf{a}}

\global\long\def\bfm{\mathbf{m}}

\global\long\def\bft{\mathbf{t}}
\global\long\def\bfu{\mathbf{u}}

\global\long\def\bfx{\mathbf{x}}
\global\long\def\bfy{\mathbf{y}}
\global\long\def\bfz{\mathbf{z}}

\global\long\def\bfzero{\mathbf{0}}

\global\long\def\bfA{\mathbf{A}}
\global\long\def\bfB{\mathbf{B}}
\global\long\def\bfC{\mathbf{C}}

\global\long\def\bfI{\mathbf{I}}
\global\long\def\eye{\mathbf{I}}
\global\long\def\bfK{\mathbf{K}}

\global\long\def\bfM{\mathbf{M}}

\global\long\def\bfS{\mathbf{S}}

\global\long\def\bfW{\mathbf{W}}
\global\long\def\bfX{\mathbf{X}}
\global\long\def\bfY{\mathbf{Y}}
\global\long\def\bfZ{\mathbf{Z}}

\global\long\def\la{\leftarrow}

\global\long\def\T{{\top}}

\global\long\def\cut#1{}

\global\long\def\detail#1{}



%% file: graphicalModels.tex
\tikzstyle{latent} = [circle,inner sep=1pt,minimum size=20pt, font=\fontsize{10}{10}\selectfont, node distance=1,draw=\blackColor,fill=\whiteColor]
\tikzstyle{obs} = [latent,fill=\blackColor!30]
\tikzstyle{pixel} = [latent,minimum size=10pt, inner sep=0pt,node distance=0,font=\fontsize{5}{5}\selectfont]
\tikzstyle{const} = [rectangle, inner sep=0pt, node distance=1]
\tikzstyle{dash plate} = [draw, rectangle, rounded corners, fit=#1]
\tikzstyle{dash plate caption} = [caption, node distance=0, inner sep=0pt,
above=5pt and 0pt of #1.north] %

%% file: abstract.tex

\begin{abstract}%
  The Gaussian process latent variable model (GP-LVM) provides a
  flexible approach for non-linear dimensionality reduction that has been widely
  applied. However, the current approach for training GP-LVMs is based
  on maximum likelihood, where the latent projection variables are
  maximized over rather than integrated out.  In this paper we present
  a Bayesian method for training GP-LVMs by introducing a non-standard
  variational inference framework that allows to approximately
  integrate out the latent variables and subsequently train a GP-LVM
  by maximizing an analytic lower bound on the exact marginal
  likelihood.  We apply this method for learning a GP-LVM from iid
  observations and for learning non-linear dynamical systems where the
  observations are temporally correlated. We show that a benefit
  of the variational Bayesian procedure is its robustness to
  overfitting and its ability to automatically select the
  dimensionality of the nonlinear latent space.  The resulting
  framework is generic, flexible and easy to extend for other
  purposes, such as Gaussian process regression with uncertain inputs
  and semi-supervised Gaussian processes.
  We demonstrate our method on synthetic data and standard machine
  learning benchmarks, as well as challenging real world datasets,
  including high resolution video data.
\end{abstract}


%% file: introduction.tex
  \begin{matlab}
    importTool('dimred')
    dimredToolboxes
    randn('seed', 3e9)
    rand('seed', 3e9)
    if ~isoctave
      colordef white
    end
    patchColor = [0.8 0.8 0.8];
    lineColor = [0.2 0.2 0.2];
    redColor = [1 0 0];
    colorFigs = true;
    textWidth = 16
  \end{matlab}
\section{Introduction}

Consider a non linear function, $\mappingFunction(\latentScalar)$. A very general class of probability densities can be recovered by mapping a simpler density through the non linear function. For example, we might decide that $\latentScalar$ should be drawn from a Gaussian density,
\[
\latentScalar \sim \gaussianSamp{0}{1}
\]
and we observe $\dataScalar$, which is given by passing samples from $\latentScalar$ through a non linear function, perhaps with some corrupting noise,
\begin{equation}
\dataScalar = \mappingFunction(\latentScalar) + \epsilon
\label{generative}
\end{equation}
where $\epsilon$ could also be drawn from a Gaussian density,
\[
\epsilon \sim \gaussianSamp{0}{\dataStd^2},
\]
this time with variance $\dataStd^2$. Whilst the resulting density for $\dataScalar$, denoted by $p(\dataScalar)$, can now have a very general form, these models present particular problems in terms of tractability.
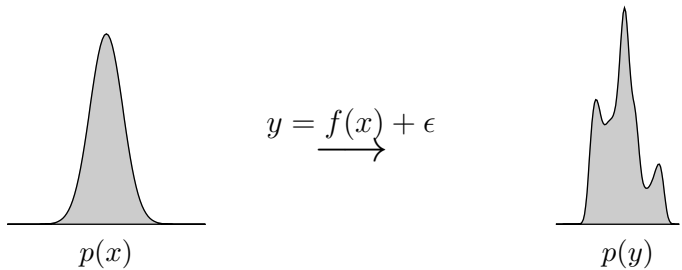
\begin{figure}
\begin{matlab}
    clf
    number = 200;
    rbfWidth = 2;
    numBasisFunc = 3;
    mu = linspace(-4, 4, numBasisFunc);
    x = linspace(-6, 6, number)';

    numSamples = 10;
    alpha = 1;
    dataStd = 0.2;

    numSamples = 10000;
    xsamp = randn(numSamples, 1);

    mu = linspace(-4, 4, 100);
    numBasisFunc = size(mu, 2);
    rbfWidth = 0.1;
    Phi = exp(-dist2(xsamp, mu')/(2*rbfWidth*rbfWidth));
    W = randn(1, numBasisFunc)*sqrt(alpha);
    f = Phi*W';

    axes('position', [0 0 0.4 1])
    hold on
    p = exp(-0.5/alpha*x.^2)*1/sqrt(2*pi*alpha);
    a = patch(x, p, patchColor);
    axis off
    set(a, 'linewidth', 2);
    set(gca, 'ylim', [0 0.5]);
    set(gca, 'xlim', [-6 6]);
    xlabel('$p(\latentScalar)$', 'horizontalalignment', 'center')

    axes('position', [0.6 0 0.4 1])
    y = linspace(min(f)-3*dataStd, max(f)+3*dataStd, 100)';
    p = mean(exp(-0.5/(dataStd*dataStd)*dist2(y, f))*1/(sqrt(2*pi)*dataStd), 2);
    a = patch(y, p, patchColor);
    axis off
    set(a, 'linewidth', 2);
    set(gca, 'ylim', [0 0.5])
    set(gca, 'xlim', [min(y) max(y)]);
    xlabel('$p(\dataScalar)$', 'horizontalalignment', 'center')

    axes('position', [0.1 0 0.8 1])
    set(gca, 'xlim', [0 1])
    set(gca, 'ylim', [0 1])
    axis off
    text(0.5, 0.45, '$\dataScalar = \mappingFunction(\latentScalar) + \noiseScalar$', 'horizontalalignment', 'center')
    text(0.5, 0.25, '$\longrightarrow$', 'horizontalalignment', 'center')

    options = printLatexOptions;
    options.maintainAspect = false;
    options.height = 0.1*textWidth;
    printLatexPlot('gaussianThroughNonlinear2', 'diagrams', textWidth, options);
    printLatexText(['\global\long\def\rbfWidthVal{' numsf2str(rbfWidth, 0) '}\global\long\def\dataStdVal{' strrep(numsf2str(dataStd, 1), ' ', '\ ') '}'], 'gaussianThroughNonlinear2Values.tex', 'diagrams');
  \end{matlab}
  \begin{centering}
    \input{gaussianThroughNonlinear2}
    \input{gaussianThroughNonlinear2Values}
  \end{centering}
  
  \caption{A Gaussian distribution propagated through a non-linear mapping. $\dataScalar_i=\mappingFunction(\latentScalar_i) + \noiseScalar_i$. $\noiseScalar \sim \gaussianSamp{0}{\dataStdVal^2}$ and $\mappingFunction(\cdot)$ uses RBF basis, 100 centres between -4 and 4 and $\rbfWidth=\rbfWidthVal$. The new distribution over $\dataScalar$ (right) is multimodal and difficult to normalize. }
  
\end{figure}

Models of this form appear in several domains. They can be used for  autoregressive prediction in time series \citep[see \eg][]{Girard:uncertain01} or prediction of a regression model output when the input is uncertain \citep[see \eg][]{Oakley:computer02}. \cite{MacKay:wondsa95} considered the same form for dimensionality reduction where several latent variables, $\latentVector = \{\latentScalar_j\}_{j=1}^\latentDim$ are used to represent a high dimensional vector $\dataVector = \{\dataScalar_j\}_{j=1}^\dataDim$ and we normally have  $\dataDim>\latentDim$,
\[
\dataVector = \mappingFunctionVector(\latentVector).
\] Adding a dynamical component to these nonlinear dimensionality reduction approaches leads to nonlinear state space models \citep{Sarkka:book13}, where the states often have a physical interpretation and are propagated through time in an autoregressive manner,
\[
\latentVector_t = \mappingFunctionTwoVector(\latentVector_{t-1}),
\]
where $\mappingFunctionTwoVector(\cdot)$ is a vector valued function. The observations are then observed through a separate nonlinear vector valued function,
\[
\dataVector_t = \mappingFunctionVector(\latentVector_t) + \noiseVector.
\]

The intractabilities of mapping a distribution through a nonlinear
function have resulted in a range of different approaches. In density
networks sampling was proposed; in particular, in
\citep{MacKay:wondsa95} \emph{importance sampling} was used. When
extending importance samplers dynamically, the degeneracy in the
weights needs to be avoided, thus leading to the resampling approach
suggested for the bootstrap particle filter of
\cite{Gordon:novel93}. Other approaches in nonlinear state space
models include the Laplace approximation as used in extended Kalman
filters and unscented and ensemble transforms
\citep[see][]{Sarkka:book13}. In dimensionality reduction the
generative topographic mapping \citep[GTM][]{Bishop:gtmncomp98}
reinterpreted the importance sampling approach of
\cite{MacKay:wondsa95} as a mixture of Gaussians model, using a
discrete representation of the latent space.

In this paper we suggest a variational approach to dealing with input
uncertainty that can be applied to Gaussian process models. Gaussian
processes provide a probabilistic framework for performing inference
over functions. A Gaussian process prior can be combined with a data
set (through an appropriate likelihood) to obtain a posterior process
that represents all functions that are consistent with the data and
our prior.

%
%
%
%
%
%
%


Our initial focus will be application of Gaussian process models in
the context of dimensionality reduction. In dimensionality reduction
we assume that our high dimensional data set is really the result of
some low dimensional control signals which are, perhaps, nonlinearly
related to our observed functions. In other words we assume that our
data, $\dataMatrix\in\Re^{\numData\times \dataDim}$, can be
approximated by a lower dimensional matrix, $\latentMatrix \in
\Re^{\numData\times\latentDim}$ through a vector valued function where
each row, $\dataVector_{i, :}$ of $\dataMatrix$ represents an observed
data point and is approximated through
\[
\dataVector_{i, :} = \mappingFunctionVector(\latentVector_{i, :}) + \noiseVector_{i, :},
\]
so that the data is a lower dimensional subspace immersed in the
original, high dimensional space. If the mapping is linear,
e.g. $\mappingFunctionVector(\latentVector_{i, :}) = \mappingMatrix
\latentVector_{i, :}$ with $\mappingMatrix \in \Re^{\latentDim \times
  \dataDim}$, methods like principal component analysis, factor
analysis and (for non-Gaussian $p(\latentVector_{i, :}$)) independent
component analysis \citep{Hyvarinen:icabook01} follow. For Gaussian $p(\latentVector_{i, :})$ the marginalization of the latent variable
is tractable because placing a Gaussian density through  an affine transformation retains the Gaussianity of the data density, $p(\dataVector_{i,
  :})$. However, the linear assumption is very restrictive so it is
natural to look to go beyond it through a non linear mapping.

  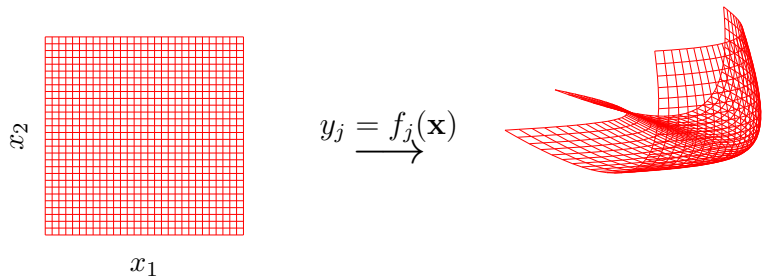
\begin{figure}
    \begin{matlab}
    number = 200;
    numBasisFunc = 3;
    rbfWidth = 2;
    mu = linspace(-4, 4, numBasisFunc);
    x = linspace(-6, 6, number)';
    numSamples = 10;
    alpha = 1;
    W = randn(numSamples, numBasisFunc)*sqrt(alpha);
    clf
    numberAcross = 30;
    x1 = linspace(-1, 1, numberAcross);
    x2 = x1;
    mu1 = mu;
    mu2 = mu;
    [MU1, MU2] = meshgrid(mu1, mu2);
    [X1, X2] = meshgrid(x1, x2);
    X = [X1(:) X2(:)];
    MU = [MU1(:) MU2(:)];
    numBasisFunc = size(MU, 1);
    number = size(X, 1);
    Phi = exp(-dist2(X, MU)/(2*rbfWidth*rbfWidth));
    numSamples = 3;
    W = randn(numSamples, numBasisFunc)*sqrt(alpha);
    F = Phi*W';
    axes('position', [0.1 0.333 0.333 0.3333])
    hold on
    startVal = 1;
    for i = 1:numberAcross
      endVal = numberAcross*i;
      a = plot(X(startVal:endVal, 1), X(startVal:endVal, 2), '-', 'color', lineColor);
      startVal = endVal + 1;
    end
    X1 = X1';
    X2 = X2';
    X = [X1(:) X2(:)];
    startVal = 1;
    for i = 1:numberAcross
      endVal = numberAcross*i;
      a = plot(X(startVal:endVal, 1), X(startVal:endVal, 2), '-', 'color', lineColor);
      startVal = endVal + 1;
    end
    hold off
    axis equal
    axis off
    xlabel('$\inputScalar_1$', 'horizontalalignment', 'center')
    ylabel('$\inputScalar_2$', 'horizontalalignment', 'center')

    axes('position', [0.5 0.3 0.4 0.4])
    hold on
    startVal = 1;
    for i = 1:numberAcross
      endVal = numberAcross*i;
      a = plot3(F(startVal:endVal, 1), F(startVal:endVal, 2), F(startVal:endVal, 3), '-', 'color', lineColor);
      startVal = endVal + 1;
    end
    F1 = reshape(F(:, 1), size(X1, 1), size(X1, 2))';
    F2 = reshape(F(:, 2), size(X1, 1), size(X1, 2))';
    F3 = reshape(F(:, 3), size(X1, 1), size(X1, 2))';
    F = [F1(:) F2(:) F3(:)];
    startVal = 1;
    for i = 1:numberAcross
      endVal = numberAcross*i;
      a = plot3(F(startVal:endVal, 1), F(startVal:endVal, 2), F(startVal:endVal, 3), '-', 'color', lineColor);
      startVal = endVal + 1;
    end
    hold off
    axis equal
    axis off
    pos = get(gca, 'position');
    npos = pos;
    npos(3:4) = pos(3:4)*3;
    npos(1) = pos(1) - 0.5*(npos(3)-pos(3));
    npos(2) = pos(2) - 0.5*(npos(4)-pos(4));
    set(gca, 'position', npos);

    axes('position', [0 0 1 1])
    set(gca, 'xlim', [0 1])
    set(gca, 'ylim', [0 1])
    axis off
    text(0.5, 0.55, '\large$\dataScalar_j = \mappingFunction_j(\inputVector)$', 'horizontalalignment', 'center')
    text(0.5, 0.45, '\Large$\longrightarrow$', 'horizontalalignment', 'center')

    paperPos = get(gcf, 'paperPosition');
    paperPos(4) = paperPos(4)/2;
    set(gcf, 'paperPosition', paperPos)
    options = printLatexOptions;
    options.maintainAspect = false;
    options.height = 0.5*textWidth;
    printLatexPlot('nonlinearMapping3DPlot', 'diagrams', 0.95*textWidth, options);

    \end{matlab}
    \begin{center}
      \input{nonlinearMapping3DPlot}
    \end{center}
    \caption{A three dimensional manifold formed by mapping from a two dimensional space to a three dimensional space.}
    \label{fig:nonlinearMapping3DPlot}
  \end{figure}

In the context of dimensionality reduction a range of approaches have
been suggested that consider neighborhood structures or the
preservation of local distances to find a low dimensional
representation. In the machine learning community, spectral methods
such as isomap \citep{Tenenbaum:isomap00}, locally linear embeddings
\citep[LLE,][]{Roweis:lle00} and Laplacian eigenmaps
\citep{Belkin:laplacian03} have attracted a lot of attention. These
spectral approaches are all closely related to kernel PCA
\citep{Scholkopf:nonlinear98} and classical multi-dimensional scaling
(MDS) \citep[see \eg][]{Mardia:multivariate79}. These methods do have a
probabilistic interpretation as described by
\cite{Lawrence:unifying12}, but it does not explicitly include an
assumption of underlying reduced data dimensionality. Other iterative
methods such as metric and non-metric approaches to MDS
\citep{Mardia:multivariate79}, Sammon mappings
\citep{Sammon:nonlinear69} and $t$-SNE \citep{Maaten:tsne08} also lack
an underlying generative model.

Probabilistic approaches, such as the generative topographic mapping
\citep[GTM,][]{Bishop:gtmncomp98} and density networks
\citep{MacKay:wondsa95}, view the dimensionality reduction problem
from a different perspective, since they seek a mapping from a
low-dimensional latent space to the observed data space 
(as illustrated in Figure \ref{fig:nonlinearMapping3DPlot}), and come with
certain advantages. More precisely, their generative nature and the
forward mapping that they define, allows them to be extended more
easily in various ways (\eg with additional dynamics modelling), to be
incorporated into a Bayesian framework for parameter learning and to
handle missing data. This approach to dimensionality
reduction provides a useful archetype for the algorithmic solutions we
are providing in this paper, as they require approximations that allow
latent variables to be propagated through a nonlinear function.

Our framework takes the generative approach prescribed by density
networks and the nonlinear variants of Kalman filters one step
further. Because, rather than considering a specific function,
$\mappingFunction(\cdot)$, to map from the latent variables to the
data space, we will consider an entire family of functions. One that
subsumes the more restricted class of either Gauss Markov processes
(such as the \emph{linear} Kalman filter/smoother) and Bayesian basis
function models (such as the RBF network used in the GTM, with a
Gaussian prior over the basis function weightings). These models can
all be cast within the framework of Gaussian processes
\citep{Rasmussen:book06}. Gaussian processes are probabilistic kernel
methods, where the kernel has an interpretation of a covariance
associated with a prior density. This covariance specifies a
distribution over functions that subsumes the special cases mentioned
above.

The Gaussian process latent variable model
\citep[GP-LVM,][]{Lawrence:pnpca05} is a more recent probabilistic
dimensionality reduction method which has been proven to be very
robust for high dimensional problems \citep{Lawrence:larger07,
  Damianou:vgpds11}. GP-LVM can be seen as a non-linear generalisation
of probabilistic PCA \citep[PPCA,][]{Tipping:probpca99,Roweis:spca97},
which also has a Bayesian interpretation \citep{Bishop:bayesPCA98}.
In contrast to PPCA, the non-linear mapping of GP-LVM makes a Bayesian
treatment much more challenging.  Therefore, GP-LVM itself and all of
its extensions, rely on a maximum a posteriori (MAP) training
procedure.  However, a principled Bayesian formulation is highly
desirable, since it would allow for robust training of the model,
automatic selection of the latent space's dimensionality as well as
more intuitive exploration of the latent space's structure.

In this paper we formulate a variational inference framework which
allows us to propagate uncertainty through a Gaussian process and
obtain a rigorous lower bound on the marginal likelihood of the
resulting model. The procedure followed here is non-standard, as
computation of a closed-form Jensen's lower bound on the true log
marginal likelihood of the data is infeasible with classical
approaches to variational inference. Instead, we build on, and
significantly extend, the variational GP method of
\cite{Titsias:variational09}, where the GP prior is augmented to
include auxiliary inducing variables so that the approximation is
applied on an expanded probability model. The resulting framework
defines an approximate bound on the evidence of the GP-LVM which, when
optimised, gives as a by-product an approximation to the true
posterior distribution of the latent variables given the data.

Considering a posterior distribution rather than point estimates for
the latent points means that our framework is generic and can be
easily extended for multiple practical scenarios. For example, if we
treat the latent points as noisy measurements of given inputs we
obtain a method for Gaussian process regression with uncertain inputs
\citep{Girard:uncertain01} or, in the limit, with partially observed inputs.
 On the other hand, considering a latent
space prior that depends on a time vector, allows us to obtain a
Bayesian model for dynamical systems \citep{Damianou:vgpds11} that
significantly extends classical Kalman filter models with a nonlinear
relationship between the state space, $\latentMatrix$, and the
observed data $\dataMatrix$, along with non-Markov assumptions in the
latent space which can be based on continuous time observations. This
is achieved by placing a Gaussian process prior on the latent space,
$\latentMatrix$ which is itself a function of time, $t$. This approach
can itself be trivially further extended by replacing the time time
dependency of the prior for the latent space with a spatial
dependency, or a dependency over an arbitrary number of high
dimensional inputs. As long as a valid \emph{covariance function}\footnote{The constraints for a valid covariance function are the same as those for a Mercer kernel. It must be a positive (semi) definite function over the space of all possible input pairs.} can be
derived (this is also possible for strings and graphs). This leads to
a Bayesian approach for warped Gaussian process regression
\citep{snelson:2004warped, lazaro:warped}.

In the next section we review the main prior work on dealing with
latent variables in the context of Gaussian processes and describe how
the model was extended with a dynamical component. We then introduce
the variational framework and Bayesian training procedure in Section
\ref{section:vgplvm}. In Section \ref{section:predictions} we describe how the
variational approach is applied to a range of predictive
tasks and this is demonstrated with experiments conducted on simulated and
real world datasets in Section \ref{section:experiments}.
In Section \ref{section:extensions} we discuss and experimentally demonstrate natural but important extensions
of our model, motivated by situations where the inputs to the GP are not fully
unobserved. These extensions give rise to an auto-regressive variant for performing
iterative future predictions and a semi-supervised GP variant.
Finally, based on the theoretical and experimental results of our work,
we present our final conclusions in Section \ref{section:conclusion}.



%% file: gaussianThroughNonlinear2.tex
\begingroup
  \makeatletter
  \providecommand\color[2][]{%
    \GenericError{(gnuplot) \space\space\space\@spaces}{%
      Package color not loaded in conjunction with
      terminal option `colourtext'%
    }{See the gnuplot documentation for explanation.%
    }{Either use 'blacktext' in gnuplot or load the package
      color.sty in LaTeX.}%
    \renewcommand\color[2][]{}%
  }%
  \providecommand\includegraphics[2][]{%
    \GenericError{(gnuplot) \space\space\space\@spaces}{%
      Package graphicx or graphics not loaded%
    }{See the gnuplot documentation for explanation.%
    }{The gnuplot epslatex terminal needs graphicx.sty or graphics.sty.}%
    \renewcommand\includegraphics[2][]{}%
  }%
  \providecommand\rotatebox[2]{#2}%
  \@ifundefined{ifGPcolor}{%
    \newif\ifGPcolor
    \GPcolortrue
  }{}%
  \@ifundefined{ifGPblacktext}{%
    \newif\ifGPblacktext
    \GPblacktexttrue
  }{}%
  \let\gplgaddtomacro\g@addto@macro
  \gdef\gplbacktext{}%
  \gdef\gplfronttext{}%
  \makeatother
  \ifGPblacktext
    \def\colorrgb#1{}%
    \def\colorgray#1{}%
  \else
    \ifGPcolor
      \def\colorrgb#1{\color[rgb]{#1}}%
      \def\colorgray#1{\color[gray]{#1}}%
      \expandafter\def\csname LTw\endcsname{\color{white}}%
      \expandafter\def\csname LTb\endcsname{\color{black}}%
      \expandafter\def\csname LTa\endcsname{\color{black}}%
      \expandafter\def\csname LT0\endcsname{\color[rgb]{1,0,0}}%
      \expandafter\def\csname LT1\endcsname{\color[rgb]{0,1,0}}%
      \expandafter\def\csname LT2\endcsname{\color[rgb]{0,0,1}}%
      \expandafter\def\csname LT3\endcsname{\color[rgb]{1,0,1}}%
      \expandafter\def\csname LT4\endcsname{\color[rgb]{0,1,1}}%
      \expandafter\def\csname LT5\endcsname{\color[rgb]{1,1,0}}%
      \expandafter\def\csname LT6\endcsname{\color[rgb]{0,0,0}}%
      \expandafter\def\csname LT7\endcsname{\color[rgb]{1,0.3,0}}%
      \expandafter\def\csname LT8\endcsname{\color[rgb]{0.5,0.5,0.5}}%
    \else
      \def\colorrgb#1{\color{black}}%
      \def\colorgray#1{\color[gray]{#1}}%
      \expandafter\def\csname LTw\endcsname{\color{white}}%
      \expandafter\def\csname LTb\endcsname{\color{black}}%
      \expandafter\def\csname LTa\endcsname{\color{black}}%
      \expandafter\def\csname LT0\endcsname{\color{black}}%
      \expandafter\def\csname LT1\endcsname{\color{black}}%
      \expandafter\def\csname LT2\endcsname{\color{black}}%
      \expandafter\def\csname LT3\endcsname{\color{black}}%
      \expandafter\def\csname LT4\endcsname{\color{black}}%
      \expandafter\def\csname LT5\endcsname{\color{black}}%
      \expandafter\def\csname LT6\endcsname{\color{black}}%
      \expandafter\def\csname LT7\endcsname{\color{black}}%
      \expandafter\def\csname LT8\endcsname{\color{black}}%
    \fi
  \fi
  \setlength{\unitlength}{0.0500bp}%
  \begin{picture}(7000.00,2210.00)%
    \gplgaddtomacro\gplbacktext{%
    }%
    \gplgaddtomacro\gplfronttext{%
      \colorrgb{0.00,0.00,0.00}%
      \put(3500,994){\makebox(0,0){\strut{}\large$\dataScalar = \mappingFunction(\latentScalar) + \noiseScalar$}}%
      \put(3500,773){\makebox(0,0){\strut{}\LARGE$\longrightarrow$}}%
    }%
    \gplgaddtomacro\gplbacktext{%
      \colorrgb{0.00,0.00,0.00}%
      \put(5588,23){\makebox(0,0){\strut{}$p(\dataScalar)$}}%
    }%
    \gplgaddtomacro\gplfronttext{%
    }%
    \gplgaddtomacro\gplbacktext{%
      \colorrgb{0.00,0.00,0.00}%
      \put(1656,23){\makebox(0,0){\strut{}$p(\latentScalar)$}}%
    }%
    \gplgaddtomacro\gplfronttext{%
    }%
    \gplbacktext
    \put(0,0){\includegraphics{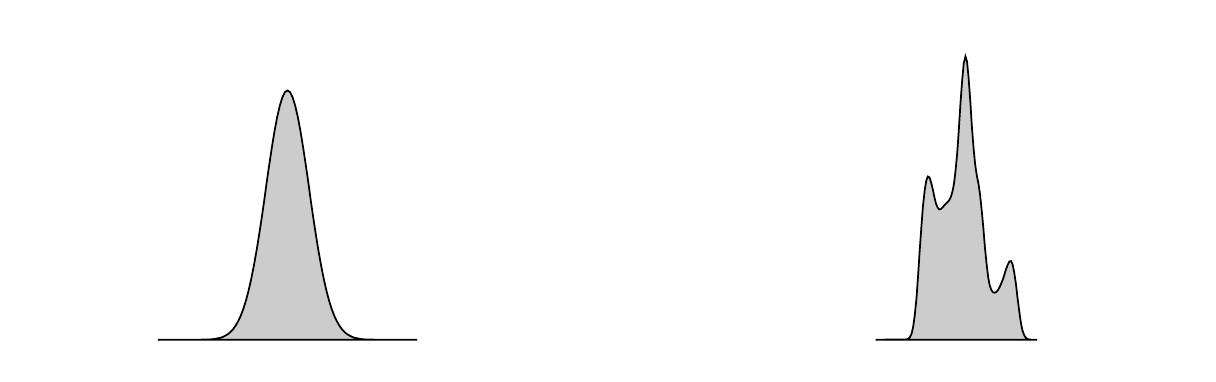}}%
    \gplfronttext
  \end{picture}%
\endgroup

%% file: gaussianThroughNonlinear2Values.tex
\global\long\def\rbfWidthVal{0.1}\global\long\def\dataStdVal{0.2}

%% file: nonlinearMapping3DPlot.tex
\begingroup
  \makeatletter
  \providecommand\color[2][]{%
    \GenericError{(gnuplot) \space\space\space\@spaces}{%
      Package color not loaded in conjunction with
      terminal option `colourtext'%
    }{See the gnuplot documentation for explanation.%
    }{Either use 'blacktext' in gnuplot or load the package
      color.sty in LaTeX.}%
    \renewcommand\color[2][]{}%
  }%
  \providecommand\includegraphics[2][]{%
    \GenericError{(gnuplot) \space\space\space\@spaces}{%
      Package graphicx or graphics not loaded%
    }{See the gnuplot documentation for explanation.%
    }{The gnuplot epslatex terminal needs graphicx.sty or graphics.sty.}%
    \renewcommand\includegraphics[2][]{}%
  }%
  \providecommand\rotatebox[2]{#2}%
  \@ifundefined{ifGPcolor}{%
    \newif\ifGPcolor
    \GPcolortrue
  }{}%
  \@ifundefined{ifGPblacktext}{%
    \newif\ifGPblacktext
    \GPblacktexttrue
  }{}%
  \let\gplgaddtomacro\g@addto@macro
  \gdef\gplbacktext{}%
  \gdef\gplfronttext{}%
  \makeatother
  \ifGPblacktext
    \def\colorrgb#1{}%
    \def\colorgray#1{}%
  \else
    \ifGPcolor
      \def\colorrgb#1{\color[rgb]{#1}}%
      \def\colorgray#1{\color[gray]{#1}}%
      \expandafter\def\csname LTw\endcsname{\color{white}}%
      \expandafter\def\csname LTb\endcsname{\color{black}}%
      \expandafter\def\csname LTa\endcsname{\color{black}}%
      \expandafter\def\csname LT0\endcsname{\color[rgb]{1,0,0}}%
      \expandafter\def\csname LT1\endcsname{\color[rgb]{0,1,0}}%
      \expandafter\def\csname LT2\endcsname{\color[rgb]{0,0,1}}%
      \expandafter\def\csname LT3\endcsname{\color[rgb]{1,0,1}}%
      \expandafter\def\csname LT4\endcsname{\color[rgb]{0,1,1}}%
      \expandafter\def\csname LT5\endcsname{\color[rgb]{1,1,0}}%
      \expandafter\def\csname LT6\endcsname{\color[rgb]{0,0,0}}%
      \expandafter\def\csname LT7\endcsname{\color[rgb]{1,0.3,0}}%
      \expandafter\def\csname LT8\endcsname{\color[rgb]{0.5,0.5,0.5}}%
    \else
      \def\colorrgb#1{\color{black}}%
      \def\colorgray#1{\color[gray]{#1}}%
      \expandafter\def\csname LTw\endcsname{\color{white}}%
      \expandafter\def\csname LTb\endcsname{\color{black}}%
      \expandafter\def\csname LTa\endcsname{\color{black}}%
      \expandafter\def\csname LT0\endcsname{\color{black}}%
      \expandafter\def\csname LT1\endcsname{\color{black}}%
      \expandafter\def\csname LT2\endcsname{\color{black}}%
      \expandafter\def\csname LT3\endcsname{\color{black}}%
      \expandafter\def\csname LT4\endcsname{\color{black}}%
      \expandafter\def\csname LT5\endcsname{\color{black}}%
      \expandafter\def\csname LT6\endcsname{\color{black}}%
      \expandafter\def\csname LT7\endcsname{\color{black}}%
      \expandafter\def\csname LT8\endcsname{\color{black}}%
    \fi
  \fi
  \setlength{\unitlength}{0.0500bp}%
  \begin{picture}(7000.00,2210.00)%
    \gplgaddtomacro\gplbacktext{%
    }%
    \gplgaddtomacro\gplfronttext{%
      \colorrgb{0.00,0.00,0.00}%
      \put(3500,1215){\makebox(0,0){\strut{}\large$\dataScalar_j = \mappingFunction_j(\inputVector)$}}%
      \put(3500,994){\makebox(0,0){\strut{}\LARGE$\longrightarrow$}}%
    }%
    \gplgaddtomacro\gplbacktext{%
    }%
    \gplgaddtomacro\gplfronttext{%
    }%
    \gplgaddtomacro\gplbacktext{%
      \colorrgb{0.00,0.00,0.00}%
      \put(689,1143){\rotatebox{90}{\makebox(0,0){\strut{}$\inputScalar_2$}}}%
      \colorrgb{0.00,0.00,0.00}%
      \put(1656,176){\makebox(0,0){\strut{}$\inputScalar_1$}}%
    }%
    \gplgaddtomacro\gplfronttext{%
    }%
    \gplbacktext
    \put(0,0){\includegraphics{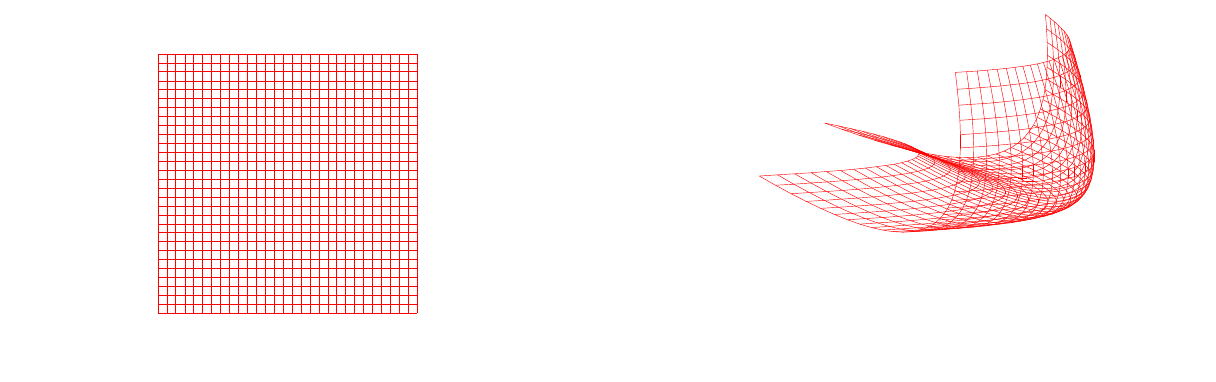}}%
    \gplfronttext
  \end{picture}%
\endgroup

%% file: background.tex

\section{Gaussian Processes with Latent Variables as Inputs \label{section:background}}

This section provides background material on current approaches for learning using 
Gaussian process latent variables models (GP-LVMs). Specifically, section \ref{sec:standardGPLVM} specifies
the general structure of such models, section
\ref{section:gplvmDynamics} reviews the standard GP-LVM for i.i.d.\ data as well as dynamic extensions 
suitable for sequence data. Finally, section \ref{mapCriticism} discusses the drawbacks of MAP 
estimation over the latent variables which is currently the standard way to train GP-LVMs. 

\subsection{Gaussian Processes for Latent Mappings \label{sec:standardGPLVM}}

The unified characteristic of all GP-LVM algorithms, as they were first introduced by 
\citet{Lawrence:pnpca05, Lawrence:gplvm03}, is the consideration of a Gaussian Process
as a prior distribution for the mapping function
$\mappingFunctionVector(\latentVector) = (\mappingFunction_1(\latentVector), \ldots, \mappingFunction_p(\latentVector))$ so that, 
\begin{equation}
\label{gppriorf}
\mappingFunction_j(\latentVector)  \sim  \mathcal{GP}(0, \kernelScalar_\mappingFunction(\latentVector,\latentVector^\prime)), \ \ \outputIndex=1,\ldots,\dataDim.
\end{equation}
Here, the individual components of $\mappingFunctionVector(\latentVector)$ are taken to be independent draws from a Gaussian
process with kernel or covariance function $\kernelScalar_\mappingFunction(\latentVector,\latentVector^\prime)$, which 
determines the properties of the latent
mapping. As shown in \citep{Lawrence:pnpca05} the use  of a linear covariance function 
makes GP-LVM equivalent to traditional PPCA. On the the other hand,  
when nonlinear covariance functions are considered the model is able to perfom non-linear dimensionality reduction. 
The non-linear covariance function considered in \citep{Lawrence:pnpca05} is the exponentiated quadratic (RBF),
\begin{align}
\kernelScalar_{\mappingFunction(\text{rbf})} \left( \latentVector_{\dataIndexOne, :}, \latentVector_{k, :} \right) = {} &  
        \sigma_{\text{rbf}}^2 \exp\left(
            - \frac{1}{2 \lengthScale^2} \sum_{j=1}^{\latentDim} \left(
                          \mathit{\latentScalar_{\dataIndexOne,j} - \latentScalar_{k,j}} \right) ^2 \right),
\label{rbf}
\end{align}
\noindent which is infinitely many times differentiable and it uses a common lengthscale
 parameter for all latent dimensions.
The above covariance function results in a non-linear but smooth mapping from the latent
to the data space.
Parameters that appear in a covariance function, such as $\sigma_{\text{rbf}}^2$ and $\ell^2$, are often 
referred to as kernel hyperparameters and will be denoted by $\bftheta_\mappingFunction$ throughout the paper.



\par Given the independence assumption across dimensions in equation \eqref{gppriorf}, the latent variables 
$\mappingFunctionMatrix \in \Re^{\numData \times \dataDim}$ (with columns $\{ \mappingFunctionVector_{:, \outputIndex} \}_{\outputIndex=1}^\dataDim$), 
which have one-to-one correspondance 
with the data points $\dataMatrix$,  follow the prior distribution
$p(\mappingFunctionMatrix | \latentMatrix, \bftheta_\mappingFunction) = \prod_{\outputIndex=1}^\dataDim p(\mappingFunctionVector_{:,\outputIndex} | \latentMatrix, \bftheta_\mappingFunction)$, where
$p(\mappingFunctionVector_{:, \outputIndex} | \latentMatrix, \bftheta_\mappingFunction)$ is given by
\begin{align}
p(\mappingFunctionVector_{:, \outputIndex} | \latentMatrix, \bftheta_\mappingFunction) &= \gaussianDist{\mappingFunctionVector_{:, \outputIndex}}{\zerosVector}{\Kff} 
                     = \det{2\pi\Kff}^{-\frac{1}{2}} 
                        \exp \left( - \frac{1}{2} \mappingFunctionVector_{:, \outputIndex}^\top \Kff^{-1} \mappingFunctionVector_{:, \outputIndex} \right) , \label{priorF}
\end{align}
and where $\Kff= \kernelScalar_\mappingFunction(\latentMatrix,\latentMatrix)$ is the covariance matrix
defined by the kernel function $\kernelScalar_\mappingFunction$.
The inputs $\latentMatrix$ in this kernel matrix are latent random variables 
following a prior distribution $p(\latentMatrix | \bftheta_\latentScalar)$ with hyperparameters 
$\bftheta_\latentScalar$. The structure of this prior can depend on the application 
at hand, such as on whether the observed data are i.i.d.\ or have a sequential dependence. 
For the remaining of this section we shall leave $p(\latentMatrix | \bftheta_\latentScalar)$
unspecified so that to keep our discussion general while specific forms for it will be given 
in the next section.
 

Given the construction outlined above, the joint probabibility density over the observed data and all latent 
variables is written as follows,
\begin{equation}
\label{joint}
p(\dataMatrix,\mappingFunctionMatrix,\latentMatrix, \bftheta_\mappingFunction, \bftheta_\latentScalar, \dataStd^2 ) = p(\dataMatrix|\mappingFunctionMatrix, \dataStd^2 ) p(\mappingFunctionMatrix|\latentMatrix, \bftheta_\mappingFunction) p(\latentMatrix | \bftheta_\latentScalar)= 
 \prod_{\outputIndex=1}^\dataDim  p(\mathbf{y}_{:,\outputIndex} | \mappingFunctionVector_{:, \outputIndex}, \dataStd^2 ) p(\mappingFunctionVector_{:, \outputIndex} | \latentMatrix, \bftheta_\mappingFunction) p(\latentMatrix | \bftheta_\latentScalar),
\end{equation}
where the term
\begin{equation}
\label{eq:yGivenF}
p(\dataMatrix|\mappingFunctionMatrix, \dataStd^2) = 
\prod_{\outputIndex=1}^\dataDim 
\gaussianDist{\bfy_{:,\outputIndex}}{\mappingFunctionVector_{:, \outputIndex}}{\dataStd^2 \eye_n}
\end{equation}
comes directly from the assumed noise model of equation \eqref{generative} while 
$p(\mappingFunctionMatrix|\latentMatrix, \bftheta_\mappingFunction)$ and $p(\latentMatrix | \bftheta_\latentScalar)$ come from the GP and the latent space.
As discussed in detail in Section \ref{section:meanField}, the interplay of the latent variables 
(i.e.\ the latent matrix $\latentMatrix$ that is passed as input in the latent matrix $\mappingFunctionMatrix$)  makes 
inference very challenging. However, when fixing $\latentMatrix$ we can treat 
 $\mappingFunctionMatrix$ analytically and marginalise it out as follows,
\begin{equation*}
 \label{gplvmLikelihood}
p(\dataMatrix | \latentMatrix) p(\latentMatrix) = \left( \int p \left( \dataMatrix | \mappingFunctionMatrix \right) p(\mappingFunctionMatrix | \latentMatrix) \intd \mappingFunctionMatrix \right) p(\latentMatrix),
\end{equation*}
where
\begin{equation}
\label{pyx}
p(\dataMatrix | \latentMatrix) = \prod_{\outputIndex = 1}^\dataDim 
\gaussianDist{\dataVector_{:,\outputIndex}}{\zerosVector}{\Kff + \dataStd^2 \eye_\numData}.
\end{equation}
Here (and for the remaining of the paper), we omit refererence to  
the parameters $\bftheta = \{ \bftheta_\mappingFunction, \bftheta_\latentScalar, \dataStd^2 \}$ in order to simplify our 
notation. The above partial tractability of the model gives rise to a straightforward 
MAP training procedure where the latent inputs $\latentMatrix$ are selected according to
\begin{equation*}
\latentMatrix_{\text{MAP}} = \arg \max_{\latentMatrix}  p(\dataMatrix|\latentMatrix)  p(\latentMatrix).
\end{equation*}
This is the approach suggested by \cite{Lawrence:pnpca05,Lawrence:gplvmtut06} and subsequently followed by other authors \citep{Urtasun:dgplvm07,Ek:pose07,Ferris:wifi07,Wang:gpdm08,Ko:bayesian09, Fusi:detecting13, Lu:GaussianFace14}. 
Finally, notice that point estimates over the hyperparameters $\bftheta$ can also be
found by maximising the same objective function.  


\subsection{Different Latent Space Priors and  GP-LVM Variants 
\label{section:gplvmDynamics}}


Different GP-LVM algorithms can result by varying the structure of the prior 
distribution $p(\latentMatrix)$ over the latent inputs. The simplest case, which is suitable for i.i.d.\ observations, is obtained 
by selecting a fully factorized (across data points and dimemsions) latent space prior: 
\begin{equation}
\label{standardNormal}
p(\latentMatrix) = \prod_{i=1}^\numData \gaussianDist{\latentVector_{i, :}}{\zerosVector}{\eye_\latentDim}
= \prod_{i=1}^\numData \prod_{j=1}^\latentDim \gaussianDist{x_{i,j}}{0}{1}.
\end{equation}

\noindent More structured
latent space priors can also be used that could incorporate available 
information about the problem at hand.  For
example, \cite{Urtasun:dgplvm07} add discriminative properties to the
GP-LVM by considering priors which encapsulate class-label
information.  Other existing approaches in the literature seek to
constrain the latent space via a smooth dynamical prior
$p(\latentMatrix)$ so as to obtain a model for dynamical systems.  
For example, \cite{Wang:gpdm05,Wang:gpdm08} extend GP-LVM with a
temporal prior which encapsulates the Markov property,
resulting in an auto-regressive model. \cite{Ko:bayesfilter09,GP-Based2} further extend these models for
Bayesian filtering in a robotics setting, whereas
\cite{Urtasun:3dpeople06} consider this idea for tracking.
In a similar direction, \cite{Lawrence:hgplvm07} consider an additional temporal model 
which employs a GP prior that is able to generate smooth paths in the
latent space.

In this paper we shall focus on dynamical variants where the dynamics are regressive, as in \citep{Lawrence:hgplvm07}. In this setting, the data are assumed to be a multivariate
timeseries $\{\dataVector_{i, :},t_i\}_{i=1}^\numData$  where $t_i \in \Re_+$ is the time at which the datapoint $\dataVector_{i, :}$ is observed. A GP-LVM  dynamical model 
is obtained by defining a temporal latent function $\latentVector(t) = (\latentScalar_1(t),\ldots, \latentScalar_{\latentDim}(t))$  where the individual components are taken to be independent draws from a Gaussian process, 
\begin{equation}
  \label{eq:xt}
  \latentScalar_k(t)  \sim \mathcal{GP}(0, \kernelScalar_\latentScalar(t_i,t_j)), \ \ k=1,\ldots,\latentDim,    
\end{equation}
where $\kernelScalar_\latentScalar(t_i,t_j)$ is the covariance function. 
The datapoint $\dataVector_{i, :}$ is assumed to be  produced via the latent vector 
$\latentVector_{i, :} = \latentVector(t_i)$, as shown in Figure
\ref{fig:graphicalModels}\subref{fig:dbgplvm}. All these latent vectors  
can be stored in the matrix $\latentMatrix$ (exactly as in the i.i.d.\ data case) 
which now follows the correlated prior distribution,  
\begin{equation}
p(\latentMatrix | \bft)  = \prod_{j=1}^\latentDim p(\latentVector_{:, j}|\mathbf{t}) = 
\prod_{j=1}^\latentDim \gaussianDist{\latentVector_{:, j}}{\zerosVector}{
  \Kx},
\label{priorXgivenT}
\end{equation}
\noindent where $\Kx =
\kernelScalar_\latentScalar(\bft,\bft)$ is the covariance matrix
obtained by evaluating the covariance function
$\kernelScalar_\latentScalar$ on the observed times $\bft$. In
contrast to the fully factorized prior in \eqref{standardNormal}, the above prior
couples all elements in each row of $\latentMatrix$. 
The covariance function $\kernelScalar_\latentScalar$ has parameters
$\bftheta_\latentScalar$ and determines the
properties of each temporal function $\latentScalar_\latentIndex(t)$.  For
instance, the use of an Ornstein-Uhlbeck covariance function yields a
Gauss-Markov process for $\latentScalar_\latentIndex(t)$, while the exponentiated
quadratic covariance function gives rise to very smooth and
non-Markovian process. The specific choices and forms of the
covariance functions used in our experiments are discussed in section
\ref{covarianceFunctions}.

\subsection{Drawbacks of the MAP Training Procedure \label{mapCriticism}}

Current GP-LVM based models found in the literature rely on MAP
training procedures, discussed in Section \ref{sec:standardGPLVM}, for optimizing the latent inputs and the hyperparameters. However,
this approach has several drawbacks. Firstly, the fact that it does not 
marginalise out the latent inputs implies that it could be sensitive to 
overfitting. Further, the MAP objective
function cannot provide any insight for selecting the optimal number of latent
dimensions, since it typically increases when more dimensions are added.
This is why most GP-LVM algorithms found in the literature require
the latent dimensionality to be either set
by hand or selected with cross-validation. The
latter case renders the whole training computationally
slow and, in practice, only a very limited subset of models can be explored in a reasonable time. 

As another consequence of the 
above, the current GP-LVMs employ simple covariance functions 
(typically having a common lengthscale over the latent input demensions as 
the one in  equation \eqref{rbf}) while more complex covariance functions, that could help to automatically 
select the latent dimensionality, are not popular. Such a latter covariance function can be an 
exponentiated quadratic, as in \eqref{rbf}, but with different 
lengthscale (or weight) per input dimension, 
\begin{align}
  \kernelScalar_{\mappingFunction(\text{ard})} \left(
    \latentVector_{i, :}, \latentVector_{k, :} \right) = {} &
  \sigma_{\text{ard}}^2 \exp \left( - \frac{1}{2}
    \sum_{j=1}^{\latentDim} w_j \left( \mathit{\latentScalar_{i,j} -
        \latentScalar_{k,j}} \right) ^2 \right).
  \label{ard}
\end{align}
This covariance function could allow an Automatic Relevance
Determination (ARD) procedure to take place, during which unnecessary
dimensions of the latent space $\latentMatrix$ are assigned a weight
$w_k$ with value almost zero. However, with the standard MAP training approach 
the benefits of using the above covariance function cannot be realised as 
typically overfitting will occur.    

Therefore, it is clear that the development of more fully Bayesian approaches 
for training GP-LVMs could make these 
models more reliable and provide rigorous solutions to the limitations of MAP training. 
The variational method presented in the next section is such an approach that, 
as demonstrated in the experiments, shows great ability in avoiding overfitting 
and permits automatic selection of the latent dimensionality.  
 

%% file: vargplvm.tex

 \section{\label{section:vgplvm}Variational Gaussian Process Latent Variable Models}
 
In this section we describe in detail our proposed method which is based 
on a non-standard variational approximation that utilises auxiliary variables. 
The resulting class of training algorithms will be referred to as 
\emph{Variational Gaussian Process Latent Variable Models}, or simply \emph{variational GP-LVMs}.
 
We start with section \ref{section:meanField} where we explain the obstacles  
we need to overcome when applying variational methods to the GP-LVM 
and specifically why the standard mean field approach is not immediately tractable.
In Section \ref{sec:lowerBoundWithAuxVars}, 
we show how the use of auxiliary variables together with a certain 
variational distribution results in a tractable approximation. 
In Section \ref{section:priors} we give specific details about how to apply 
our framework to the two different GP-LVM variants that this paper is concerned with:
the standard GP-LVM and the dynamical/warped one.
Finally, we outline two extensions of our variational method
that enable its application in more specific modelling scenarios. In the end of Section 
\ref{temporalPrior} we explain
how multiple independent time-series can be accommodated within the same dynamical model
and in Section \ref{sec:HighDimensionalData} we describe a simple trick that makes the model (and, in fact, any 
GP-LVM model) applicable to vast dimensionalities.

\subsection{\label{section:meanField} Standard Mean Field is Challenging for  GP-LVM}

A Bayesian treatment of the GP-LVM requires the computation of the log marginal likelihood
associated with the joint distribution of equation \eqref{joint}.
Both sets of unknown
random variables have to be marginalised out: the mapping values $\mappingFunctionMatrix$ (as in the standard model)
and the latent space $\latentMatrix$.
Thus, the required integral is written as,
\begin{align}
\log p(\dataMatrix) &= \log \int p(\dataMatrix, \mappingFunctionMatrix, \latentMatrix) \intd \latentMatrix \intd \mappingFunctionMatrix
                  =  \log \int p( \dataMatrix | \mappingFunctionMatrix) p(\mappingFunctionMatrix | \latentMatrix ) p(\latentMatrix) \intd  \latentMatrix \intd \mappingFunctionMatrix \label{marginalLikelihood} \\
              &= \log \int p(\dataMatrix|\mappingFunctionMatrix) \left( \int p(\mappingFunctionMatrix|\latentMatrix) p(\latentMatrix)
                   \intd \latentMatrix \right) \intd \mappingFunctionMatrix \label{marginalLikelihood2}.
\end{align}


\noindent The key difficulty with this Bayesian approach is propagating the prior
density $p(\latentMatrix)$ through the nonlinear mapping.
Indeed, the nested integral in equation \eqref{marginalLikelihood2} can be written
as $\int p(\latentMatrix)  \prod_{j=1}^\dataDim p(\mappingFunctionVector_{:, \outputIndex} | \latentMatrix) \intd \mappingFunctionMatrix$ 
where each term $p(\mappingFunctionVector_{:, \outputIndex} | \latentMatrix)$, given by \eqref{priorF}, is proportional to  
$|\Kff|^{-\frac{1}{2}} \exp \left( - \frac{1}{2} \mappingFunctionVector_{:, \outputIndex}^\top \Kff^{-1} \mappingFunctionVector_{:, \outputIndex} \right)$.
Clearly, this term contains $\latentMatrix$, which are the inputs of the kernel matrix $\Kff$, 
in a rather very complex non-linear manner and therefore analytical integration over $\latentMatrix$ is infeasible.   

To make progress, we can invoke the standard variational Bayesian methodology \citep{Bishop:book06} 
to approximate the marginal likelihood of equation \eqref{marginalLikelihood} with a variational lower bound.
Specifically, we can introduce a factorised variational distribution over the 
unknown random variables,
\begin{equation}
\label{eq:GPLVMvardistr}
q(\mappingFunctionMatrix,\latentMatrix)=q(\mappingFunctionMatrix)q(\latentMatrix),
\end{equation}
which aims at approximating the true posterior 
$p(\mappingFunctionMatrix|\dataMatrix,\latentMatrix) p(\latentMatrix|\dataMatrix)$. Based on Jensen's inequality, we can obtain the standard variational lower bound on the log marginal likelihood,
\begin{equation}
\log p(\dataMatrix) \ge 
       \int q(\mappingFunctionMatrix)q(\latentMatrix) \log \frac{p(\dataMatrix|\mappingFunctionMatrix)
       p(\mappingFunctionMatrix|\latentMatrix) p(\latentMatrix)}{q(\mappingFunctionMatrix)q(\latentMatrix)} \intd \mappingFunctionMatrix \intd \latentMatrix . \label{jensens1}
\end{equation}

%
%
%
 %
Nevertheless, this standard {\em mean field} approach remains problematic because the lower bound above is still 
intractable to compute. To isolate the intractable term, observe that \eqref{jensens1} can be written as
\begin{equation}
\log p(\dataMatrix) \ge 
       \int q(\mappingFunctionMatrix)q(\latentMatrix) \log   p(\mappingFunctionMatrix|\latentMatrix) \intd \mappingFunctionMatrix \intd \latentMatrix 
       +
       \int q(\mappingFunctionMatrix)q(\latentMatrix) \log \frac{p(\dataMatrix|\mappingFunctionMatrix)
       p(\latentMatrix)}{q(\mappingFunctionMatrix)q(\latentMatrix)} \intd \mappingFunctionMatrix \intd \latentMatrix, \label{jensens2}
\end{equation}
where the first term of the above equation contains the expectation of
$\log p(\mappingFunctionMatrix|\latentMatrix)$ under the distribution $q(\latentMatrix)$. This requires 
an integration over $\latentMatrix$ which appears nonlinearly in 
$\Kff^{-1}$ and $\log |\Kff|$ and cannot be done analytically.
Therefore, standard mean field variational methodologies do not lead to an analytically
tractable variational lower bound.

\subsection{\label{sec:lowerBoundWithAuxVars}Tractable Lower Bound by Introducing Auxiliary Variables}
\par In contrast, our framework allows us to compute
a closed-form Jensen's lower bound
by applying variational inference after expanding the GP prior so as to include auxiliary inducing
variables. Originally, inducing variables were introduced for computational speed ups in GP regression models
\citep{Csato:sparse02,Seeger:fast03,Csato:thesis02,Snelson:pseudo05,Quinonero:unifying05,Titsias:variational09}.
 In our approach, these extra variables will be used within the variational sparse 
 GP framework of \cite{Titsias:variational09}.

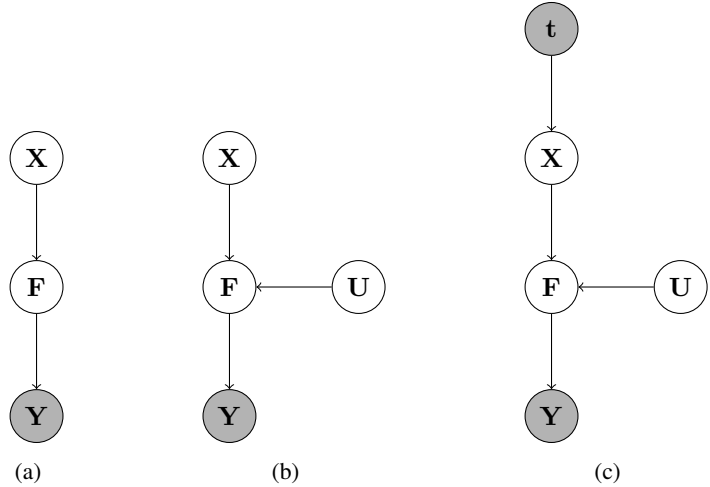
\begin{figure}[ht]
\begin{center}
  \def\layersep{0.75cm}
  \def\nodesep{1cm}
  \subfigure[]{
    \label{fig:gplvm}
    \begin{tikzpicture}[node distance=\layersep]
    
    \draw node[obs] (y) {$\dataMatrix$};
    \draw node[latent, above=of y] (f) {$\mappingFunctionMatrix$};
    \draw node[latent, above=of f] (x) {$\latentMatrix$};
    
    \draw [->] (x) to (f);%
    \draw [->] (f) to (y);%
    
    
  \end{tikzpicture}}
\hspace{0.1\textwidth}
  \subfigure[]{
  \label{fig:bgplvm}
    \begin{tikzpicture}[node distance=\layersep]
    
    \draw node[obs] (y) {$\dataMatrix$};
    \draw node[latent, above=of y] (f) {$\mappingFunctionMatrix$};
    \draw node[latent, above=of f] (x) {$\latentMatrix$};
    \draw node[latent, right=of f] (u) {$\inducingMatrix$};
    
    \draw [->] (x) to (f);%
    \draw [->] (f) to (y);%
    \draw [->] (u) to (f);%
    
    
  \end{tikzpicture}}
\hspace{0.1\textwidth}
  \subfigure[]{
    \label{fig:dbgplvm}
  \begin{tikzpicture}[node distance=\layersep]
    
    \draw node[obs] (y) {$\dataMatrix$};
    \draw node[latent, above=of y] (f) {$\mappingFunctionMatrix$};
    \draw node[latent, above=of f] (x) {$\latentMatrix$};
    \draw node[latent, right=of f] (u) {$\inducingMatrix$};
    \draw node[obs, above=of x] (t) {$\mathbf{t}$};
    
    \draw [->] (x) to (f);%
    \draw [->] (f) to (y);%
    \draw [->] (u) to (f);%
    \draw [->] (t) to (x);%
    
    
  \end{tikzpicture}}
\end{center}
\vspace{-8pt}
\caption{ \small{
The graphical model for the GP-LVM \subref{fig:gplvm} is augmented with auxiliary variables to obtain the
variational GP-LVM model \subref{fig:bgplvm} and its dynamical version
\subref{fig:dbgplvm}. 
Shaded nodes represent
observed variables. In general, the top level input in \subref{fig:dbgplvm} can be arbitrary, depending
on the application.}}
\label{fig:graphicalModels}
\end{figure}

More specifically, we expand the joint
probability model in (\ref{joint}) by including $\numInducing$ extra samples (inducing points) of
the GP latent mapping $\mappingFunctionVector({\bf x})$, so that
$\inducingVector_{\dataIndex, :} \in \mathbb{R}^\dataDim$ is such a sample. The inducing points are
collected in a matrix $\inducingMatrix \in \mathbb{R}^{\numInducing \times \dataDim}$ 
and constitute latent function evaluations at a set of pseudo-inputs 
$\latentMatrix_u \in \mathbb{R}^{\numInducing \times \latentDim}$.
%
%
The augmented joint probability density takes the form,
\begin{align}
p(\dataMatrix,\mappingFunctionMatrix, \inducingMatrix,\latentMatrix)  
=& p(\dataMatrix|\mappingFunctionMatrix) 
   p(\mappingFunctionMatrix|\inducingMatrix,\latentMatrix,\latentMatrix_u)
   p(\inducingMatrix|\latentMatrix_u) p(\latentMatrix) \nonumber \\
=& \left( \prod_{\outputIndex=1}^\dataDim 
p(\dataVector_{:, \outputIndex} | \mappingFunctionVector_{:, \outputIndex}) p(\mappingFunctionVector_{:, \outputIndex} | \inducingVector_{:, \outputIndex}, \latentMatrix, \latentMatrix_u) 
p(\inducingVector_{:, \outputIndex} | \latentMatrix_u) \right) p(\latentMatrix)     \label{augmentedJoint},
\end{align}

where
\begin{align}
p(\mappingFunctionVector_{:, \outputIndex} | \inducingVector_{:, \outputIndex},\latentMatrix,\latentMatrix_u) = \gaussianDist{\mappingFunctionVector_{:, \outputIndex}}{\mathbf{a}_\outputIndex}{\bfSigma_f} \label{priorF2},
\end{align}
with
\begin{equation}
\label{eq:conditionalGPmeanCovar}
\mathbf{a}_j = \Kfu \Kuu^{-1} \inducingVector_{:, \outputIndex} \text{\; \; and \; \; }
\bfSigma_f = \Kff - \Kfu \Kuu^{-1} \Kuf
\end{equation}
is the conditional GP prior (see \eg \cite{Rasmussen:book06}) and 
\begin{equation}
\label{pfu}
p(\inducingVector_{:, \outputIndex}|\latentMatrix_u) = \mathcal{N}(\inducingVector_{:, \outputIndex}|\mathbf{0},
\Kuu),
\end{equation}
is the marginal GP
prior over the inducing variables. In the above expressions, $\Kuu$ denotes
the covariance matrix constructed by evaluating the covariance function 
on the inducing points, $\Kuf$ is the cross-covariance between the inducing
and the latent points and $\Kfu = \Kuf^\top$.
Figure \ref{fig:bgplvm} graphically illustrates the augmented probability model.

Notice that the likelihood $p(\dataMatrix|\latentMatrix)$
can be equivalently computed from the above augmented
model by marginalizing out $(\mappingFunctionMatrix, \inducingMatrix)$ and crucially this is
true for any value of the inducing inputs $\latentMatrix_u$. This means
that, unlike $\latentMatrix$, the inducing inputs $\latentMatrix_u$ are not random variables
and neither are they model hyperparameters; they are
variational parameters. This interpretation of the inducing
inputs is key in developing our approximation and it arises
from the variational approach of \cite{Titsias09}. Taking
advantage of this observation we now simplify our notation
by dropping $\latentMatrix_u$ from our expressions.

We can now apply variational inference to approximate the true posterior, 
$p(\mappingFunctionMatrix, \inducingMatrix,  \latentMatrix | \dataMatrix) = 
p(\mappingFunctionMatrix | \inducingMatrix, \dataMatrix, \latentMatrix)$ $p(\inducingMatrix | \dataMatrix, \latentMatrix) p(\latentMatrix|\dataMatrix)$ with
a variational distribution of the form, 
%
\begin{equation}
\label{varDistr}
q(\mappingFunctionMatrix, \inducingMatrix,\latentMatrix) = q(\mappingFunctionMatrix | \inducingMatrix, \latentMatrix) q(\inducingMatrix) q(\latentMatrix) =
 \left( \prod_{j=1}^\dataDim p(\mappingFunctionVector_{:, \outputIndex} | \inducingVector_{:, \outputIndex}, \latentMatrix )q(\inducingVector_{:, \outputIndex}) \right) q(\latentMatrix) .
\end{equation}
Moreover, the distribution $q(\latentMatrix)$ is constrained to be Gaussian,
\begin{equation}
  \label{qX}
  q(\latentMatrix) =  \gaussianDist{\latentMatrix}{\mathcal{M}}{\mathcal{S}},
\end{equation}
while $q(\inducingMatrix)$ is an arbitrary (i.e.\ unrestricted) variational distribution.
We can choose the Gaussian $q(\latentMatrix)$ to factorise across latent dimensions or datapoints and, as 
will be discussed in Section \ref{section:priors}, this choice will depend on the form of the prior distribution $p(\latentMatrix)$. 
For the time being, however, we shall proceed assuming a general form 
for this Gaussian. 

\par The particular choice for the variational distribution
allows us to analytically compute a lower bound. The key reason behind this 
is that the conditional GP prior term that appears in the joint density in \eqref{augmentedJoint} 
is also part of the variational distribution. Indeed, by making use of equations 
\eqref{augmentedJoint} and \eqref{varDistr} the derivation of the lower
bound has as follows,   
\begin{align}
\F & \left( q(\latentMatrix), q(\inducingMatrix) \right) = 
  \int q(\mappingFunctionMatrix,\inducingMatrix,\latentMatrix)
      \log \frac{ p(\dataMatrix,\mappingFunctionMatrix, \inducingMatrix,\latentMatrix)}{q(\mappingFunctionMatrix, \inducingMatrix,\latentMatrix)}  
          \intd  \latentMatrix \intd \mappingFunctionMatrix \intd \inducingMatrix
      \nonumber \\
 & = \int \prod_{j=1}^\dataDim 
    p(\mappingFunctionVector_{:, \outputIndex} | \inducingVector_{:, \outputIndex}, \latentMatrix )
    q(\inducingVector_{:, \outputIndex})
    q(\latentMatrix) 
      \log  \frac{\prod_{j=1}^\dataDim
              p(\dataVector_{:, \outputIndex} | \mappingFunctionVector_{:, \outputIndex}) 
              \cancel{p(\mappingFunctionVector_{:, \outputIndex} | \inducingVector_{:, \outputIndex}, \latentMatrix)}
              p(\inducingVector_{:, \outputIndex}) 
              p(\latentMatrix )}
            {\prod_{j=1}^\dataDim 
              \cancel{p(\mappingFunctionVector_{:, \outputIndex} | \inducingVector_{:, \outputIndex}, \latentMatrix )}
               q(\inducingVector_{:, \outputIndex})
               q(\latentMatrix)} 
       \intd  \latentMatrix \intd \mappingFunctionMatrix  \intd \inducingMatrix \nonumber \\
 & = \int \prod_{j=1}^\dataDim
    p(\mappingFunctionVector_{:, \outputIndex} | \inducingVector_{:, \outputIndex}, \latentMatrix )
    q(\inducingVector_{:, \outputIndex})
    q(\latentMatrix) 
    \log  \frac{\prod_{j=1}^\dataDim 
        p(\dataVector_{:, \outputIndex} | \mappingFunctionVector_{:, \outputIndex}) 
        p(\inducingVector_{:, \outputIndex})}
      {\prod_{j=1}^\dataDim 
        q(\inducingVector_{:, \outputIndex})
      }
    \intd  \latentMatrix \intd \mappingFunctionMatrix  \intd \inducingMatrix
-  \int  q(\latentMatrix)   
    \log \frac{q(\latentMatrix)}{p(\latentMatrix )}   \intd  \latentMatrix \nonumber \\
 & = \hat{\F}\left( q(\latentMatrix), q(\inducingMatrix) \right) - \KL{q(\latentMatrix)}{p(\latentMatrix)}, \label{jensensSplit}
\end{align}
with:
 \begin{align}
\hat{\F} \left( q(\latentMatrix), q(\inducingMatrix) \right) 
&= 
\sum_{j=1}^\dataDim \left( 
    \int q(\inducingVector_{:, \outputIndex}) q(\latentMatrix) 
    \left\langle \log p(\dataVector_{:, \outputIndex} | \mappingFunctionVector_{:, \outputIndex}) 
    \right\rangle_{p(\mappingFunctionVector_{:, \outputIndex} | \inducingVector_{:, \outputIndex}, \latentMatrix)} \intd \inducingVector_{:, \outputIndex} \intd \latentMatrix +
             \log \left\langle 
                \frac{p(\inducingVector_{:, \outputIndex})}{q(\inducingVector_{:, \outputIndex})} 
               \right\rangle_{q(\inducingVector_{:, \outputIndex})} 
  \right) \nonumber \\
  &= \sum_{j=1}^\dataDim \hat{\F}_j \left( q(\latentMatrix), q(\inducingVector_{:, \outputIndex}) \right) , \label{Fd}
\end{align} 
%
where $\langle \cdot \rangle$ is a shorthand for expectation. 
Clearly,  the second $\text{KL}$ term can be easily calculated since both 
$p(\latentMatrix)$ and $q(\latentMatrix)$ are Gaussians; explicit expressions are given in Section \ref{section:priors}.  
To compute $\hat{\F}_j \left( q(\latentMatrix), q(\inducingVector_{:, \outputIndex}) \right)$, first
note that (see Appendix \ref{appendix:bound} for details), 
\begin{equation}
\label{eq:expectYgivenF}
\left\langle \log p(\dataVector_{:, \outputIndex} | \mappingFunctionVector_{:, \outputIndex}) 
    \right\rangle_{p(\mappingFunctionVector_{:, \outputIndex} | \inducingVector_{:, \outputIndex}, \latentMatrix)} 
    =  \log \gaussianDist{\dataVector_{:, \outputIndex}}{\mathbf{a}_\outputIndex}{\dataStd^2 \eye_\dataDim} - 
    \frac{1}{2\dataStd^2} \tr{\Kff }  +  \frac{1}{2\dataStd^2} \tr{\Kuu^{-1} \Kuf\Kfu},
\end{equation}
where $\mathbf{a}_j$ is given by equation \eqref{eq:conditionalGPmeanCovar}, based on which we can write 
\begin{align}
\hat{\F}_j \left( q(\latentMatrix), q(\inducingVector_{:, \outputIndex}) \right)={}& 
  \int q(\inducingVector_{:, \outputIndex}) \log \frac{
      e^{\la \log \gaussianDist{\dataVector_{:, \outputIndex}}{\mathbf{a}_\outputIndex}{\dataStd^2 \eye_\dataDim} \ra_{q(\latentMatrix)}}
    p(\inducingVector_{:, \outputIndex})}{q(\inducingVector_{:, \outputIndex})} \intd \inducingVector_{:, \outputIndex} - \mathcal{A} , \label{boundFAnalytically5}
\end{align}
where  
$\mathcal{A}=\frac{1}{2\dataStd^2} \tr{\la \Kff \ra_{q(\latentMatrix)}} -
    \frac{1}{2\dataStd^2} \tr{ 
      \Kuu^{-1} \la \Kuf \Kfu \ra_{q(\latentMatrix)}} $.
The expression in \eqref{boundFAnalytically5} is a KL-like quantity and, therefore, $q(\inducingVector_{:, \outputIndex})$ is optimally set to be proportional to 
the numerator inside the logarithm of the above equation, i.e.\ 
\begin{equation}
\label{qu}
q(\inducingVector_{:, \outputIndex}) \propto e^{\la \log \gaussianDist{\dataVector_{:, \outputIndex}}{\mathbf{a}_\outputIndex}{ \dataStd^2 \eye_\dataDim} \ra_{q(\latentMatrix)}}
    p(\inducingVector_{:, \outputIndex}) ,
\end{equation}
which is just a Gaussian distribution (see Appendix \ref{appendix:bound} for an explicit form). 

\par We can now re-insert the optimal value for $q(\inducingVector_{:, \outputIndex})$ back into 
$\hat{\F}_j \left( q(\latentMatrix), q(\inducingVector_{:, \outputIndex}) \right)$, somehow {\em reversing Jensen's inequality} 
(this trick is also explained in \citep{King:klcorrection06}), to obtain:
\begin{equation}
\label{boundFAnalyticallyFinalIntegral}
\hat{\F}_j \left( q(\latentMatrix) \right) = 
  \log \int e^{\la \log \gaussianDist{\dataVector_{:, \outputIndex}}{\mathbf{a}_\outputIndex}{\dataStd^2 \eye_\dataDim} \ra_{q(\latentMatrix)}}
    p(\inducingVector_{:, \outputIndex}) \intd \inducingVector_{:, \outputIndex}  -\mathcal{A} .
\end{equation}

\noindent Notice that by optimally eliminating $q(\inducingVector_{:, \outputIndex})$ we obtain a tighter bound which no longer depends on
this distribution, \ie 
$\hat{\F}_j \left( q(\latentMatrix) \right) \geq 
  \hat{\F}_j \left( q(\latentMatrix), q(\inducingVector_{:, \outputIndex}) \right)$.
Also notice that the expectation appearing in equation \eqref{boundFAnalyticallyFinalIntegral}
is a standard Gaussian integral and \eqref{boundFAnalyticallyFinalIntegral} can
be calculated in closed form, which turns out to be (see Appendix \ref{appendix:bound3} for details):
\begin{equation}
\label{FdFinal}
\hat{\F}_j \left( q(\latentMatrix) \right) = \log \left[   
  \frac{\dataStd^{-\numData} \vert \mathit{\Kuu} \vert ^\frac{1}{2} }
     {(2\pi)^{\frac{\numData}{2}} \vert \dataStd^{-2} \bfPsi_2 + \Kuu  \vert ^\frac{1}{2} } 
   e^{-\frac{1}{2} \dataVector_{:, \outputIndex}^\top \bfW \dataVector_{:, \outputIndex}}
   \right]   -
   \frac{\psi_0}{2\dataStd^2} + \frac{1}{2\dataStd^2} 
   \tr{ \Kuu^{-1} \bfPsi_2}
\end{equation}

\noindent where
\begin{equation}
\label{psis}
\psi_0 = \tr{\langle \mathit{\Kff} \rangle_{q(\latentMatrix)}} \;, \;\;
\bfPsi_1 = \langle \Kfu \rangle_{q(\latentMatrix)} \;, \;\;
\bfPsi_2 = \langle \Kuf \mathit{\Kfu} \rangle_{q(\latentMatrix)}
\end{equation}
\noindent are referred to as $\Psi$ statistics 
and $\bfW = \dataStd^{-2} \eye_\numData - \dataStd^{-4} \bfPsi_1 (\dataStd^{-2} \bfPsi_2 + \Kuu)^{-1} \bfPsi_1^\top$. 

The computation of $\hat{\F}_j \left( q(\latentMatrix) \right)$ only requires us
to compute matrix inverses and determinants which involve $\Kuu$
instead of $\Kff$, something which
is tractable since $\Kuu$ does not depend on $\latentMatrix$. Therefore, this expression is straightforward to compute, as long as
the covariance function $\kernelScalar_f$  is selected so that the $\Psi$ quantities of
equation \eqref{psis} can be computed analytically. 

It is worth noticing that the $\Psi$ statistics are computed in a decomposable way 
since the covariance matrices appearing in them are evaluated in pairs of
inputs $\latentVector_{\dataIndex, :}$ and $(\latentVector_u)_{k, :}$ taken from $\latentMatrix$ and $\latentMatrix_u$ respectively.
 In particular, the statistics 
$\psi_0$ and $\bfPsi_2$ are written as sums of independent terms
where each term is associated with a data point and similarly 
each column of the matrix $\bfPsi_1$ is associated with only one data point.
This decomposition is useful when a new data vector 
is inserted into the model and can also help to speed up computations during
test time as discussed in Section \ref{section:predictions}. 
It can also allow for parallelization in the computations as suggested in \citep{Gal:Distributed14}.
%
Therefore, the averages of the covariance matrices
over $q(\latentMatrix)$ in equation \eqref{psis} of the $\Psi$ statistics
can be computed separately for each marginal
$q(\latentVector_{\dataIndex, :}) = \gaussianDist{\latentVector_{\dataIndex, :}}{\bfmu_{\dataIndex,:}}{\bfS_\dataIndex}$ taken from the full $q(\latentMatrix)$ of equation \eqref{qX}.
We can, thus, write that $\psi_0 = \sum_{\dataIndex=1}^\numData \psi_0^\dataIndex$
where
\begin{equation}
\psi_0^\dataIndex = \int \kernelScalar_\mappingFunction(\latentVector_{\dataIndex, :},\latentVector_{\dataIndex, :}) \gaussianDist{\latentVector_{\dataIndex, :}}{\bfmu_{\dataIndex,:}}{ \bfS_\dataIndex} \intd \latentVector_{\dataIndex, :}.
\label{eq:psi0}
\end{equation}
Further,
$\bfPsi_1$ is an $\numData \times \numInducing$ matrix such that  
\begin{equation} (\Psi_1)_{\dataIndex,k} = 
\int \kernelScalar_\mappingFunction(\latentVector_{\dataIndex, :},(\latentVector_u)_{k, :}) 
\gaussianDist{\latentVector_{\dataIndex, :}}{\bfmu_{\dataIndex,:}}
  { \bfS_\dataIndex} \intd \latentVector_{\dataIndex, :}, \label{eq:psi1}
\end{equation}
where $(\latentVector_u)_{k, :}$ denotes the $k$th row of $\latentMatrix_u$. Finally,
$\bfPsi_2$ is an $\numInducing \times \numInducing$ matrix which is written as
 $\bfPsi_2 = \sum_{\dataIndex=1}^\numData \Psi_2^\dataIndex$ where $\Psi_2^\dataIndex$ is such that 
\begin{equation}
  (\Psi^\dataIndex_2)_{k, k'} = \int \kernelScalar_\mappingFunction(\latentVector_{\dataIndex, :},(\latentVector_u)_{k, :})
  \kernelScalar_\mappingFunction((\latentVector_u)_{k',:},\latentVector_{\dataIndex, :}) \gaussianDist{\latentVector_{\dataIndex, :}}{\bfmu_{\dataIndex,:}}{ \bfS_\dataIndex} \intd \latentVector_{\dataIndex, :}.
\label{eq:psi2}
\end{equation}

Notice that these
statistics constitute convolutions of the covariance function $\kernelScalar_f$ with Gaussian densities and 
are tractable for many standard covariance functions, such as the ARD exponentiated quadratic or the linear one.
The analytic forms of the $\Psi$ statistics for the aforementioned covariance functions are
given in Appendix \ref{PsiQuantities}.



To summarize, the final form of the variational lower bound on the marginal
likelihood $p(\dataMatrix)$ is written as
\begin{equation}
\F \left( q(\latentMatrix) \right) = \hat{\F} \left( q(\latentMatrix) \right) - \KL{q(\latentMatrix)}{p(\latentMatrix)}, \label{jensensSplit2}
\end{equation}
where $\hat{\F} \left( q(\latentMatrix) \right)$  can be obtained by summing both sides of \eqref{FdFinal} over the
$\dataDim$ outputs,
\begin{equation}
\label{eq:hatF}
\hat{\F} \left( q(\latentMatrix) \right) = \sum_{j=1}^\dataDim \hat{\F}_j \left( q(\latentMatrix) \right).
\end{equation}
We note that the above framework is, in essence, computing the following
approximation analytically,
\begin{equation}
\label{approxF1}
\hat{\F} \left( q(\latentMatrix) \right) \le \int q(\latentMatrix) \log p(\dataMatrix|\latentMatrix) \intd \latentMatrix .
\end{equation}

The lower bound \eqref{jensensSplit} can be jointly maximized over the
model parameters $\bftheta$ and variational parameters 
$ \{ \mathcal{M}, \mathcal{S}, \latentMatrix_u \}  $ 
by applying a gradient-based optimization algorithm. This approach is
similar to the optimization of the MAP objective function employed in the 
standard GP-LVM \citep{Lawrence:pnpca05} with the main
difference being that instead of optimizing the random variables $\latentMatrix$, we 
now optimize a set of \textit{variational
parameters} which govern the approximate posterior mean and variance for $\latentMatrix$. Furthermore, the inducing inputs
$\latentMatrix_u$ are variational parameters and the optimisation over them simply improves the 
approximation similarly to variational sparse GP regression  \citep{Titsias09}. 

By investigating more carefully the resulting expression of the bound allows us to observe that 
each term $\hat{\F}_j \left( q(\latentMatrix) \right)$ from \eqref{FdFinal}, that depends on the single column of data ${\bf y}_{:,j}$, 
closely resembles the corresponding variational lower bound 
obtained by applying the method of \cite{Titsias:variational09} in standard sparse GP regression.
The difference in variational GP-LVM is that now
$\latentMatrix$ is marginalized out so that the terms containing $\latentMatrix$, \ie the kernel
quantities $\tr{\Kff}$, $\Kfu$ and $\Kfu \Kuf$, are transformed
into averages (\ie the $\Psi$ quantities in \eqref{psis}) with respect to the
variational distribution $q(\latentMatrix)$. 

Finally, notice that the application of the variational method developed in this paper is not restricted
to the set of latent points. As in \citep{Titsias:VarInferenceMahalanobis13}, a fully Bayesian approach
can be obtained by additionally placing priors on the kernel parameters and, subsequently,
integrating them out variationally with the methodology that we described in this section.

\subsection{\label{section:priors} Applying the Variational Framework to Different GP-LVM Variants}


Different variational GP-LVM algorithms  can be obtained by varying the form of the latent space prior $p(\latentMatrix)$
which so far has been left unspecified. 
One useful property of the variational lower bound is that 
$p(\latentMatrix)$ appears only in the separate KL divergence 
term, as can be seen by equation \eqref{jensensSplit}, which can be tractably computed when  $p(\latentMatrix)$  is Gaussian.
This allows our  framework to easily accommodate different Gaussian forms for the 
latent space prior which give rise to different GP-LVM variants. 
%
In particular, incorporating a specific prior 
mainly requires us to specify a suitable factorisation for $q(\latentMatrix)$
and compute the corresponding KL term. In contrast, the general structure of the more complicated $\hat{\F} \left( q(\latentMatrix) \right)$ term remains unaffected.
Next we demonstrate these ideas by giving further details about how to apply the variational method to the two GP-LVM variants discussed in Section 
\ref{section:gplvmDynamics}. For both cases we follow 
the recipe that the factorisation of the variational 
distribution $q(\latentMatrix)$ resembles the factorisation of the prior $p(\latentMatrix)$. 

\subsubsection{\label{staticPrior}The Standard Variational GP-LVM for I.i.d.\ Data}

In the simplest case, the latent space prior is just a standard
normal density, fully factorised across datapoints and latent dimensions,  as shown in \eqref{standardNormal}. 
This is the typical assumption in latent variable models, such as factor analysis and PPCA \citep{Bartholomew:book87,Basilevsky:book94,Tipping:probpca99}.  
%
%
We choose a variational distribution $q(\latentMatrix)$ that follows the factorisation of the prior,
\begin{equation}
  \label{qXstatic}
  q(\latentMatrix) = \prod_{\dataIndex=1}^\numData \gaussianDist{\latentVector_{\dataIndex, :}}{\bfmu_{\dataIndex, :}}{\bfS_\dataIndex},
\end{equation} 
where each covariance matrix $\bfS_\dataIndex$ is diagonal. Notice that this variational distribution depends on $2 n q$ free parameters. 
The corresponding $\text{KL}$ quantity appearing in \eqref{jensensSplit2} takes the explicit form 
\begin{equation}
\label{KLstatic}
\KL{q(\latentMatrix)}{p(\latentMatrix)} = 
 \frac{1}{2}\sum_{\dataIndex=1}^\numData \tr{\bfmu_{\dataIndex, :} \bfmu_{\dataIndex, :}^\T + \bfS_\dataIndex - \log \bfS_\dataIndex}  - \frac{n q}{2} ,
\end{equation}
where $\log \bfS_\dataIndex$ denotes the diagonal matrix resulting from $\bfS_\dataIndex$ by taking the logarithm of its diagonal elements. 
To train the model we simply need to substitute the above term in the final form of the variational lower in \eqref{jensensSplit2} and follow
the gradient-based optimisation procedure. 

\par The resulting variational GP-LVM can be seen as a non-linear version of Bayesian probabilistic PCA \citep{Bishop:bayesPCA98,Minka:automatic01}. 
In the experiments, we consider this model for non-linear dimensionality reduction and demonstrate its ability to automatically select the
latent dimensionality.

\subsubsection{\label{temporalPrior}  The Dynamical Variational GP-LVM for Sequence Data}

We now turn into the second model discussed in Section \ref{section:gplvmDynamics}, which is 
suitable for sequence data. Again we define a variational distribution $q(\latentMatrix)$ so that it resembles fully the 
factorisation of the prior, i.e.\  
%
\begin{equation}
\label{eq:qTemporal}
  q(\latentMatrix) = \prod_{\outputIndex=1}^\latentDim \gaussianDist{\latentVector_{:, \outputIndex}}{\bfmu_{:,\outputIndex}}{\bfS_\outputIndex},
\end{equation}
where $\bfS_\outputIndex$ is a $n \times n$ full covariance matrix.  
The corresponding $\text{KL}$ term takes the form 
\begin{equation}
\label{KLasSummationOverQ} 
\KL{q(\latentMatrix)}{p(\latentMatrix | \bft)} = \frac{1}{2} \sum_{j=1}^{\latentDim} \left[ 
    \tr{\Kx^{-1} \bfS_\outputIndex 
                   + \Kx^{-1} \boldsymbol \mu_{:,\outputIndex} \boldsymbol \mu_{:,\outputIndex}^\top}
    +\log \vert \Kx \vert
    - \log \vert \bfS_\outputIndex \vert
 \right] - \frac{\numData\latentDim}{2}.
\end{equation}



This term can be substituted into the final form of the variational lower bound in \eqref{jensensSplit2} and allow training 
using a gradient-based optimisation procedure.
If implemented naively, such a procedure, will require too many parameters to tune
since the variational distribution depends on $nq + \frac{n(n+1)}{2}q$ free parameters. However,  
by applying the reparametrisation trick suggested by \cite{OpperFixedPointCovariance}
we can reduce the number of parameters in the variational distribution 
to just $2 n q$. Specifically, the stationary conditions obtained by setting to zero the 
first derivatives of the variational bound w.r.t. $\bfS_\outputIndex$ and
$\bfmu_{:,\outputIndex}$ take the form,
\begin{equation}
\bfS_\outputIndex = \left( \Kx^{-1} + \bfLambda_j \right)^{-1} \;\;\; \text{and}  \;\;\;  \bfmu_{:,\outputIndex} = \Kx \bar{\bfmu}_{:,\outputIndex}, \label{SFixedPointQ}
\end{equation}
where 
\begin{equation}
\label{lambdaMubar}
\bfLambda_j = - 2\frac{\vartheta \hat{\F} \left( q(\latentMatrix) \right)}{\vartheta \mathit{\bfS_\outputIndex}}
\;\;\; \text{and}  \;\;\;
\bar{\bfmu}_{:,\outputIndex} = \frac{\vartheta \hat{\F} \left( q(\latentMatrix) \right) }{\vartheta \bfmu_{:,\outputIndex}}.
\end{equation}
Here, $\bfLambda_j$ is a $\numData \times \numData$ diagonal positive definite matrix and 
$\bar{\bfmu}_{:,\outputIndex}$ is a $\numData-$dimensional vector.
The above stationary conditions tell us that, since $\bfS_\outputIndex$ depends on a
diagonal matrix $\bfLambda_j$, we can reparametrise it using only the
diagonal elements of that matrix, denoted by the $\numData-$dimensional vector $\boldsymbol
\lambda_j$.  Then, we can optimise the $2 \latentDim \numData$ parameters 
$(\boldsymbol \lambda_j$, $\bar{\bfmu}_{:,\outputIndex})$ and obtain the original
parameters using the transformation in \eqref{SFixedPointQ}. 

There are two optimisation strategies, depending
on the way we choose to treat the newly introduced parameters $\bflambda_j$ and $\bar{\bfmu}_{:,\outputIndex}$.
Firstly, inspired by \cite{OpperFixedPointCovariance} we can construct an iterative optimisation scheme.
More precisely, 
the variational bound $\F$ in equation \eqref{jensensSplit2} depends on the {\em actual} variational parameters
$\bfmu_{:,\outputIndex}$ and $\bfS_\outputIndex$ of $q(\latentMatrix)$, which through equation \eqref{SFixedPointQ} depend on the newly introduced
quantities $\bar{\bfmu}_{:,\outputIndex}$ and $\bflambda_j$ which, in turn, are associated with $\F$
through equation \eqref{lambdaMubar}. These observations can lead to 
an EM-style algorithm which alternates
between estimating one of the parameter sets $\{\bftheta, \latentMatrix_u \}$ and $\{ \mathcal{M}, \mathcal{S} \}$ by keeping the other set fixed.
%
%
An alternative approach, which is the one we use in our implementation, is to treat the new parameters
$\bflambda_\outputIndex$ and $\bar{\bfmu}_{:,\outputIndex}$ as completely free ones so that equation \eqref{lambdaMubar} is never used. In this case, the
variational parameters are optimised directly with a gradient based optimiser, jointly with the model
hyperparameters and the inducing inputs.

Overall, the above reparameterisation is appealing not only because of improved
complexity, but also because of optimisation robustness. Indeed, equation \eqref{SFixedPointQ} 
confirms that the original variational parameters are coupled via $\Kx$, which is a full-rank covariance matrix. By reparametrising according to equation \eqref{SFixedPointQ} and treating
the new parameters as free ones, we manage to approximately break this coupling and apply our optimisation 
algorithm on a set of less correlated parameters.

Furthermore, the methodology described above can be readily applied to model dependencies of a different nature
(\eg spatial rather than temporal), as any kind of high dimensional input variable can replace
the temporal inputs of the graphical model in fig.\ \ref{fig:graphicalModels}\subref{fig:dbgplvm}.
Therefore, by simply
replacing the input $\bft$ with any other kind of observed input
$\bfZ$ we trivially obtain a Bayesian framework for
warped GP regression \citep{snelson:2004warped, lazaro:warped}
for which we can predict the latent function values in new inputs
$\bfZ_*$ through a non-linear, latent warping layer,
using exactly the same architecture and equations described in
this section and in Section \ref{predictionsDynamical}. Similarly, if
the observed inputs of the top layer are taken to be the outputs themselves,
then we obtain a probabilistic auto-encoder (\eg \cite{kingma2013auto}) 
which is non-parametric and based on Gaussian processes.


Finally, the above dynamical variational GP-LVM algorithm can be easily extended to deal with datasets consisting of multiple independent sequences (probably of different length) such as those arising in human motion capture applications. Let, for example, the dataset be a group of
$\numSequences$ independent sequences  $\left( \dataMatrix^{(1)}, ..., \dataMatrix^{(\numSequences)} \right)$. We would like the dynamical version of our
model to capture the underlying
commonality of these data. We handle this by allowing a different \emph{temporal} latent function for each of the independent
sequences, so that $\latentMatrix^{(i)}$ is the set of latent variables corresponding to the sequence $i$.
%
These sets are a priori assumed to be independent since they correspond to separate sequences,
i.e.\ $p\left( \latentMatrix^{(1)}, \latentMatrix^{(2)}, ..., \latentMatrix^{(\numSequences)} \right) = 
\prod_{i=1}^\numSequences p(\latentMatrix^{(i)})$.
%
This factorisation leads to a block-diagonal structure for the time covariance matrix $\Kx$, where each block corresponds to one sequence.
 In this setting, each block of observations $\dataMatrix^{(i)}$ is generated from its corresponding $\latentMatrix^{(i)}$
according to $\dataMatrix^{(i)} = \mappingFunctionMatrix^{(i)} + \noiseVector$, where the latent function which governs this mapping is shared across all sequences and 
$\noiseVector$ is Gaussian noise.

\subsection{Time Complexity and Handling Very High Dimensional Datasets\label{sec:HighDimensionalData}}
Our variational framework makes use of inducing point representations which 
provide low-rank approximations to the covariance $\Kff$. For the standard variational
GP-LVM, this allows us to avoid the typical cubic complexity of Gaussian processes,
reducing the computational cost to $O(\numData \numInducing^2)$. 
Since we typically select a small set of inducing points, $\numInducing \ll \numData$,
the variational GP-LVM can handle relatively large training sets (thousands of points, $\numData$).
The \emph{dynamical} variational GP-LVM, however, still requires the inversion of
 the covariance matrix $\kernelMatrix_\latentScalar$ of size $\numData \times \numData$, 
as can be seen in equation \eqref{SFixedPointQ}, thereby inducing a computational
cost of $O(\numData^3)$. 
Further, the models scale only linearly with the
number of dimensions $\dataDim$. Specifically, the number of dimensions only
matters when performing calculations involving the data matrix $\dataMatrix$. In
the final form of the lower bound (and consequently in all of the
derived quantities, such as gradients) this matrix only appears in the
form $\dataMatrix \dataMatrix^\top$ which can be precomputed. This means that, when $\numData \ll
\dataDim$, we can calculate $\dataMatrix \dataMatrix^\top$ only once and then substitute $\dataMatrix$ with
the SVD (or Cholesky decomposition) of $\dataMatrix \dataMatrix^\top$. In this way, we can
work with an $\numData \times \numData$ instead of an $\numData \times \dataDim$
matrix. Practically speaking, this allows us to work with data sets
involving millions of features. In our experiments we model directly
the pixels of HD quality video, exploiting this trick.


%% file: predictions.tex

\section{Predictions with the Variational GP-LVM \label{section:predictions}} 

In this section, we explain how the proposed Bayesian models can
accomplish various kinds of prediction tasks. We will use a star ($*$) to denote test quantities, 
\eg a test data matrix will be denoted by $\dataMatrix_* \in \Re^{\numData_* \times \dataDim}$ while test row
and column vectors of such a matrix will be denoted by $\dataVector_{i,*}$ and 
$\dataVector_{*,j}$. 

The first type of
inference we are interested in is the calculation of the probability density
$p(\dataMatrix_* | \dataMatrix)$. The computation of this quantity can allow us to use the
model as a density estimator which, for instance, can represent the
class conditional distribution in a generative based classification
system.  We will exploit such a use in Section
\ref{sec:classification}.  Secondly, we discuss how from a test data matrix $\dataMatrix_* = (\dataMatrix_*^{\unobservedSet}, \dataMatrix_*^{\observedSet})$,
we can probabilistically reconstruct the unobserved part $\dataMatrix_*^{\unobservedSet}$ 
based on the observed part $\dataMatrix_*^{\observedSet}$ and where $\unobservedSet$ and $\observedSet$ denote non-overlapping 
sets of indices such that their union is $\{1,\ldots,\dataDim\}$.  For this second problem the missing dimensions are reconstructed by
approximating the mean and the covariance of the  Bayesian predictive density $p(\dataMatrix_*^{\unobservedSet}|
\dataMatrix_*^{\observedSet}, \dataMatrix)$.  

Section \ref{sec:predStandardGPLVM} discusses how to solve the above tasks in the standard variational GP-LVM case while
Section \ref{predictionsDynamical} discusses the dynamical case. Furthermore, for the dynamical case the test points $\dataMatrix_*$ are accompanied
by their corresponding timestamps $\bft_*$ based on which  we can perform an additional {\em forecasting} prediction 
task, where we are given only a test time vector $\bft_*$ and we wish to predict the corresponding outputs.


\subsection{Predictions with the Standard Variational GP-LVM \label{sec:predStandardGPLVM}}


We first discuss how to approximate the density $p(\dataMatrix_*|\dataMatrix)$. 
By introducing the latent variables $\latentMatrix$ (corresponding to
the training outputs $\dataMatrix$) and the new test latent variables
$\latentMatrix_* \in \Re^{\numData_* \times \latentDim}$, we can write
the density of interest as the ratio of two marginal likelihoods,

\begin{equation}
  \label{pyystar1}
  p(\dataMatrix_*|\dataMatrix) = \frac{p(\dataMatrix_*,\dataMatrix)}{p(\dataMatrix)} = 
  \frac{\int p(\dataMatrix_*, \dataMatrix | \latentMatrix, \latentMatrix_*) p(\latentMatrix,\latentMatrix_*) 
    \intd \latentMatrix \intd \latentMatrix_*}{\int p(\dataMatrix|\latentMatrix) p(\latentMatrix) \intd \latentMatrix}.
\end{equation}
In the denominator we have the marginal likelihood of the GP-LVM for
which we have already computed a variational lower bound.  The
numerator is another marginal likelihood that is obtained by
augmenting the training data $\dataMatrix$ with the test points
$\dataMatrix_*$ and integrating out both $\latentMatrix$ and the newly
inserted latent variable $\latentMatrix_*$. In the following, we
explain in more detail how to approximate the density $p(\dataMatrix_*
|\dataMatrix)$ of equation \eqref{pyystar1} through constructing a ratio
of lower bounds.

The quantity $ \int p(\dataMatrix | \latentMatrix) p(\latentMatrix)
\intd \latentMatrix$ appearing in the denominator of equation
\eqref{pyystar1} is approximated by the lower bound
$e^{\F(q(\latentMatrix))}$ where
$\F(q(\latentMatrix))$ is the variational lower bound as
computed in Section \ref{sec:lowerBoundWithAuxVars} and is given in equation
\eqref{jensensSplit2}. The maximization of this lower bound specifies
the variational distribution $q(\latentMatrix)$ over the latent
variables in the training data. Then, this distribution remains fixed
during test time.  The quantity $\int p(\dataMatrix_*,\dataMatrix |
\latentMatrix, \latentMatrix_*) p(\latentMatrix,\latentMatrix_*) \intd
\latentMatrix \intd \latentMatrix_*$ appearing in the numerator of
equation \eqref{pyystar1} is approximated by the lower bound
$e^{\F(q(\latentMatrix,\latentMatrix_*))}$
which has exactly analogous form to \eqref{jensensSplit2}.  This
optimisation is fast, because the factorisation imposed for the
variational distribution in equation \eqref{qXstatic} means that
$q(\latentMatrix, \latentMatrix_*)$ is also a fully factorised
distribution so that
we can write $q(\latentMatrix,\latentMatrix_*) = q(\latentMatrix)
q(\latentMatrix_*)$. Then, if $q(\latentMatrix)$ is held fixed\footnote{
Ideally $q(\latentMatrix)$ would be optimised during test time as well.} 
during test time, we only need to optimise with respect to the $2
\numData_* \latentDim$ parameters of the variational Gaussian
distribution $q(\latentMatrix_*)=\prod_{i=1}^{\numData_*}
q(\latentVector_{i, *}) = \prod_{i=1}^{\numData_*}
\mathcal{N}(\bfmu_{i,*},\bfS_{i,*})$ (where $\bfS_{i,*}$ is a
diagonal matrix).  Further, since the $\Psi$ statistics decompose
across data, during test time we can re-use the already estimated
$\Psi$ statistics corresponding to the averages over
$q(\latentMatrix)$ and only need to compute the extra average terms
associated with $q(\latentMatrix_*)$.  Note that optimization of the
parameters $(\bfmu_{i,*}, \bfS_{i,*})$ of $q(\latentVector_{i,*})$ are
subject to local minima. However, sensible initializations of
$\bfmu_*$ can be employed based on the mean of the variational
distributions associated with the nearest neighbours of each test
point $\dataVector_{i,*}$ in the training data $\dataMatrix$.
%
%
%
Given the above, the approximation of $p(\dataMatrix_* | \dataMatrix)$
is given by rewriting equation \eqref{pyystar1} as,
\begin{equation}
p(\dataMatrix_* |\dataMatrix) \approx e^{ \F(q(\latentMatrix,\latentMatrix_*)) - \F(q(\latentMatrix))  }. 
\label{eq:predictive1}
\end{equation}

We now discuss the second prediction problem where a set of partially 
observed test points $\dataMatrix_* = (\dataMatrix_*^\unobservedSet,\dataMatrix_*^\observedSet)$ are given and 
we wish to reconstruct the missing part $\dataMatrix_*^\unobservedSet$.
The predictive density is, thus, $p(\dataMatrix_*^\unobservedSet | \dataMatrix_*^\observedSet, \dataMatrix)$.
Notice that $\dataMatrix_*^\unobservedSet$ is totally unobserved and, therefore, we cannot apply the
methodology described previously. Instead, our objective now is
to just approximate the moments of the predictive density. 
To achieve this, we will first need to introduce the underlying latent function
values $\mappingFunctionMatrix_*^{\unobservedSet}$ (the noise-free version of $\dataMatrix_*^{\unobservedSet}$)
and the latent variables $\latentMatrix_*$ so that we can decompose the exact predictive density as follows,
\begin{equation}
\label{eq:predictive2}
p(\dataMatrix_*^\unobservedSet | \dataMatrix_*^\observedSet, \dataMatrix) =  \int p(\dataMatrix_*^\unobservedSet | \mappingFunctionMatrix_*^\unobservedSet)  
p(\mappingFunctionMatrix_*^\unobservedSet | \latentMatrix_*, \dataMatrix_*^\observedSet, \dataMatrix) p(\latentMatrix_*| \dataMatrix_*^\observedSet, \dataMatrix) \intd  \mappingFunctionMatrix_*^\unobservedSet \intd  \latentMatrix_* .
\end{equation}
Then, we can introduce the approximation coming from the variational distribution so that 
\begin{align}
p(\dataMatrix_*^\unobservedSet | \dataMatrix_*^\observedSet, \dataMatrix) & \approx q(\dataMatrix_*^\unobservedSet|\dataMatrix_*^\observedSet,\dataMatrix) = 
    \int p(\dataMatrix_*^\unobservedSet | \mappingFunctionMatrix_*^\unobservedSet) q(\mappingFunctionMatrix_*^\unobservedSet |\latentMatrix_*) q(\latentMatrix_*)  
    	\intd \mappingFunctionMatrix_*^\unobservedSet  \intd \latentMatrix_* \label{eq:predictive3a},
\end{align}
based on which we wish to predict $\dataMatrix_*^{\unobservedSet}$ by estimating its mean $\mathbb{E}(\dataMatrix_*^{\unobservedSet})$ and covariance 
$\text{Cov}(\dataMatrix_*^{\unobservedSet})$. 
This problem takes the form of GP prediction with uncertain inputs similar 
to \citep{Oakley:computer02,quinonero2003propagation,Girard:uncertain01}, where the distribution $q(\latentMatrix_*)$ expresses the uncertainty over these inputs. 
The first term of the above integral comes from the Gaussian likelihood so $\dataMatrix_*^\unobservedSet$ is 
just a noisy version of
$\mappingFunctionMatrix_*^\unobservedSet$, as shown in equation \eqref{eq:yGivenF}. The remaining two terms together   $q(\mappingFunctionMatrix_*^\unobservedSet| \latentMatrix_*) q(\latentMatrix_*)$ 
are obtained by applying the variational methodology  
in order to optimise a variational lower bound on the following log marginal likelihood: 
\begin{align}
\log p(\dataMatrix_*^\observedSet, \dataMatrix) ={}&  \log \int p(\dataMatrix_*^\observedSet, \dataMatrix|\latentMatrix_*, \latentMatrix) 
p(\latentMatrix_*, \latentMatrix) \intd  \latentMatrix_* \intd  \latentMatrix \nonumber \\
={}&  \log \int p(\dataMatrix^\unobservedSet | \latentMatrix) p(\dataMatrix_*^\observedSet, \dataMatrix^\observedSet|\latentMatrix_*, \latentMatrix) p(\latentMatrix_*, \latentMatrix) 
\intd  \latentMatrix_* \intd  \latentMatrix,  \label{eq:marginalPredictions2}
\end{align}
which is associated with the total set of observations $(\dataMatrix_*^\observedSet, \dataMatrix)$. By following exactly Section 
\ref{section:vgplvm}, we can construct and optimise a lower bound $\F(q(\latentMatrix,\latentMatrix_*))$ on the above quantity, 
which along the way it allows us to compute a Gaussian variational distribution $q(\mappingFunctionMatrix, \mappingFunctionMatrix_*^\unobservedSet,\latentMatrix,\latentMatrix_*)$ 
from which $q(\mappingFunctionMatrix_*^\unobservedSet|\latentMatrix_*) q(\latentMatrix_*)$ is just a marginal. 
Further details about the form of the variational lower bound and how 
$q(\mappingFunctionMatrix_*^\unobservedSet|\latentMatrix_*)$ is computed are given in the Appendix \ref{app:partialtest}.  
In fact, the explicit form of $q(\mappingFunctionMatrix_*^\unobservedSet|\latentMatrix_*)$
takes the form of the projected process predictive distribution from sparse GPs \citep{Csato:sparse02,Smola:sparsegp00,Seeger:fast03,Rasmussen:book06}:
\begin{equation}
\label{eq:testPosteriorF}
q(\mappingFunctionMatrix_*^\unobservedSet | \latentMatrix_*) = \mathcal{N} \left( \mappingFunctionMatrix_*^\unobservedSet | \Kstaru \bfB, 
 \bfK_{* *} - 
 \Kstaru \left[ \Kuu^{-1}  - (\Kuu + \dataStd^{-2} \bfPsi_2)^{-1} \right] 
 \Kstaru^\top \right) ,
\end{equation}
where $\bfB = \dataStd^{-2} \left( \Kuu + \dataStd^{-2} \bfPsi_2 \right)^{-1} \bfPsi_1^\top \dataMatrix$,
 $\kernelMatrix_{* *} = \kernelScalar_\mappingFunction(\latentMatrix_*, \latentMatrix_*)$ and $\Kstaru = \kernelScalar_\mappingFunction(\latentMatrix_*, \latentMatrix_u)$.
By substituting now the above Gaussian $q(\mappingFunctionMatrix_*^\unobservedSet|\latentMatrix_*)$ in equation (\ref{eq:predictive3a}) and using the fact that 
$q(\latentMatrix_*)$  is also a Gaussian, we can analytically compute the  mean and covariance  of the 
predictive density which, based on the results of \cite{Girard:uncertain01}, take the form
\begin{align}
 \mathbb{E}(\mappingFunctionMatrix_*^\unobservedSet) ={}&  \bfB^\top \bfPsi_1^* \label{meanFstar} \\
 \text{Cov}(\mappingFunctionMatrix_*^\unobservedSet) ={}& \bfB^\top \left( \bfPsi_2^* - \bfPsi_1^* (\bfPsi_1^*)^\top \right) \bfB + \psi_0^* \eye - \text{tr} \left( \left( \Kuu^{-1} - \left( \Kuu + \dataStd^{-2} \bfPsi_2 \right)^{-1} \right) \bfPsi_2^* \right) \eye , \label{covFstar}
\end{align}
where $\psi_0^* = \tr{\la \kernelMatrix_{**} \ra}$, $\bfPsi_1^* = \la \Kustar \ra$
and $\bfPsi_2^* = \la \Kustar \Kustar^\top \ra$. All expectations are taken
w.r.t. $q(\latentMatrix_*)$ and can be calculated analytically for several kernel functions as explained in Section
\ref{sec:lowerBoundWithAuxVars} and Appendix \ref{PsiQuantities}.
Using the above expressions and the Gaussian noise model of equation \eqref{eq:yGivenF}, 
the predicted mean of $\dataMatrix_*^\unobservedSet$
is equal to $\mathbb{E}\left[ \mappingFunctionMatrix_*^\unobservedSet \right]$
and the predicted covariance is equal to 
$\text{Cov}(\mappingFunctionMatrix_*^\unobservedSet) + \dataStd^{2} \eye_{\numData_*}$.


\subsection{\label{predictionsDynamical} Predictions in the Dynamical Model}

The two prediction tasks described in the previous section for the standard variational GP-LVM can 
also be solved for the dynamical variant in a very similar fashion. Specifically, 
 the two predictive approximate densities
take exactly the same form as those in equations \eqref{eq:predictive1} and \eqref{eq:predictive3a}
while again the whole approximation relies on the maximisation of a variational lower bound
$\mathcal{F}(q(\latentMatrix, \latentMatrix_*))$. However, in the dynamical case where
the inputs $(\latentMatrix, \latentMatrix_*)$ are a priori correlated, the variational distribution 
$q(\latentMatrix, \latentMatrix_*)$ does not factorise across $\latentMatrix$ and $\latentMatrix_*$. 
This makes the optimisation of this distribution computationally more challenging,
as it has to be optimised with respect to its all 
$2(\numData+\numData_*)\latentDim$ parameters. 
This issue is further explained in Appendix \ref{app:partialtestBound}.

Finally, we shall discuss how to solve the forecasting problem with our dynamical model.
This problem is similar to the second predictive task described in 
Section \ref{sec:predStandardGPLVM}, but now the observed set is empty.
We can therefore write the predictive density similarly to equation \eqref{eq:predictive3a}
as follows, 
\begin{align}
p(\dataMatrix_* | \dataMatrix) & \approx \int p(\dataMatrix_* | \mappingFunctionMatrix_*)  
q(\mappingFunctionMatrix_* |\latentMatrix_*) q(\latentMatrix_*)  \intd \latentMatrix_* \intd \mappingFunctionMatrix_*  \label{eq:predictiveDyn3} .
\end{align}
The inference procedure then follows exactly as before, by making
use of equations \eqref{eq:predictive3a}, \eqref{meanFstar} and \eqref{covFstar}.
The only difference is that the computation of $q(\latentMatrix_*)$ (associated with a fully unobserved  $\dataMatrix_*$)
is obtained from standard GP prediction and does not require optimisation, i.e.,\ 
 \begin{align}
  q(\latentMatrix_*) = 
\int p(\latentMatrix_*|\latentMatrix) q(\latentMatrix) \intd \latentMatrix =  \prod_{j=1}^\latentDim   \int  p(\latentVector_{*,j} | \latentVector_{:, j}) q(\latentVector_{:, j}) \intd \latentVector_{:, j},
 \label{qxstar}
 \end{align}
 where $p(\latentVector_{*,j} | \latentVector_{:, j})$ is a Gaussian found from the conditional GP prior
 (see \cite{Rasmussen:book06}). Since $q(\latentMatrix)$ is  Gaussian, the above is also a Gaussian with 
 mean and variance given by,
 \begin{align}
  \bfmu_{\latentScalar_{*,j}} = {}& \kernelMatrix_{*\numData} \bar{\bfmu}_{:,j} \\
   \text{var}(\bfx_{*,j}) = {}& \kernelMatrix_{**} - \kernelMatrix_{*\numData} (\Kx + \bfLambda_j^{-1})^{-1} \kernelMatrix_{\numData*},
 \end{align}
 where $\kernelMatrix_{*\numData} = \kernelScalar_\latentScalar(\bft_*, \bft)$, $\kernelMatrix_{*\numData} = \kernelMatrix_{*\numData}^\top$ and 
 $\kernelMatrix_{**} = \kernelScalar_\latentScalar(\bft_*, \bft_*)$. Notice that these equations have
 exactly the same form as found in standard GP regression problems.

%% file: matlab.tex
\begin{matlab}


clear all
close all

retrainModels = 0;

rePredict = 0;


save 'opts.mat' 'retrainModels' 'rePredict'
system('mkdir ../diagrams');

clear; load opts
fprintf(1,'\n\n#-----  OCEAN DEMO ----#\n');
dataSetName = 'ocean';
experimentNo=60;
indPoints=-1; latentDim=20;
fixedBetaIters=200; reconstrIters = 2000;
itNo=[1000 2000 5000 8000 4000];
dynamicKern={'rbf','white','bias'};
whiteVar = 0.1;  vardistCovarsMult=1.7;
dataSetSplit = 'randomBlocks';
if retrainModels
    demHighDimVargplvm3
elseif rePredict
    demHighDimVargplvmTrained
else
    load demOceanVargplvm60Pred
    demHighDimVargplvmLoadPred
end
fr=reshape(Varmu(27,:),height,width); imagesc(fr); colormap('gray'); 
print -depsc oceanGpdsframe27.eps; system('epstopdf oceanGpdsframe27.eps');
fr=reshape(NNmuPartBest(27,:),height,width); imagesc(fr); colormap('gray'); 
print -depsc oceanNNframe27.eps; system('epstopdf oceanNNframe27.eps');

clear; load opts
fprintf(1,'\n\n#-----  MISSA DEMO ----#\n');
experimentNo = 59;
dataSetName = 'missa';
indPoints = -1; latentDim=25;
fixedBetaIters=50; reconstrIters = 4000;
itNo=[1000 2000 5000 8000 2000]; 
dynamicKern={'matern32','white','bias'};
vardistCovarsMult=1.6;
dataSetSplit = 'blocks';
blockSize = 4; whiteVar = 0.1;
msk = [48 63 78 86 96 111 118];
if retrainModels
    demHighDimVargplvm3
elseif rePredict
    demHighDimVargplvmTrained
else
    load demMissaVargplvm59Pred
    demHighDimVargplvmLoadPred
end

fr=reshape(Varmu(46,:),height,width); imagesc(fr); colormap('gray'); 
print -depsc missaGpdsframe46.eps; system('epstopdf missaGpdsframe46.eps');
fr=reshape(YtsOriginal(46,:),height,width); imagesc(fr); colormap('gray'); 
print -depsc missaYtsOrigframe46.eps; system('epstopdf missaYtsOrigframe46.eps');
fr=reshape(NNmuPartBest(46,:),height,width); imagesc(fr); colormap('gray'); 
print -depsc missaNNframe46.eps; system('epstopdf missaNNframe46.eps');
fr=reshape(Yts(17,:),height,width); imagesc(fr); colormap('gray'); 
print -depsc missaGpdsPredFrame17_part1.eps; system('epstopdf missaGpdsPredFrame17_part1.eps');
fr=reshape(Varmu(17,:),height,width); imagesc(fr); colormap('gray');
print -depsc missaGpdsPredFrame17_part2.eps; system('epstopdf missaGpdsPredFrame17_part2.eps');


clear; load opts
fprintf(1,'\n\n#-----  DOG DEMO: Generation ----#\n');
dataSetName = 'dog';
experimentNo=61;
indPoints=-1; latentDim=35;
fixedBetaIters=400;
reconstrIters = 1; 
itNo=[1000 1000 1000 1000 1000 1000 500 500 500 500 1000 1000 1000 1000 1000 1000 1000 500 500]; 
periodicPeriod = 4.3983; 
dynamicKern={'rbfperiodic','whitefixed','bias','rbf'};
vardistCovarsMult=0.8;
whiteVar = 1e-6;
dataToKeep = 60; dataSetSplit = 'custom';
indTr = [1:60];
indTs = 60; 
learnSecondVariance = 0;
if retrainModels
    demHighDimVargplvm3
end

clear; load opts; close all 
dataSetName = 'dog'; 
experimentNo=61; dataToKeep = 60; dataSetSplit = 'custom';
indTr = [1:60]; indTs = 60;
futurePred = 40; doSampling = 0; demHighDimVargplvmTrained

bar(prunedModelInit.kern.comp{1}.inputScales)
print -depsc dog_scalesInit.eps; system('epstopdf dog_scalesInit.eps');
bar(model.kern.comp{1}.inputScales)
print -depsc dog_scalesOpt.eps; system('epstopdf dog_scalesOpt.eps');

fr=reshape(Ytr(end,:),height,width); imagesc(fr); colormap('gray'); 
print -depsc dogGeneration_lastOfTraining.eps; system('epstopdf dogGeneration_lastOfTraining.eps');
fr=reshape(Varmu2(1,:),height,width); imagesc(fr); colormap('gray');  
print -depsc dogGeneration_firstOfTest.eps; system('epstopdf dogGeneration_firstOfTest.eps');
fr=reshape(Varmu2(13,:),height,width); imagesc(fr); colormap('gray'); 
print -depsc dogGeneration_frame14.eps; system('epstopdf dogGeneration_frame14.eps');



clear; load opts
fprintf(1,'\n\n#-----  DOG DEMO: Reconstruction ----#\n');
dataSetName = 'dog';
experimentNo=65;
indPoints=-1; latentDim=35;
fixedBetaIters=400;
reconstrIters = 2;
itNo=[1000 1000 1000 1000 1000 1000 500 500 500 500 1000 1000 1000 1000 1000 1000 1000 500 500]; 
periodicPeriod = 2.8840;
dynamicKern={'rbfperiodic','whitefixed','bias','rbf'};
vardistCovarsMult=0.8;
whiteVar = 1e-6;
dataSetSplit = 'custom';
indTr = 1:54;
indTs = 55:61;
learnSecondVariance = 0;
if retrainModels
    demHighDimVargplvm3
end


clear; load opts
dataSetName = 'dog';
experimentNo=65;
dataSetSplit = 'custom';
indTr = 1:54; indTs = 55:61;
predWithMs = 1; 
reconstrIters = 18000; 
doSampling = 0;
if rePredict
    demHighDimVargplvmTrained
else
    load demDogVargplvm65Pred
    demHighDimVargplvmLoadPred
end
fr=reshape(Varmu(5,:),height,width); imagesc(fr); colormap('gray'); 
print -depsc supplDogPredGpds5.eps; system('epstopdf supplDogPredGpds5.eps');
fr=reshape(Yts(5,:),height,width); imagesc(fr); colormap('gray'); 
print -depsc supplDogPredYts5.eps; system('epstopdf supplDogPredYts5.eps');
fr=reshape(Varmu(6,:),height,width); imagesc(fr); colormap('gray'); 
print -depsc supplDogPredGpds6.eps; system('epstopdf supplDogPredGpds6.eps');
fr=reshape(Yts(6,:),height,width); imagesc(fr); colormap('gray'); 
print -depsc supplDogPredYts6.eps; system('epstopdf supplDogPredYts6.eps');


clear; load opts
fprintf(1,'\n\n#-----  CMU DEMO: Rbf ----#\n');
experimentNo=34; 
itNo = [300 300 400 200 200 300 400 400];
dynamicKern = {'rbf', 'white', 'bias'};
vardistCovarsMult = 0.152;
if retrainModels 
    if rePredict
        doReconstr = 1;
    else
        doReconstr=0;
    end
    demCmu35gplvmVargplvm3;
elseif rePredict
    demCmu35vargplvmReconstructTaylor
end
predictPart = 'Legs';  plotRange = [];
demCmu35VargplvmPlotsScaled
fprintf(1,'# VGPDS RBF error on Legs reconstr:');
errStruct

predictPart = 'Body';  plotRange = [];
demCmu35VargplvmPlotsScaled
fprintf(1,'# VGPDS RBF error on Body reconstr:');
errStruct

bar(model.kern.comp{1}.inputScales);
print -depsc supplMocapScalesRbf.eps; system('epstopdf supplMocapScalesRbf.eps');

clear; load opts
fprintf(1,'\n\n#-----  CMU DEMO: Matern32 ----#\n');
experimentNo=33; 
itNo = [300 300 400 200 200 300 400 400];
dynamicKern = {'matern32', 'white', 'bias'};
vardistCovarsMult = 0.24;
if retrainModels 
    if rePredict
        doReconstr = 1;
    else
        doReconstr=0;
    end
    demCmu35gplvmVargplvm3;
elseif rePredict
    demCmu35vargplvmReconstructTaylor
end
predictPart = 'Legs'; plotRange = 10;
demCmu35VargplvmPlotsScaled
print -depsc supplMocapLeg5GpdsMatern.eps; system('epstopdf supplMocapLeg5GpdsMatern.eps');
fprintf(1,'# VGPDS Matern error on Legs reconstr:');
errStruct

predictPart = 'Body'; plotRange = 28;
demCmu35VargplvmPlotsScaled
print -depsc supplMocapBody28GpdsMatern.eps; system('epstopdf supplMocapBody28GpdsMatern.eps');
fprintf(1,'# VGPDS Matern error on Body reconstr:');
errStruct
close all
bar(model.kern.comp{1}.inputScales);
print -depsc supplMocapScalesMatern.eps; system('epstopdf supplMocapScalesMatern.eps');

fprintf(1,'\n\n#---- FINISHED reproducing plots and results!! \n');
delete opts.mat

a = ver('octave');
if length(a) == 0
  a = ver('matlab');
end
fid = fopen('vers.tex', 'w');
fprintf(fid, [a.Name ' version ' a.Version]);
fclose(fid);

fid = fopen('computer.tex', 'w');
fprintf(fid, ['\\verb+' computer '+']);
fclose(fid);

fid = fopen('date.tex', 'w');
fprintf(fid, datestr(now, 'dd/mm/yyyy'));
fclose(fid);

\end{matlab}

%% file: experiments.tex

\section{\label{section:experiments} Demonstration of the Variational Framework}

In this section we investigate the performance of the variational
GP-LVM and its dynamical extension. The
variational GP-LVM allows us to handle very high dimensional data and,
using ARD, to determine the undelying low dimensional subspace size
automatically. The generative construction allows us to impute missing
values when presented with only a partial observation.

We evaluate the models' performance in a variety of tasks, namely
visualisation, prediction, reconstruction, generation of data or timeseries
and class-conditional density estimation.
Matlab source code for repeating the following experiments is available
on-line from: \\
\url{https://github.com/SheffieldML/vargplvm} \\
 and supplementary videos from: \\
\url{http://htmlpreview.github.io/?https://github.com/SheffieldML/vargplvm/blob/master/vargplvm/html/index.html#vgpds}.

The experiments section is structured as follows; 
in Section \ref{covarianceFunctions} we outline the covariance
functions used for the experiments.
In Section \ref{sec:experimentsVisualisation} we demonstrate our
method in a standard visualisation benchmark. In Section \ref{sec:experimentsMocap}
we test both, the standard and dynamical variant of our method in
a real-world motion capture dataset. In Section \ref{sec:experimentsVideo}
we illustrate how our proposed model is able to handle a very large
number of dimensions by working directly with the raw pixel values of
high resolution videos. Additionally, we show how the dynamical model
can interpolate but also extrapolate in certain scenarios.
In Section \ref{sec:classification} we consider a classification task
on a standard benchmark, exploiting the fact that our framework gives
access to the model evidence, thus enabling Bayesian classification.

\subsection{Covariance Functions\label{covarianceFunctions}}

Before proceeding to the actual evaluation of our method, we first
review and give the forms of the covariance functions that will be
used for our experiments.  The mapping between the input and output
spaces $\latentMatrix$ and $\dataMatrix$ is nonlinear and, thus, we
use the covariance function of equation \eqref{ard} which also allows
simultaneous model selection within our framework.  In experiments
where we use our method to also model dynamics, apart from the
infinitely differentiable exponantiated quadratic covariance function
defined in equation \eqref{rbf}, we will also consider for the
dynamical component the Mat\'ern $3/2$ covariance function which is
only once differentiable, and a periodic one \citep{Rasmussen:book06,
  MacKay98} which can be used when data exhibit strong
periodicity. These covariance functions take the form:
\begin{eqnarray}
  \kernelScalar_{\latentScalar(\text{mat})} \left( t_i, t_j \right)& = &  
  \sigma_{\text{mat}}^2 \left( 1 + \frac{\sqrt{3} |t_i - t_j|}{\lengthScale} \right)
  \exp \left( \frac{ - \sqrt{3} |t_i - t_j|}{\lengthScale} \right), \nonumber \\
  \kernelScalar_{\latentScalar(\text{per})} \left( \mathit{t_i, t_j} \right) 
  & = & 
  \sigma_{\text{per}}^2 \exp \left( -\frac{1}{2} \frac{\sin^2 \left( \frac{2
          \pi}{T} \left( t_i - t_j \right) \right) }{\lengthScale} \right),
  \label{eq:temporalkernels}
\end{eqnarray}
where $\lengthScale$ denotes the characteristic lengthscale and $T$
denotes the period of the periodic covariance function.

\begin{figure}[ht]
  \begin{center}
    \subfigure[]{
      \includegraphics[width=0.31\textwidth]{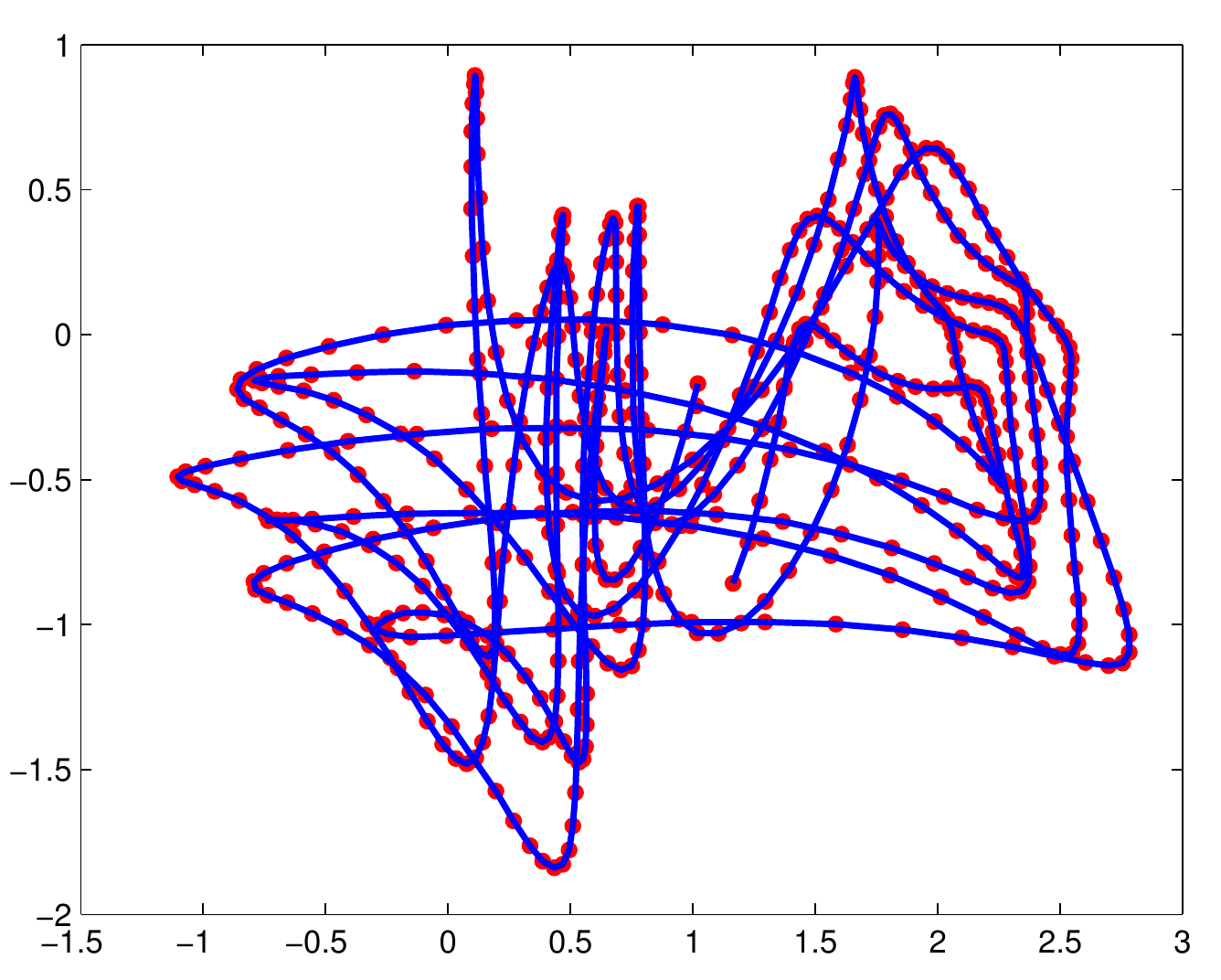}
      \label{fig:rbfPeriodic1}
    }
    \subfigure[]{
      \includegraphics[width=0.31\textwidth]{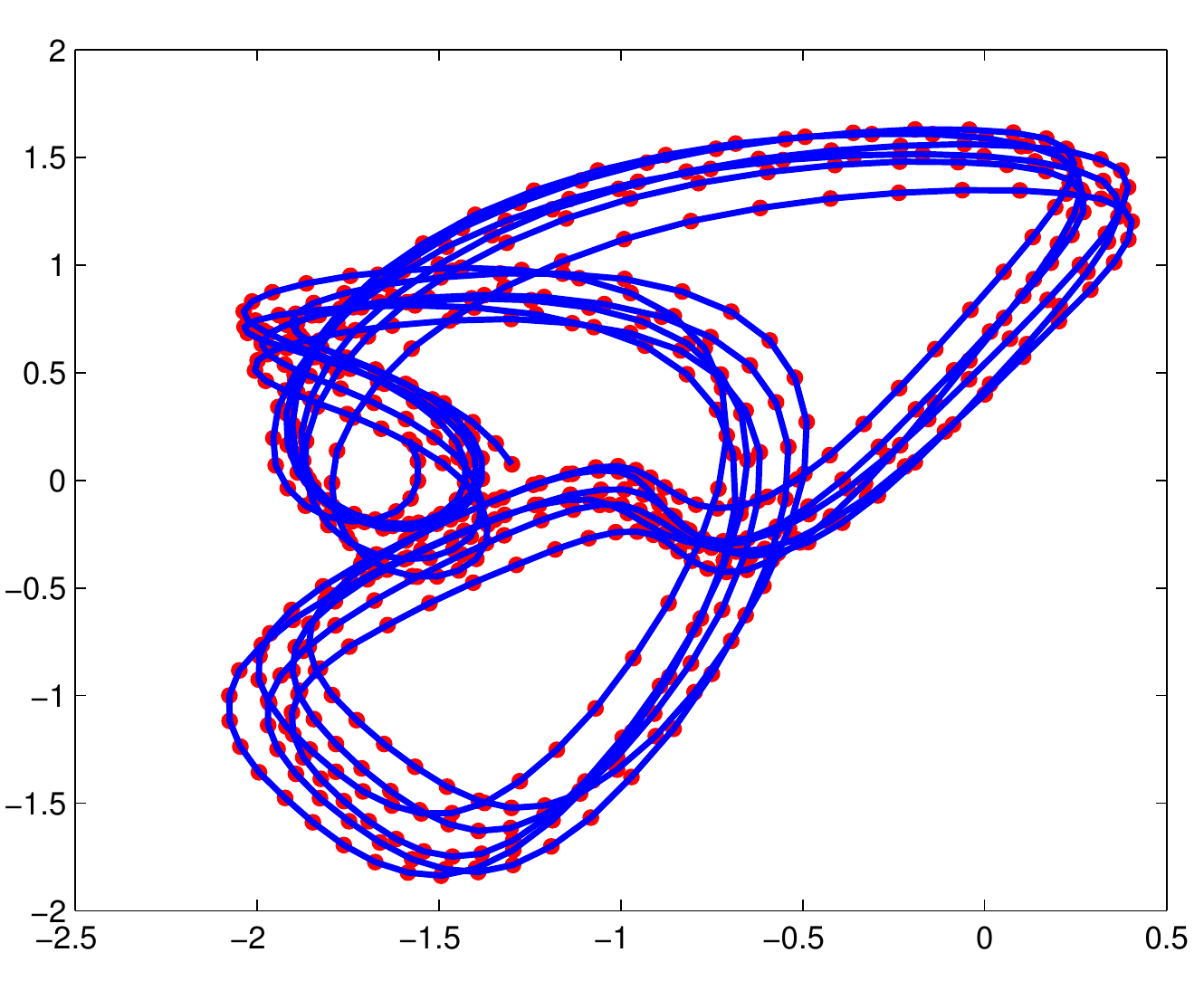}
      \label{fig:rbfPeriodic2}
    }
    \subfigure[]{
      \includegraphics[width=0.31\textwidth]{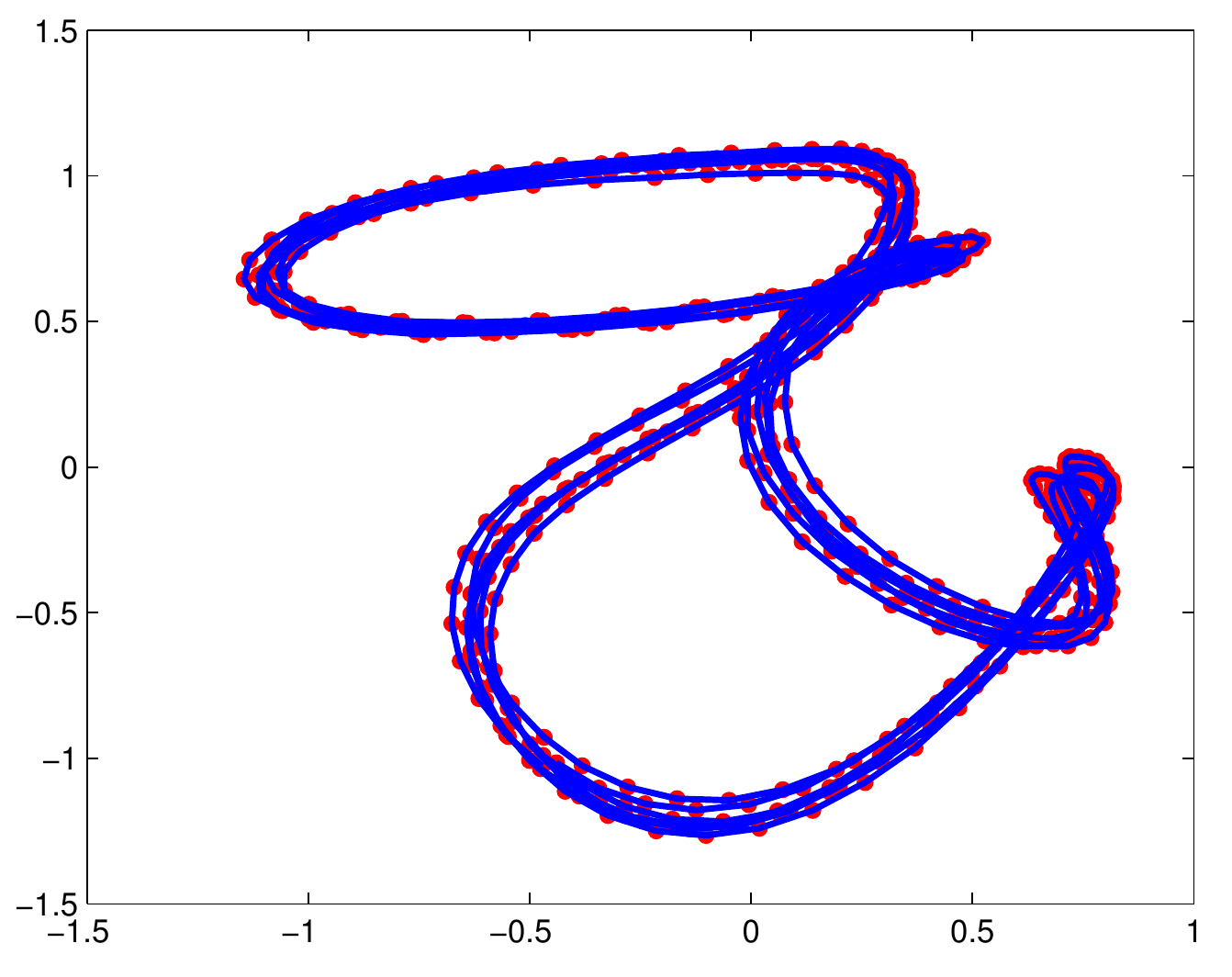}
      \label{fig:rbfPeriodic3}
    }
  \end{center}
  \caption{Typical sample paths drawn from the
    $\kernelScalar_{\latentScalar(\text{per})} +
    \kernelScalar_{\latentScalar(\text{rbf})}$ covariance function.
    The variances are fixed for the two terms, controlling their
    relative effect. In Figures \subref{fig:rbfPeriodic1},
    \subref{fig:rbfPeriodic2} and \subref{fig:rbfPeriodic3}, the ratio
    $\sigma_{\text{rbf}}^2 / \sigma_{\text{per}}^2$ of the two
    variances was large, intermediate and small respectively, causing
    the periodic pattern to be shifted proportionally each period.}
  \label{fig:rbfPeriodic}
\end{figure}

Introducing a separate GP model for the dynamics is a very convenient
way of incorporating any prior information we may have about the
nature of the data in a nonparametric and flexible manner.  In
particular, more sophisticated covariance functions can be constructed
by combining or modifying existing ones. For example, in our
experiments we consider a compound covariance function,
$\kernelScalar_{\latentScalar(\text{per})} +
\kernelScalar_{\latentScalar(\text{rbf})}$ which is suitable for
dynamical systems that are known to be only approximately periodic.
The first term captures the periodicity of the dynamics whereas the
second one corrects for the divergence from the periodic pattern by
enforcing the datapoints to form smooth trajectories in time.  By
fixing the two variances, $\sigma_{\text{per}}^2$ and
$\sigma_{\text{rbf}}^2$ to particular ratios, we are able to control
the relative effect of each kernel. Example sample paths drawn from
this compound covariance function are shown in Figure
\ref{fig:rbfPeriodic}.

For our experiments we additionally include a noise covariance function 
\begin{equation}
  \label{whiteNoise}
  k_{\text{white} }(\latentVector_{i,:}, \latentVector_{k,:}) = \theta_\text{white} \delta_{i,k},
\end{equation}
where $\delta_{i,k}$ is the Kronecker delta function. In that way, we can
define a compound kernel $\kernelScalar + \kernelScalar_{\text{white} }$,
 so that the noise level $\theta_{\text{white} }$ can be
jointly optimised along with the rest of the kernel
hyperparameters. Similarly, one can also include a bias term
$\theta_{\text{bias}}\onesVector$. 

\subsection{Visualisation Tasks \label{sec:experimentsVisualisation}}

Given a dataset with known structure, we can apply our algorithm and
evaluate its performance in a simple and intuitive way, by checking if
the form of the discovered low dimensional manifold agrees with our
prior knowledge.

\begin{figure}
  \begin{center}
      \includegraphics[width=0.31\textwidth]{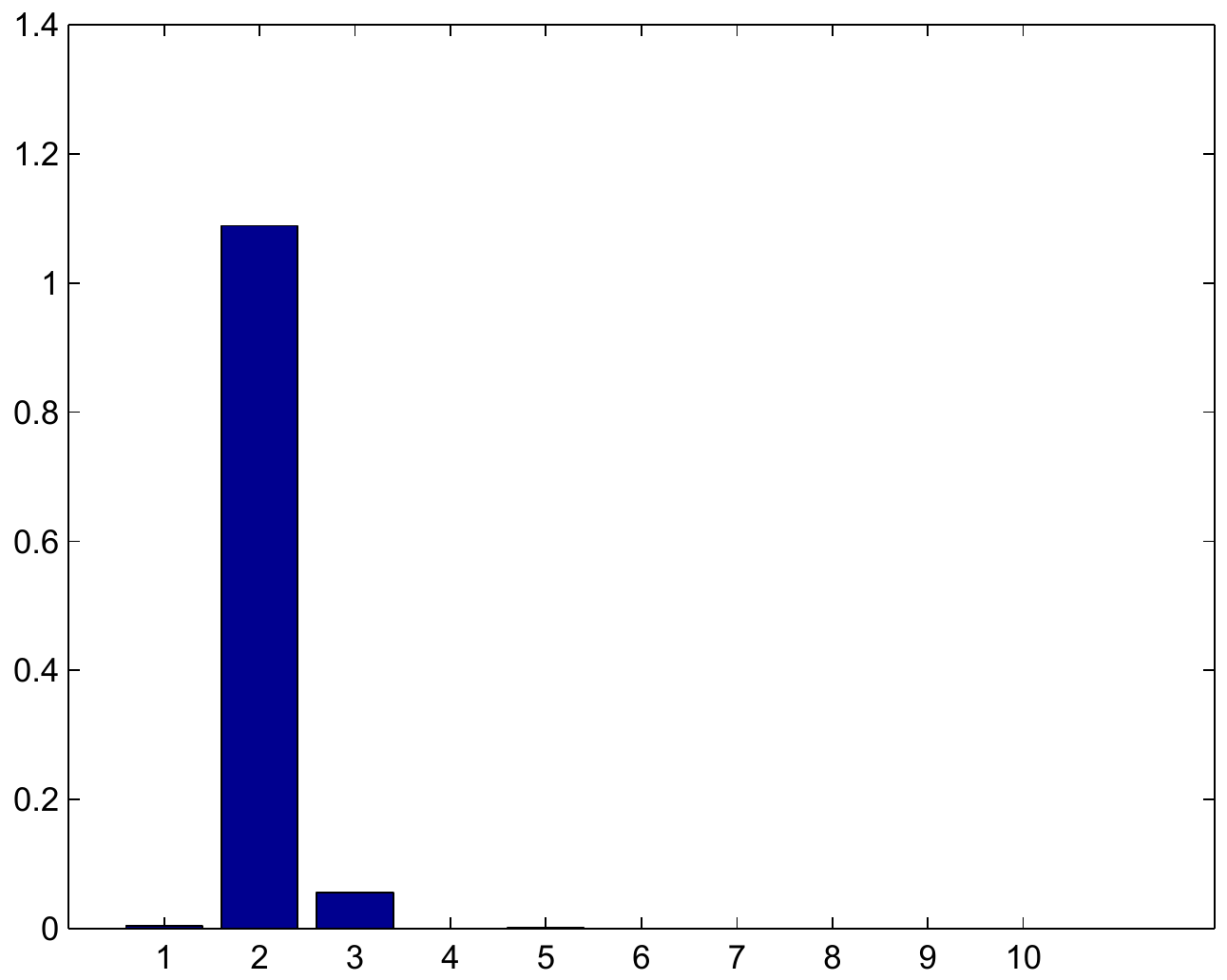}
  \end{center}
\caption{The inverse lengthscales
    found by applying the variational GP-LVM with ARD EQ kernel on the
    oil flow data. }
\label{fig:oilScales}
\end{figure}

\begin{figure}
  \begin{center}
    \subfigure[]{
      \includegraphics[width=0.48\textwidth]{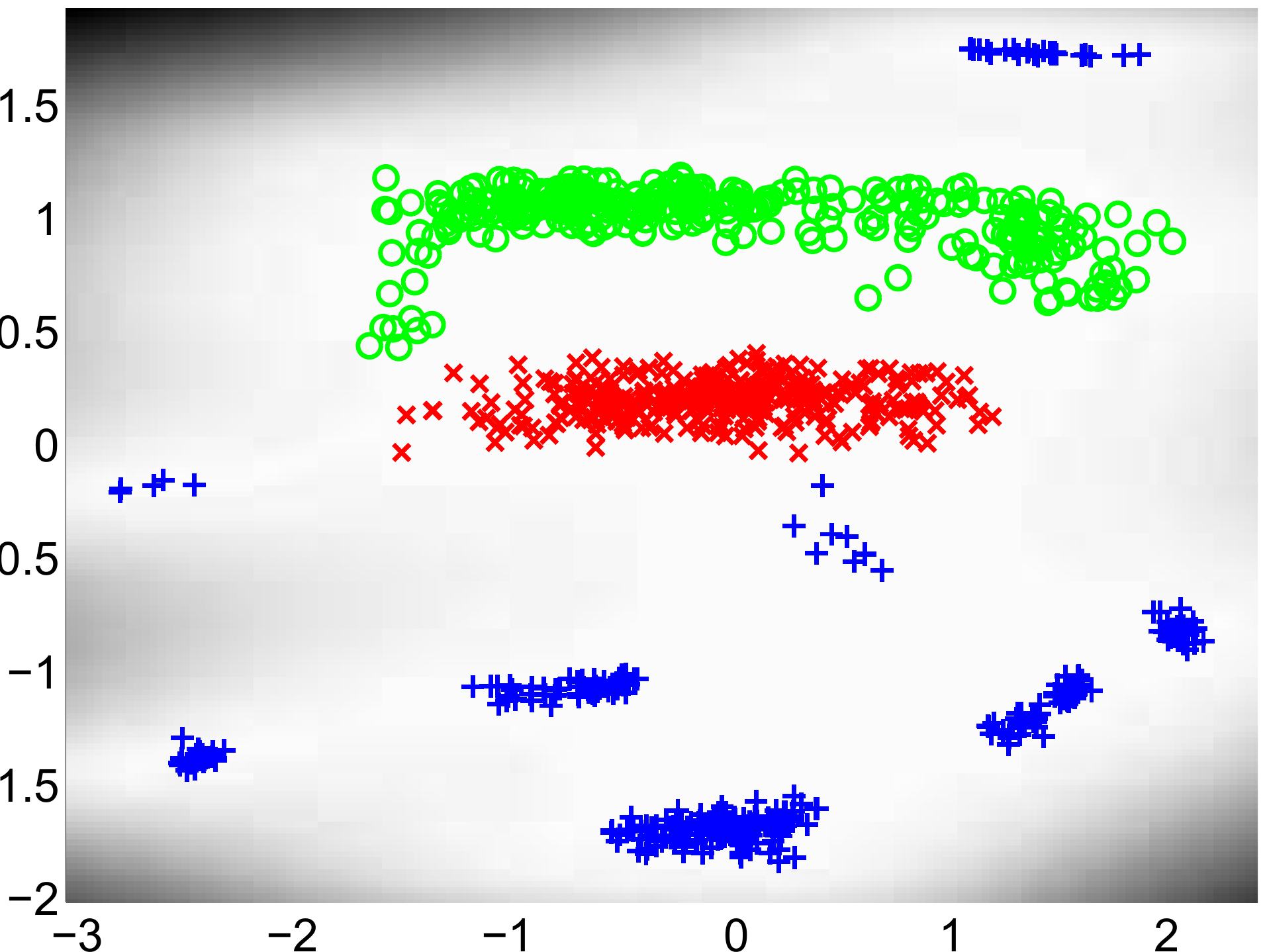} 
      \label{fig:oilBGPLVM}
    }
    \subfigure[]{
      \includegraphics[width=0.48\textwidth]{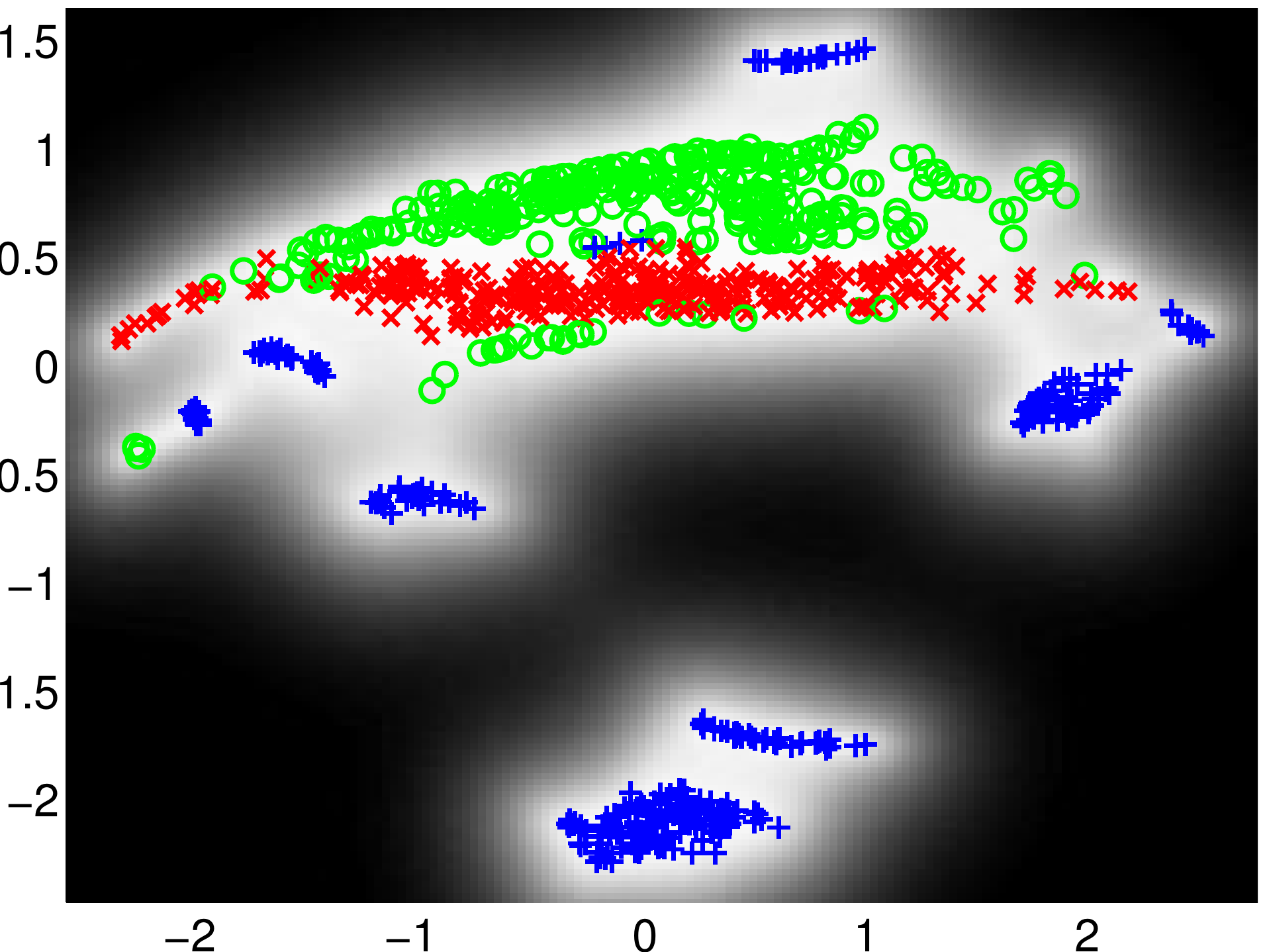}
      \label{fig:oilGPLVM}
    }
  \end{center}
  \caption{Panel \ref{fig:oilBGPLVM} shows the latent space for the variational GP-LVM. Here the dominant latent dimensions are 2 and 3. Dimension 2 is
    plotted on the $y$-axis and 3 and on the $x$-axis. Plot
    \ref{fig:oilGPLVM} shows the visualization found by standard
    sparse GP-LVM initialized with a two dimensional latent space. The nearest neighbor error count for the variational GP-LVM is one. For the standard sparse GP-LVM it is 26.}
  \label{fig:oil}
\end{figure}

We illustrate the method in the multi-phase oil flow data
\citep{Bishop:oil93} that consists of $1,000$, $12$ dimensional
observations belonging to three known classes corresponding to
different phases of oil flow.  Figure \ref{fig:oil} shows the results
for these data obtained by applying the variational GP-LVM with $10$
latent dimensions using the exponentiated quadratic ARD kernel. The
means of the variational distribution were initialized based on PCA,
while the variances in the variational distribution are initialized to
neutral values around $0.5$. As shown in Figure \ref{fig:oilScales},
the algorithm switches off $8$ out of $10$ latent dimensions by making
their inverse lengthscales almost zero. Therefore, the
two-dimensional nature of this dataset is automatically
revealed. Figure \ref{fig:oilBGPLVM} shows the visualization obtained
by keeping only the dominant latent directions which are the
dimensions $2$ and $3$. This is a remarkably high quality two
dimensional visualization of this data. For comparison, Figure
\ref{fig:oilGPLVM} shows the visualization provided by the standard
sparse GP-LVM that runs by a priori assuming only $2$ latent
dimensions.  Both models use $50$ inducing variables, while the latent
variables $\bfX$ optimized in the standard GP-LVM are initialized
based on PCA.  Note that if we were to run the standard GP-LVM with
$10$ latent dimensions, the model would overfit the data, it would not
reduce the dimensionality in the manner achieved by the variational
GP-LVM. The quality of the class separation in the two-dimensional
space can also be quantified in terms of the nearest neighbour error; 
the total error equals the number of training points whose
closest neighbour in the latent space corresponds to a data point of a different
class (phase of oil flow). The number of nearest
neighbour errors made when finding the latent embedding with the standard sparse GP-LVM was $26$ out
of $1000$ points, whereas the variational GP-LVM resulted in only one error.


%


%
%

\subsection{Human Motion Capture Data\label{sec:experimentsMocap}}

In this section we consider a data set associated with temporal information,
as the primary focus of this experiment is on evaluating the dynamical version of the variational GP-LVM.
We followed \cite{Taylor:motion06,Lawrence:larger07} in considering
motion capture data of walks and runs taken from subject 35 in the CMU
motion capture database. We used the dynamical version of our model
and treated each motion as an independent sequence.  The data set was
constructed and preprocessed as described in
\citep{Lawrence:larger07}. This results in 2,613 separate
59-dimensional frames split into 31 training sequences with an average
length of 84 frames each. Our model does not require explicit
timestamp information, since we know a priori that there is a constant
time delay between poses and the model can construct equivalent
covariance matrices given any vector of equidistant time points.

The model is jointly trained, as explained in the last paragraph of Section \ref{temporalPrior},
on both walks and runs, i.e. the algorithm learns a common latent
space for these motions. At test time we investigate the ability of
the model to reconstruct test data from a previously unseen sequence
given partial information for the test targets. This is tested once by
providing only the dimensions which correspond to the body of the
subject and once by providing those that correspond to the legs.
We compare with results in \citep{Lawrence:larger07}, which used MAP
approximations for the dynamical models, and against nearest
neighbour. We can also indirectly compare with the binary latent
variable model (BLV) of \cite{Taylor:motion06} which used a slightly different
data preprocessing. Furthermore, we additionally tested the
non-dynamical version of our model, in order to explore the structure
of the distribution found for the latent space. In this case, the
notion of sequences or sub-motions is not modelled explicitly, as the
non-dynamical approach does not model correlations between
datapoints. However, as will be shown below, the model manages to
discover the dynamical nature of the data and this is reflected in
both, the structure of the latent space and the results obtained on
test data.

The performance of each method is assessed by using the
cumulative error per joint in the scaled space defined in
\citep{Taylor:motion06} and by the root mean square error in the angle space
suggested by \cite{Lawrence:larger07}. Our models were initialized with nine
latent dimensions. For the dynamical version, we performed two runs,
once using the Mat\'ern covariance function for the dynamical prior
and once using the exponentiated quadratic.

The appropriate latent space dimensionality for the data was
automatically inferred by our models.
The non-dynamical model selected a $5$-dimensional latent space.
The model which employed the Mat\'ern covariance to govern the dynamics
retained four dimensions, whereas the model that used the
exponentiated quadratic kept only three.
%
%
The other latent dimensions were completely switched off by the ARD
parameters.

\begin{figure}[ht]
  \begin{center}
    \subfigure[]{
      \includegraphics[width=0.31\textwidth]{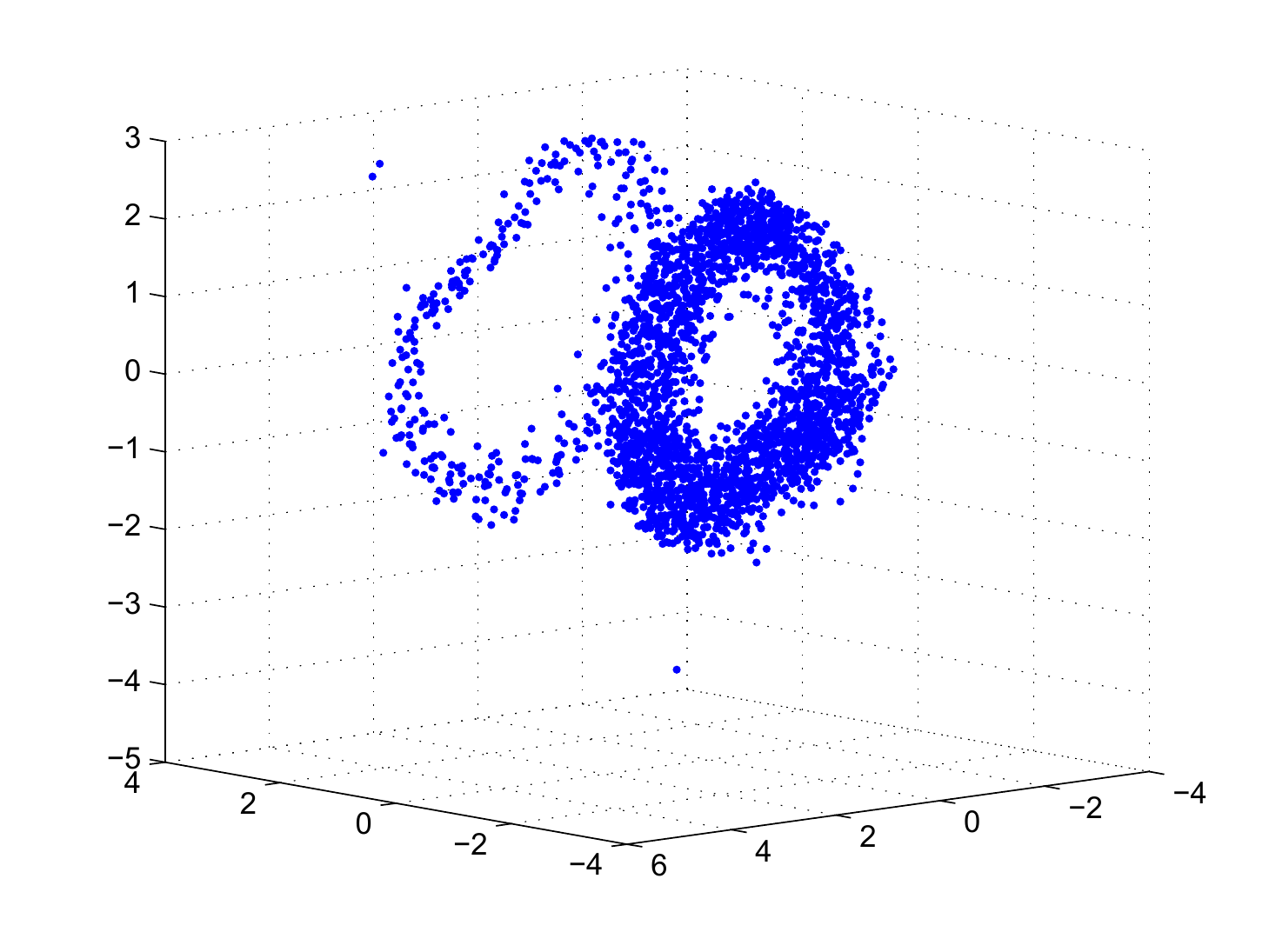}
      \label{fig:cmuLatentSpaceStatic}
    }
    \subfigure[]{
      \includegraphics[width=0.31\textwidth]{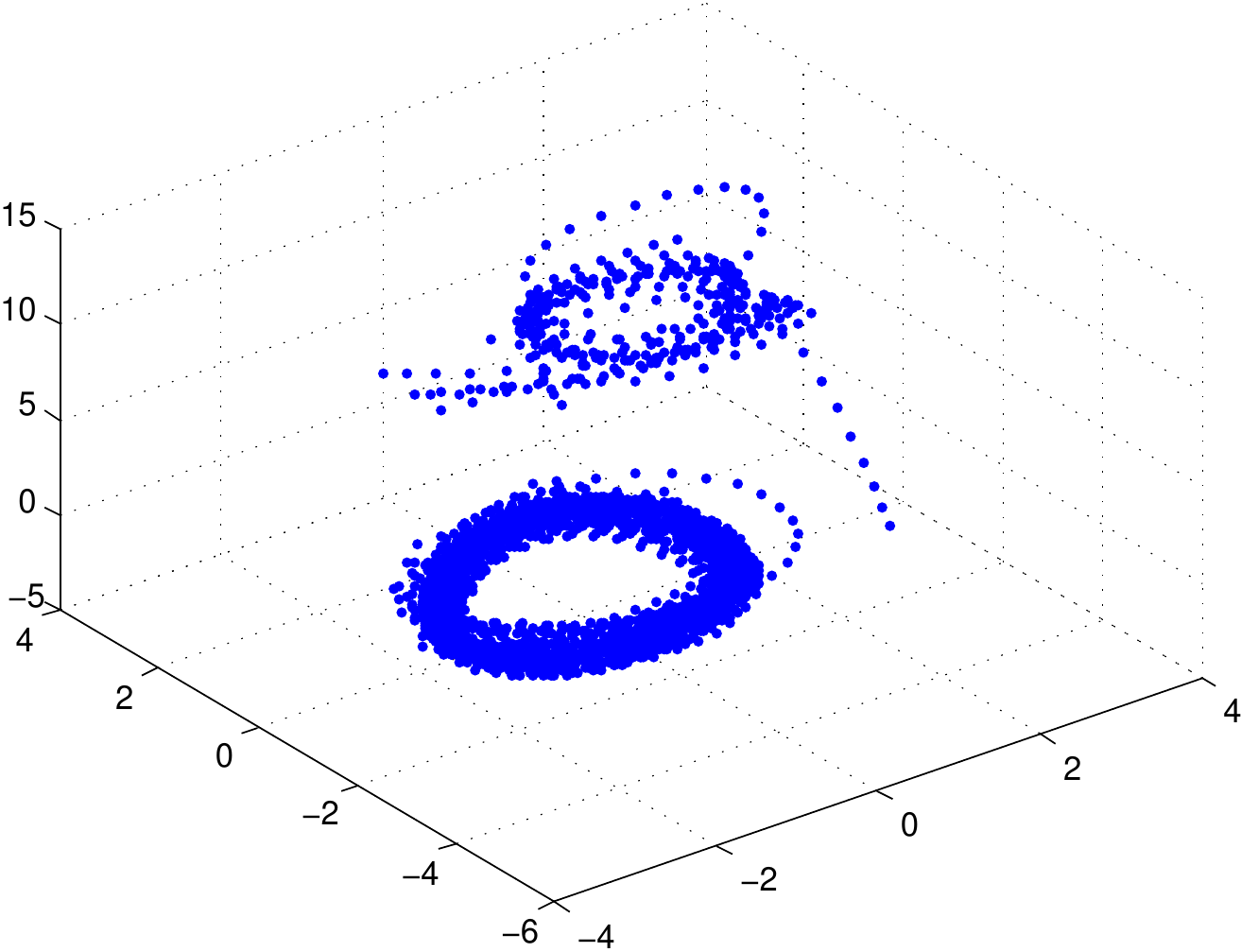}
      \label{fig:cmuLatentSpaceMatern}
    }
    \subfigure[]{
      \includegraphics[width=0.31\textwidth]{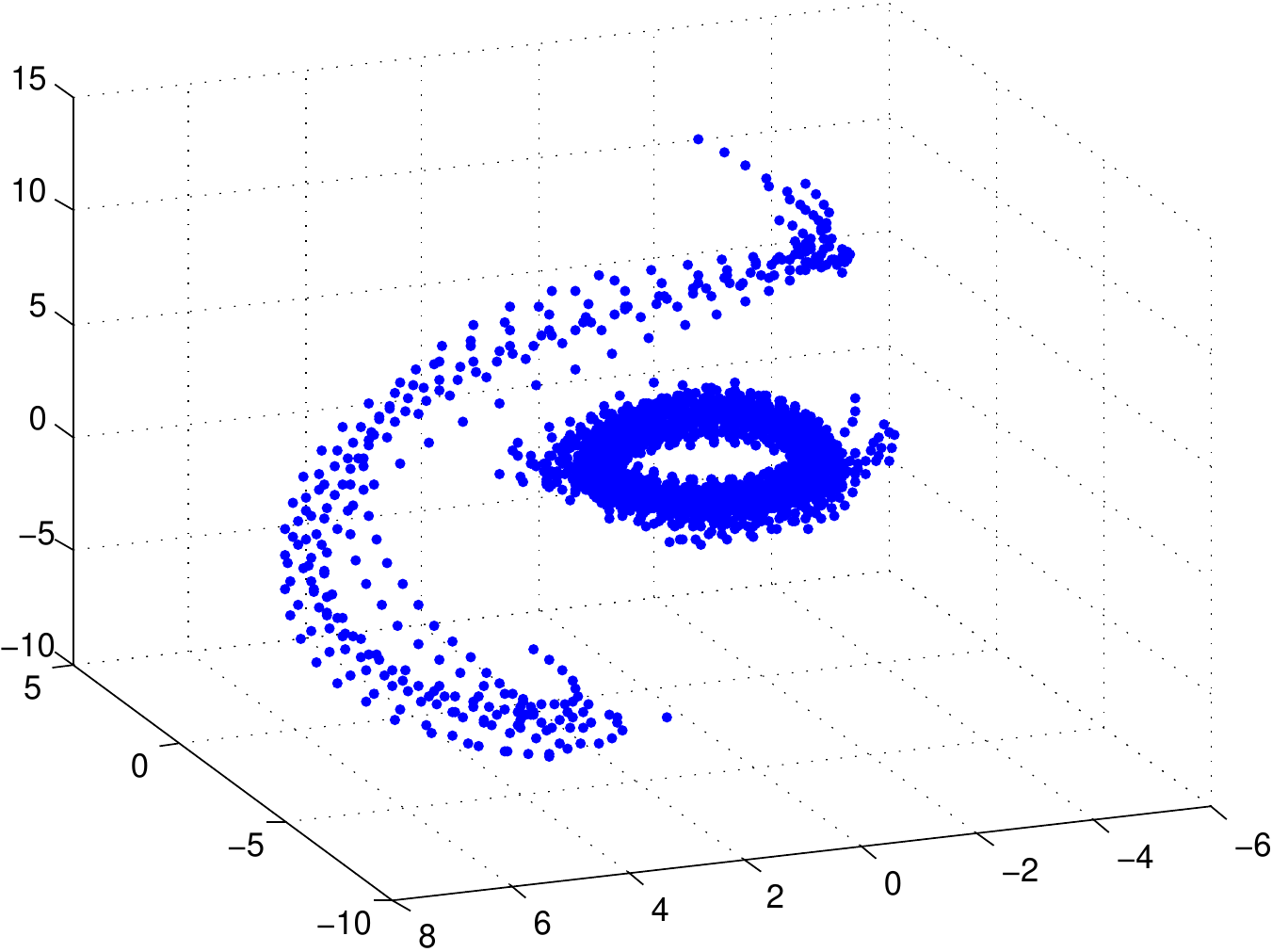}
      \label{fig:cmuLatentSpaceRbf}
    }
  \end{center}
  \caption{ The latent space discovered by our models, projected into
    its three principle dimensions. The latent space found by the
    non-dynamical variational GP-LVM is shown in
    \subref{fig:cmuLatentSpaceStatic}, by the dynamical model which
    uses the Mat\'ern in \subref{fig:cmuLatentSpaceMatern} and by the
    dynamical model which uses the exponentiated quadratic in
    \subref{fig:cmuLatentSpaceRbf}.  }
  \label{fig:cmuLatentSpaces}
\end{figure}

From Table \ref{motionCaptureTable} we see that the dynamical
variational GP-LVM considerably outperforms the other approaches.  The
best performance for the legs and the body reconstruction was achieved
by our dynamical model that used the Mat\'ern and the
exponentiated quadratic covariance function respectively. This is an
intuitive result, since the smoother body movements are expected to be
better modelled using the infinitely differentiable exponentiated
quadratic covariance function, whereas the Mat\'ern one can easier fit
the rougher leg motion.  However, although it is important to take
into account any available information about the nature of the data,
the fact that both models outperform significantly other approaches
shows that the Bayesian training manages successfully to fit the
covariance function parameters to the data in any case.
%
%
Furthermore, the non-dynamical variational GP-LVM, not only manages to
discover a latent space with a dynamical structure, as can be seen in
Figure \ref{fig:cmuLatentSpaceStatic}, but is also proven to be very
robust when making predictions. Indeed, Table \ref{motionCaptureTable}
shows that the non-dynamical variational GP-LVM typically outperforms
nearest neighbor and its performance is comparable to the GP-LVM which
explicitly models dynamics using MAP approximations.
Finally, it is worth highlighting the intuition gained by investigating Figure
\ref{fig:cmuLatentSpaces}. As can be seen, all models split the encoding for the
``walk'' and ``run'' regimes into two subspaces. Further, we notice that the smoother
the latent space is constrained to be, the less ``circular'' is the shape of the
``run'' regime latent space encoding. This can be explained by noticing the ``outliers''
in the top left and bottom positions of plot \subref{fig:cmuLatentSpaceStatic}. 
These latent points correspond to training positions that are very dissimilar to
the rest of the training set but, nevertheless, a temporally constrained model is forced to accommodate
them in a smooth path.
The above intuitions can be confirmed by interacting with the model in real time
graphically, as is presented in the supplementary video.

\begin{table}[h]
  \caption{Errors obtained for the motion capture dataset considering nearest neighbour in the angle space (NN) and in the scaled space (NN sc.), GP-LVM, BLV, variational GP-LVM (VGP-LVM) and Dynamical Variational GP-LVM (Dyn. VGP-LVM). CL / CB are the leg and body data sets as preprocessed in \citep{Taylor:motion06}, L and B the corresponding datasets from \cite{Lawrence:larger07}. SC corresponds to the error in the scaled space, as in Taylor et al. while RA is the error in the angle space. The best error per column is in bold. }
  \label{motionCaptureTable}
  \begin{center}
    \begin{tabular}{c||c|c|c|c|c|c}
      Data & CL & CB & L & L & B & B \\  \hline
      Error Type & SC & SC & SC & RA & SC & RA \\
      \hline \hline
      BLV 			       & 11.7 & \textbf{8.8} & - & - & - & - \\  \hline
      NN sc.   		       & 22.2 & \textbf{20.5} & - & - & - & - \\ \hline
      GP-LVM (\latentDim = 3)	       & - & - & 11.4 & 3.40 & 16.9 & 2.49 \\ \hline
      GP-LVM (\latentDim = 4)	       & - & - & 9.7  & 3.38 & 20.7 & 2.72 \\ \hline
      GP-LVM (\latentDim = 5)	       & - & - & 13.4 & 4.25 & 23.4 & 2.78 \\ \hline
      NN sc.  		       & - & - & 13.5 & 4.44 & 20.8 & 2.62 \\ \hline
      NN 		 	       & - & - & 14.0 & 4.11 & 30.9 & 3.20 \\ \hline
      VGP-LVM                         & - & - & 14.22& 5.09 & 18.79& 2.79 \\ \hline
      Dyn. VGP-LVM (Exp. Quadr.)      & - & - & 7.76 & 3.28 & \textbf{11.95} & \textbf{1.90} \\ \hline
      Dyn. VGP-LVM (Mat\'ern 3/2)     & - & - & \textbf{6.84} & \textbf{2.94} & 13.93 & 2.24 \\
    \end{tabular}
  \end{center}
\end{table}


\subsection{Modeling Raw High Dimensional Video Sequences\label{sec:experimentsVideo}}

For this set of experiments we considered video 
sequences (which are included in the supplementary videos available on-line).
Such sequences are typically preprocessed before modeling to extract
informative features and reduce the dimensionality of the
problem. Here we work directly with the raw pixel values to
demonstrate the ability of the dynamical variational GP-LVM to model data with a vast number
of features. This also allows us to directly sample video from the
learned model.
\par
\begin{figure}[ht]
    \subfigure[]{
      \includegraphics[width=0.48\textwidth]{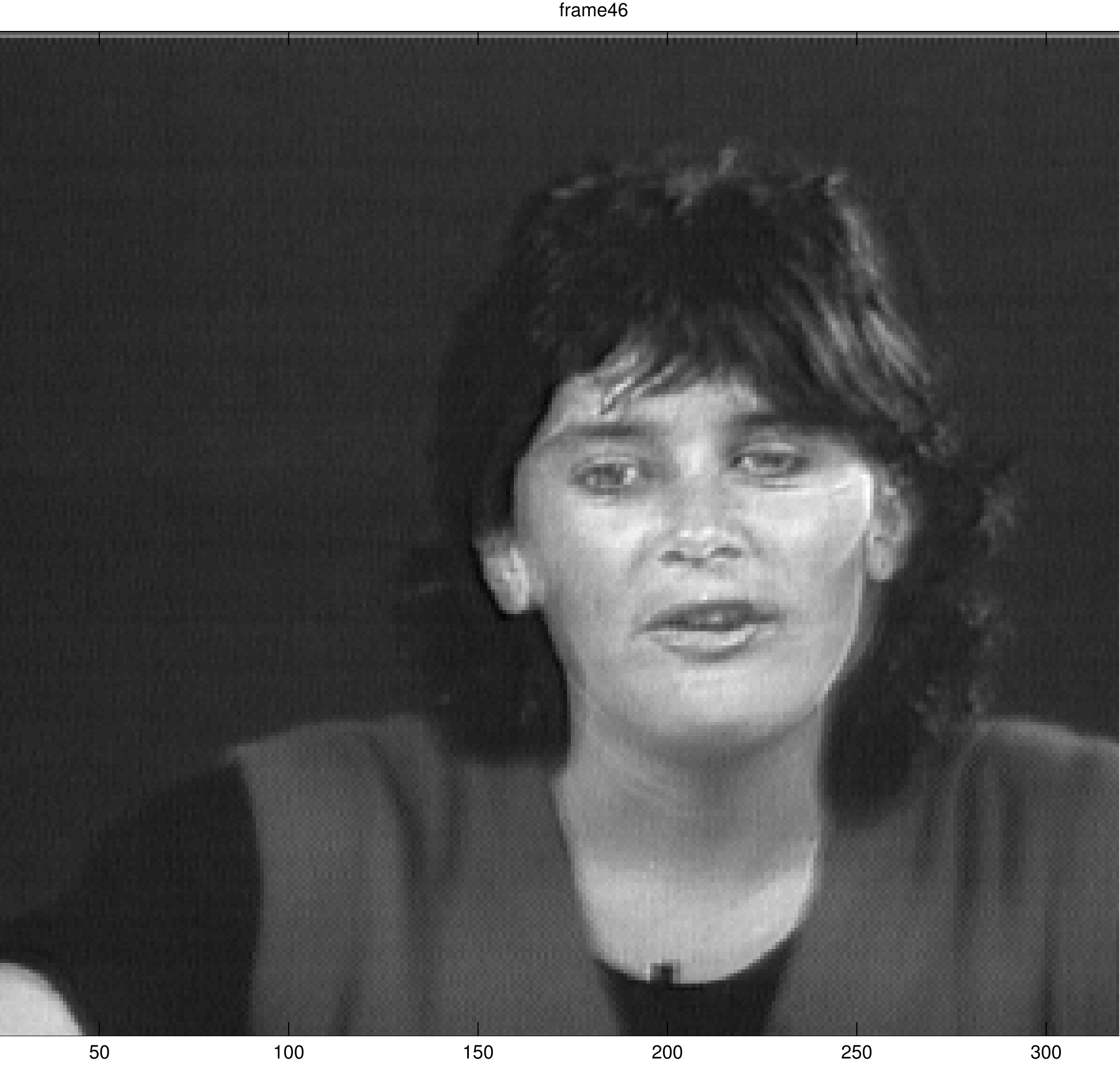}
      \label{fig:missa1}
    }
    \subfigure[]{
      \includegraphics[width=0.48\textwidth]{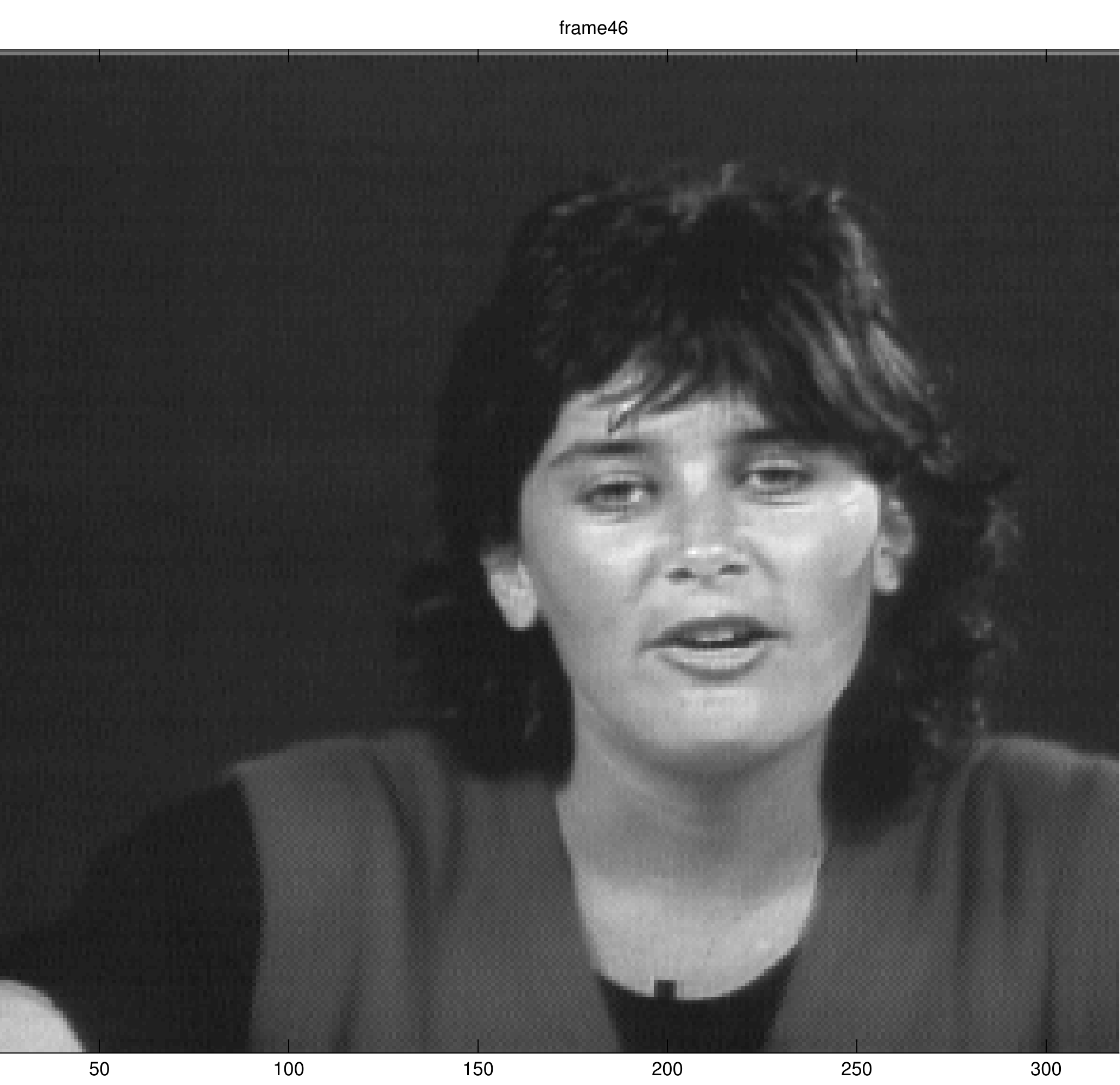}
      \label{fig:missa2}
    }
    \subfigure[]{
      \includegraphics[width=0.48\textwidth]{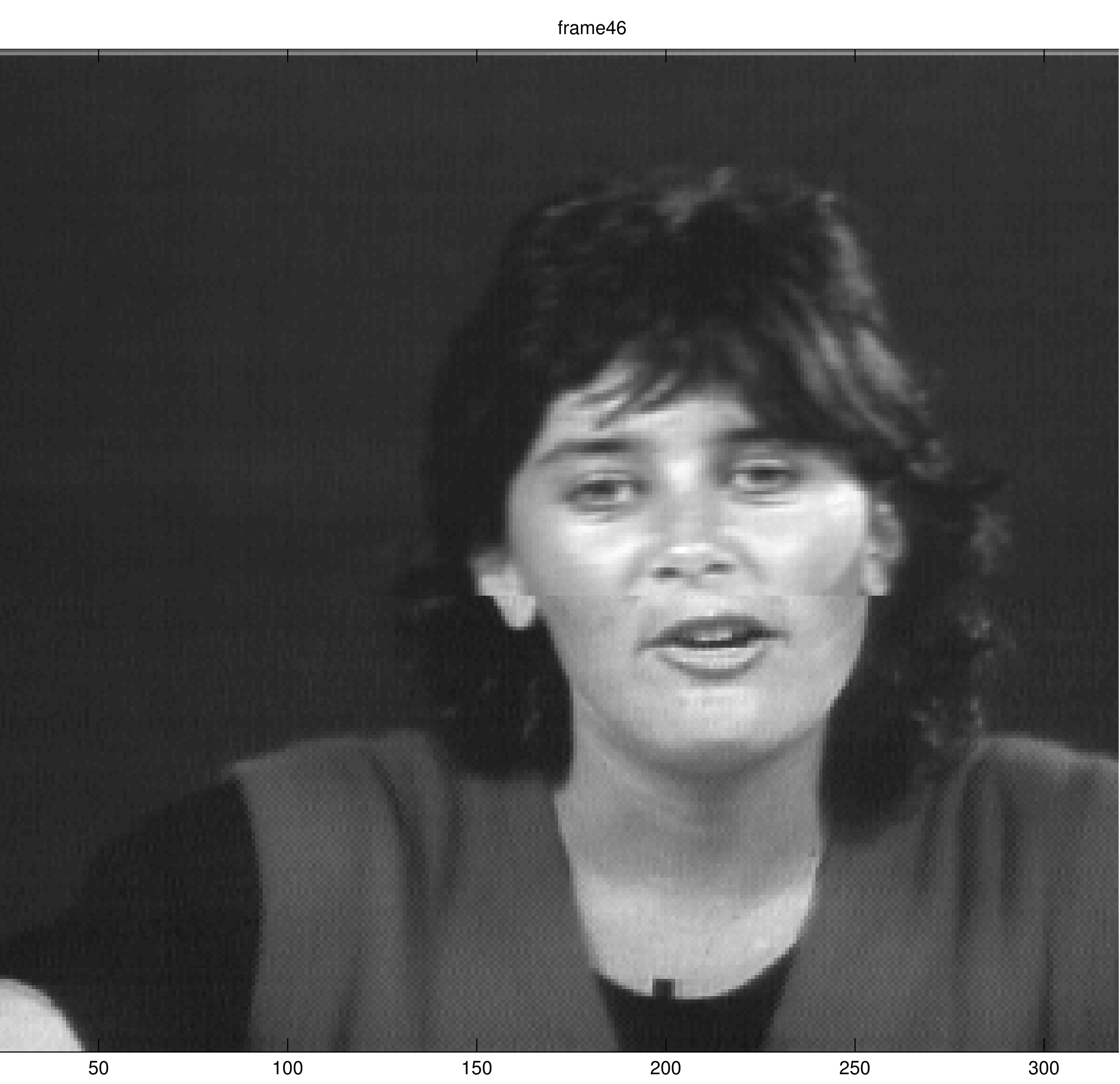}
      \label{fig:missa3}
    }
  \caption{\subref{fig:missa1} and \subref{fig:missa3} demonstrate the reconstruction achieved by dynamical variational GP-LVM and NN respectively for one of the most challenging frames \subref{fig:missa2} of the `missa' video, i.e.\ when translation occurs. In contrast to the NN method, which works in the whole high dimensional pixel space, our method reconstructed the images using a $12$-dimensional compression for the `missa' video.}
  \label{fig:video1}
\end{figure}

 Firstly, we used the model to reconstruct partially observed
frames from test video sequences \footnote{`Missa' dataset:
  cipr.rpi.edu. `Ocean': cogfilms.com. `Dog': fitfurlife.com.  See
  details in supplementary. The logo appearing in the `dog' images in
  the experiments that follow, has been added with post-processing.}.
For the first video discussed here we gave as partial information
approximately 50\% of the pixels while for the other two we gave
approximately 40\% of the pixels on each frame.  The mean squared
error per pixel was measured to compare with the $k-$nearest neighbour
(NN) method, for $k \in (1,..,5)$ (we only present the error achieved
for the best choice of $k$ in each case). The datasets considered are
the following: firstly, the `Missa' dataset, a standard benchmark used
in image processing. This is a 103,680-dimensional video, showing a
woman talking for 150 frames. The data is challenging as there are
translations in the pixel space. We also considered an HD video of
dimensionality $9 \times 10^5$ that shows an artificially created
scene of ocean waves as well as a $230,400-$dimensional video showing
a dog running for $60$ frames. The later is approximately periodic in
nature, containing several paces from the dog. For the first two
videos we used the Mat\'ern and exponentiated quadratic covariance
functions respectively to model the dynamics and interpolated to
reconstruct blocks of frames chosen from the whole sequence. For the
`dog' dataset we constructed a compound kernel $k_x =
k_{x(\text{rbf})} + k_{x(\text{per})}$ presented in section
\ref{covarianceFunctions}, where the exponentiated quadratic (RBF) term is
employed to capture any divergence from the approximately periodic
pattern. We then used our model to reconstruct the last 7 frames
extrapolating beyond the original video. As can be seen in Table
\ref{videoResultsTable}, our method outperformed NN in all cases. The
results are also demonstrated visually in Figures \ref{fig:video1}, \ref{fig:video2}, \ref{fig:video3} and
\ref{fig:dogRec} and the reconstructed videos are available in the
supplementary material.

\begin{table}[h]
  \caption{
    The mean squared error per pixel for Dyn. VGP-LVM and NN for the three datasets (measured only in the missing inputs).
    The number of latent dimensions selected by our model is in parenthesis. 
  } 
  \label{videoResultsTable}
  \begin{center}
    \begin{tabular}{c||l|l|l}
      & Missa & Ocean & Dog \\
      \hline \hline
      Dyn. VGP-LVM  & 2.52 ($\latentDim = 12$) & 9.36 ($\latentDim = 9$)  & 4.01 ($\latentDim = 6$) \\  \hline
      NN           & 2.63 & 9.53 & 4.15 \\
    \end{tabular}
  \end{center}
\end{table}

\begin{figure}[ht]
  \subfigure[]{
      \includegraphics[width=0.48\textwidth]{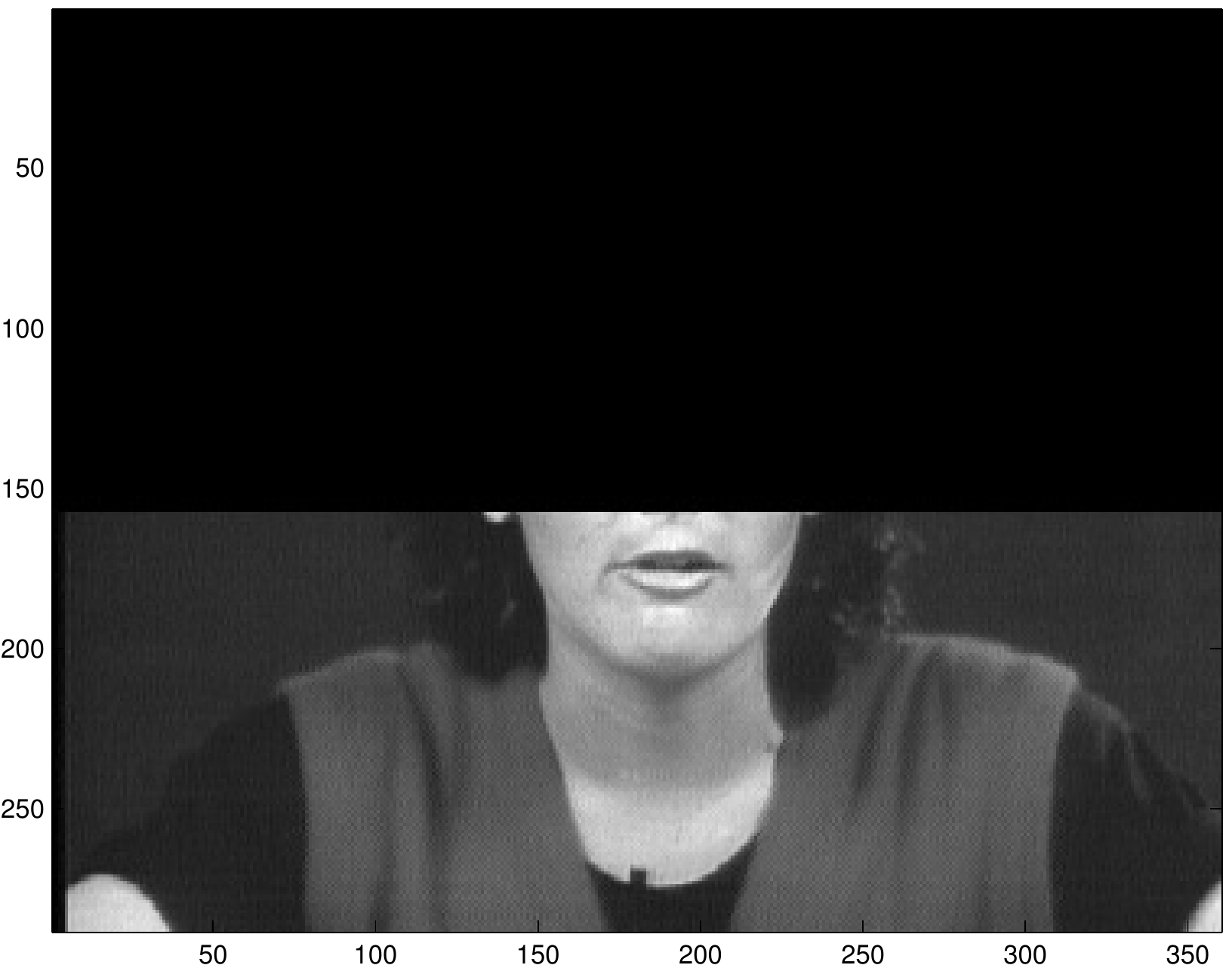}
  }
  \subfigure[]{
      \includegraphics[width=0.48\textwidth]{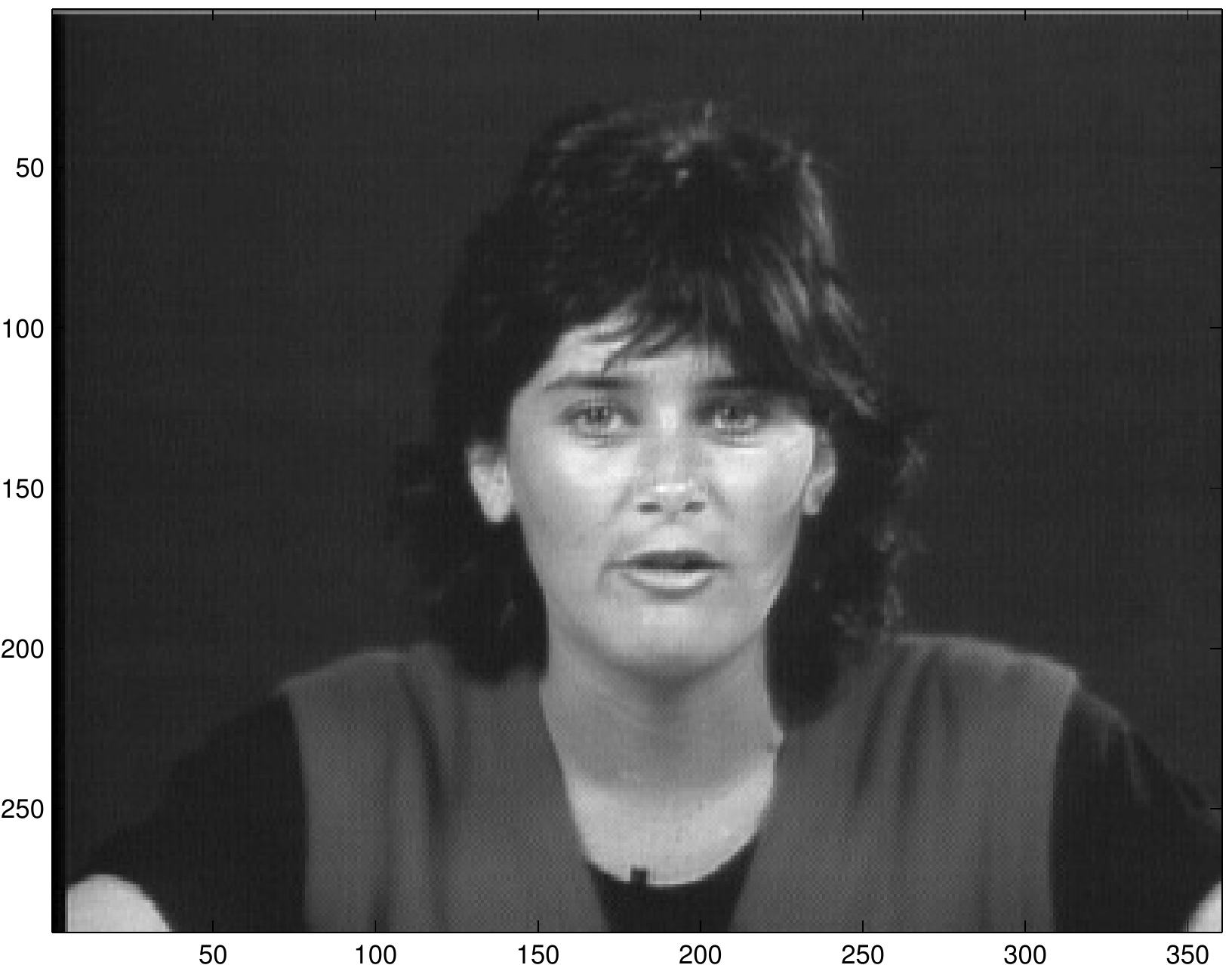}
  }
  \caption{Another example of the reconstruction achieved by the dynamical variational GP-LVM given the partially observed image.}
  \label{fig:video2}
\end{figure}

\begin{figure}
    \subfigure[]{
      \includegraphics[width=0.39\textwidth]{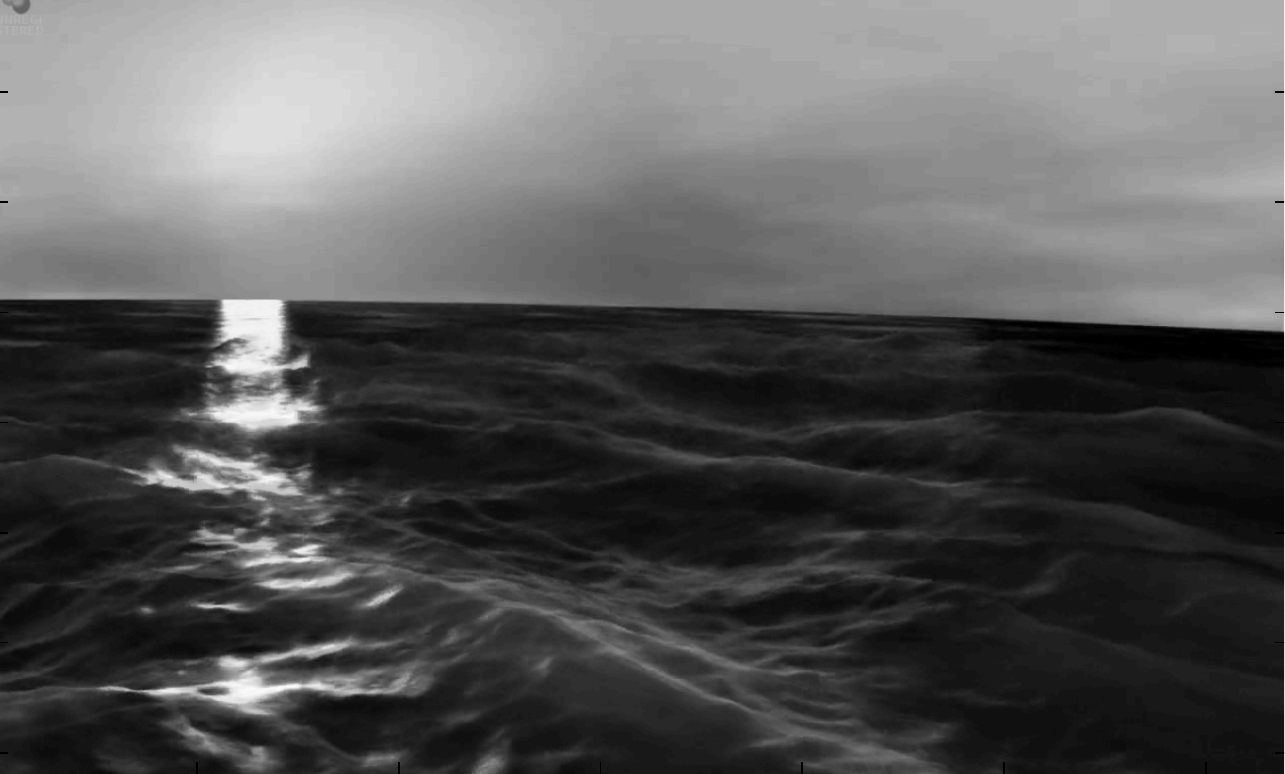}
      \label{fig:ocean1}
    }
    \subfigure[]{
      \includegraphics[width=0.39\textwidth]{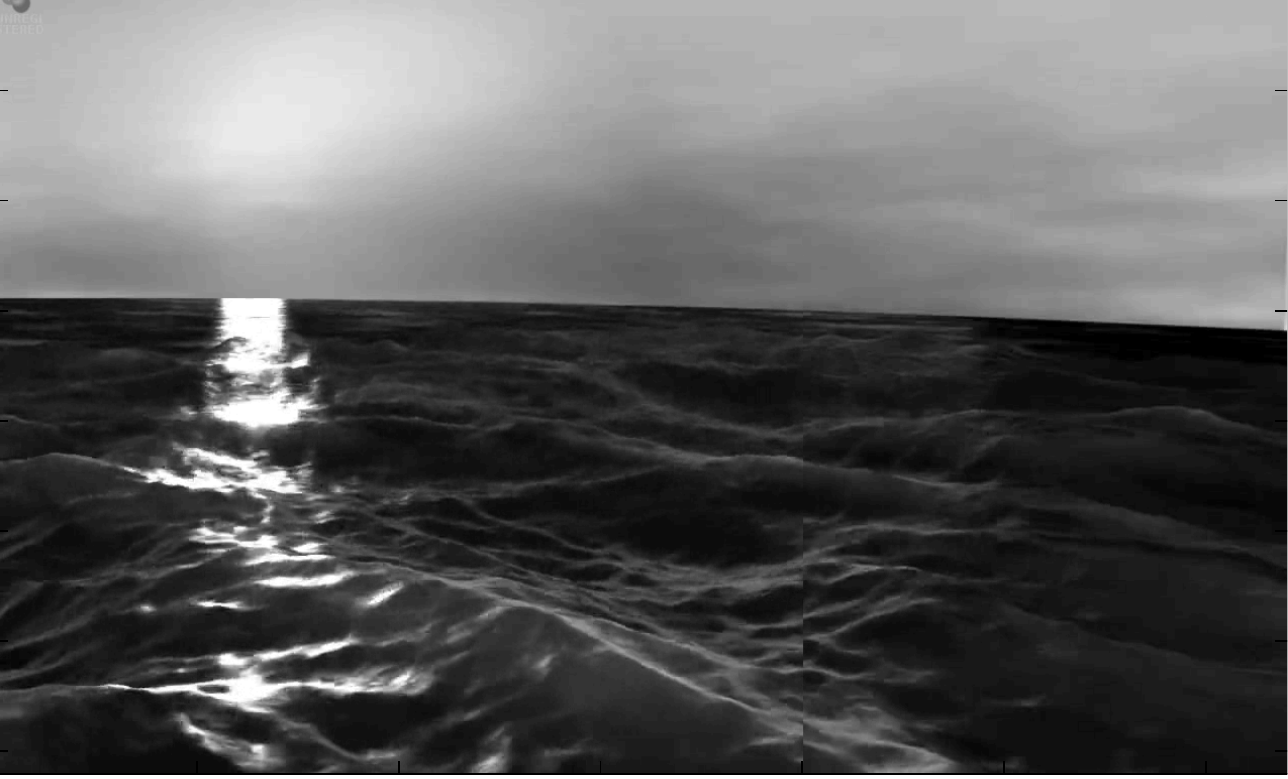}
      \label{fig:ocean3}
    }
    \subfigure[]{
      \includegraphics[width=0.08\textwidth]{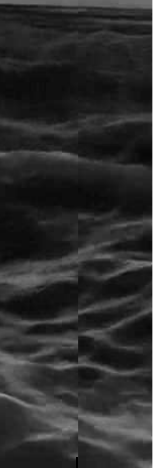}
      \label{fig:ocean4}
    }
  \caption{
    \subref{fig:ocean1} (Dynamical variational GP-LVM) and \subref{fig:ocean3} (NN) depict the reconstruction achieved for a frame of the `ocean' dataset. 
    Notice that in both of the aforementioned datasets, our method recovers a smooth image, in contrast to the simple NN (a close up of this problem with NN for the `ocean' video is shown in Figure \subref{fig:ocean4}). The dynamical var. GP-LVM reconstructed the ocean images using a $9$-dimensional compression for the video.
  }
  \label{fig:video3}
\end{figure}
As can be seen in Figures \ref{fig:video1}, \ref{fig:video2} and \ref{fig:video3},
the dynamical variational GP-LVM predicts pixels which
are smoothly connected with the observed part of the image, whereas the NN method
cannot fit the predicted pixels in the overall context.
Figure \ref{fig:video1}\subref{fig:ocean4} focuses on this specific problem with NN, but it can be seen more evidently in the corresponding video files.

\begin{figure}[ht]
  \begin{center}
    \subfigure[]{
      \includegraphics[width=0.92\textwidth]{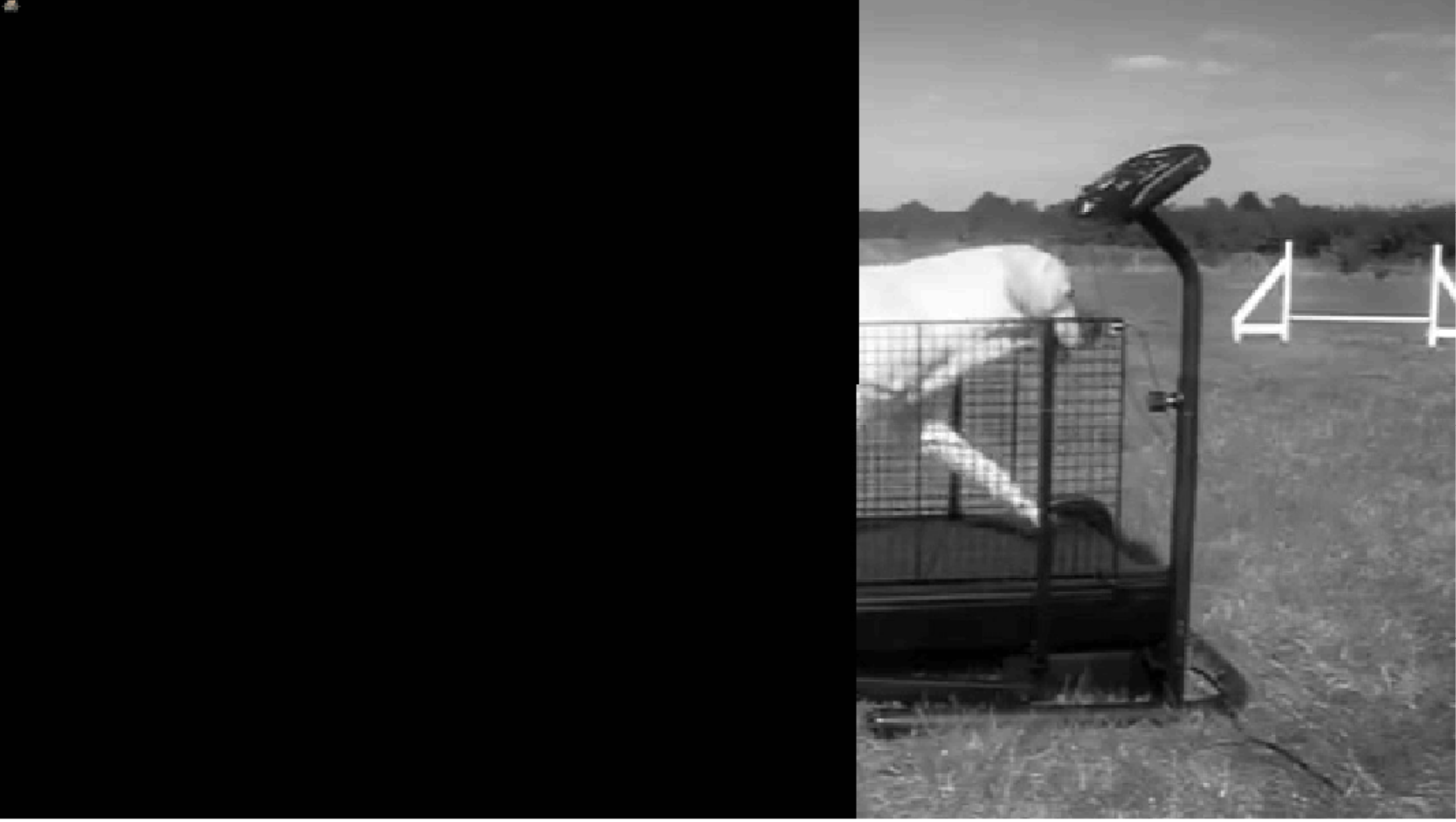}
      \label{fig:dog3}
    }\\
    \subfigure[]{
      \includegraphics[width=0.92\textwidth]{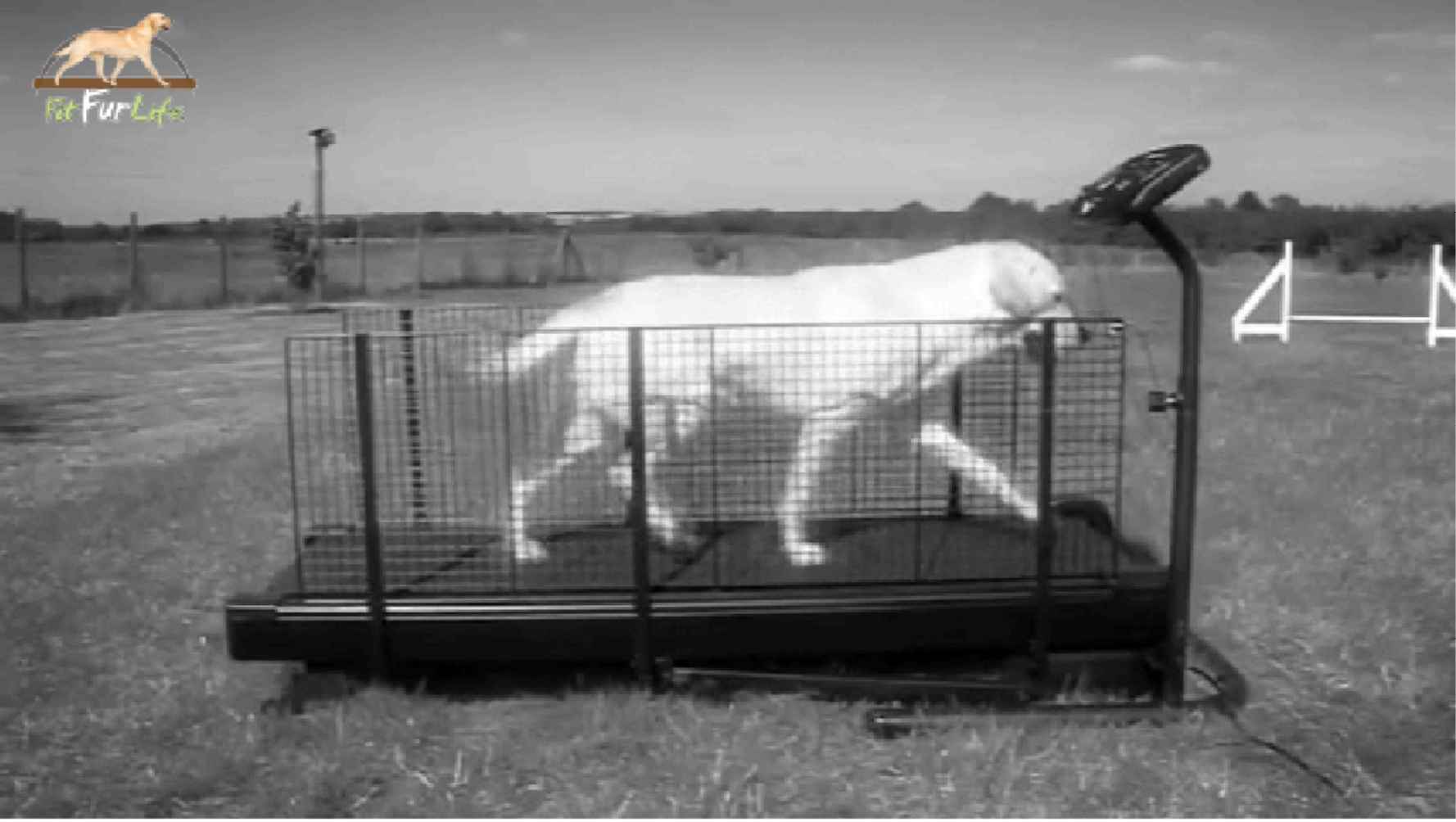}
      \label{fig:dog4}
    }
  \end{center}
  \caption{ An example for the reconstruction achieved for the `dog' dataset. $40\%$ of the test image's pixels (Figures \subref{fig:dog3}  were presented  to the model, which was able to successfully reconstruct them, as can be seen in \subref{fig:dog4}.}
  \label{fig:dogRec}
\end{figure}

\begin{figure}
  \begin{center}
    \subfigure[]{
      \includegraphics[width=0.46\textwidth]{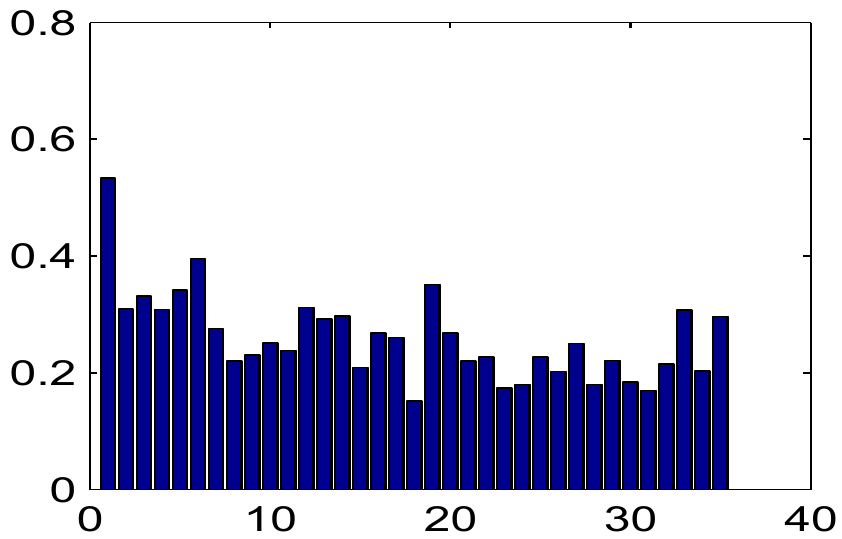}
      \label{fig:scalesDogInit}
    }
    \subfigure[]{
      \includegraphics[width=0.46\textwidth]{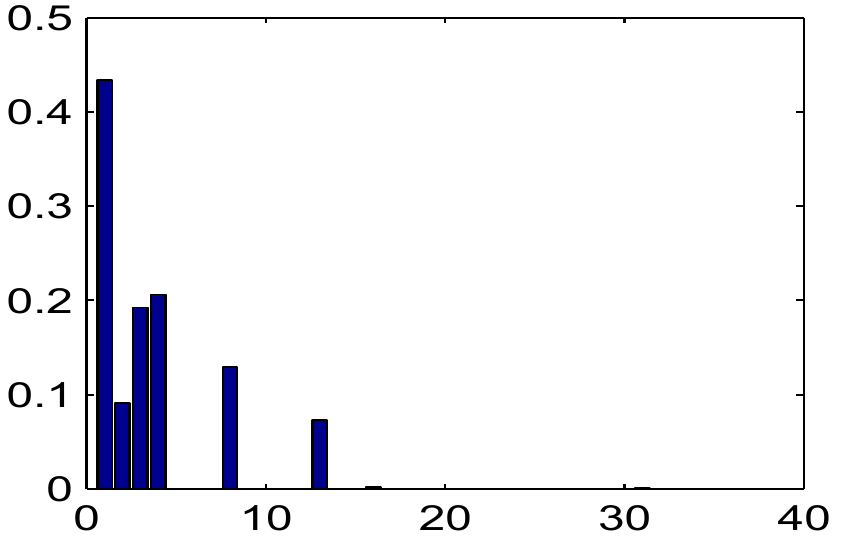}
      \label{fig:scalesDogOpt}
    }      
  \end{center}
  \caption{
    Here, we also demonstrate the ability of the model to automatically select the latent dimensionality by showing the initial lengthscales (fig: \subref{fig:scalesDogInit}) of the ARD covariance function and the values obtained after training (fig: \subref{fig:scalesDogOpt}) on the `dog' data set.
  }
\end{figure}


\par As a second task, we used our generative model to create new
samples and generate a new video sequence. This is most effective for
the `dog' video as the training examples were approximately periodic
in nature. The model was trained on 60 frames (time-stamps $[t_1,
t_{60}]$) and we generated new frames which correspond to the next
40 time points in the future. The only input given for this generation
of future frames was the time-stamp vector, $[t_{61}, t_{100}]$. The
results show a smooth transition from training to test and amongst the
test video frames. The resulting video of the dog continuing to run is
sharp and high quality. This experiment demonstrates the ability of
the model to reconstruct massively high dimensional images without
blurring. Frames from the result are shown in Figure
\ref{fig:dog}. The full video is available in the supplementary
material.

\begin{figure}[ht]
  \begin{center}
    \subfigure[]{
      \includegraphics[width=0.6\textwidth]{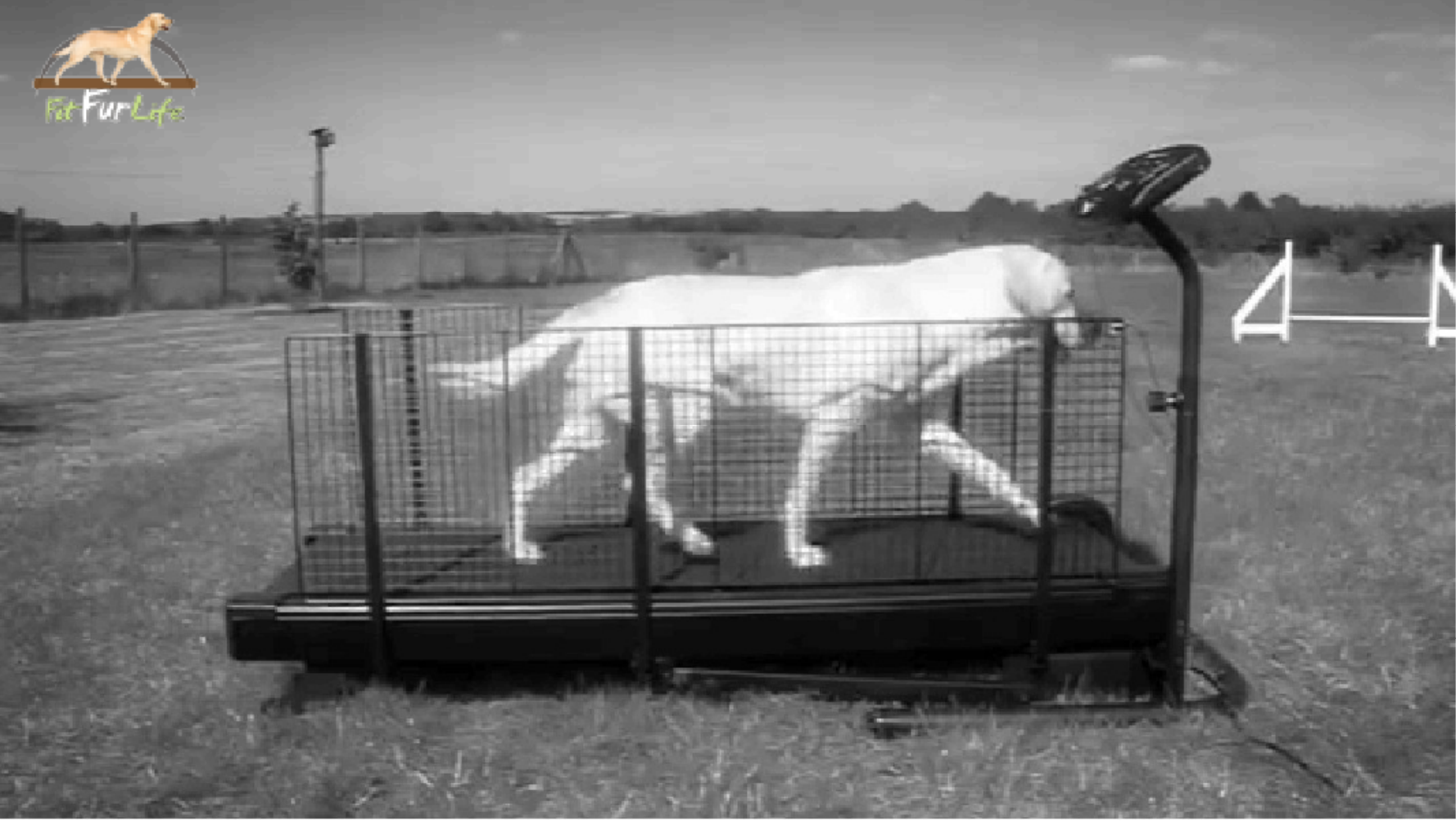}
      \label{fig:dogTrain}
    }\\
    \subfigure[]{
      \includegraphics[width=0.6\textwidth]{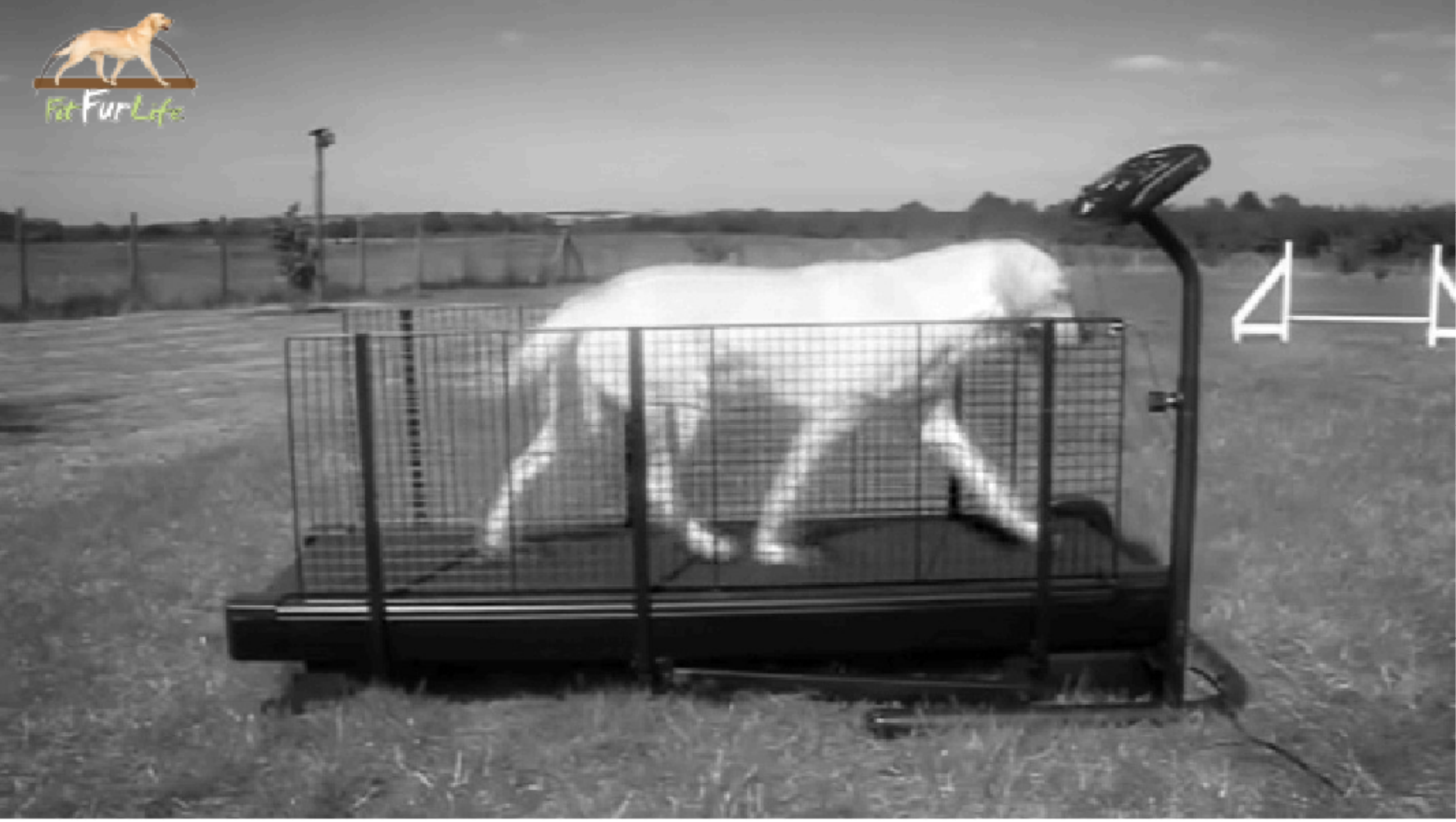}
      \label{fig:dogTest1}
    }
    \subfigure[]{
      \includegraphics[width=0.6\textwidth]{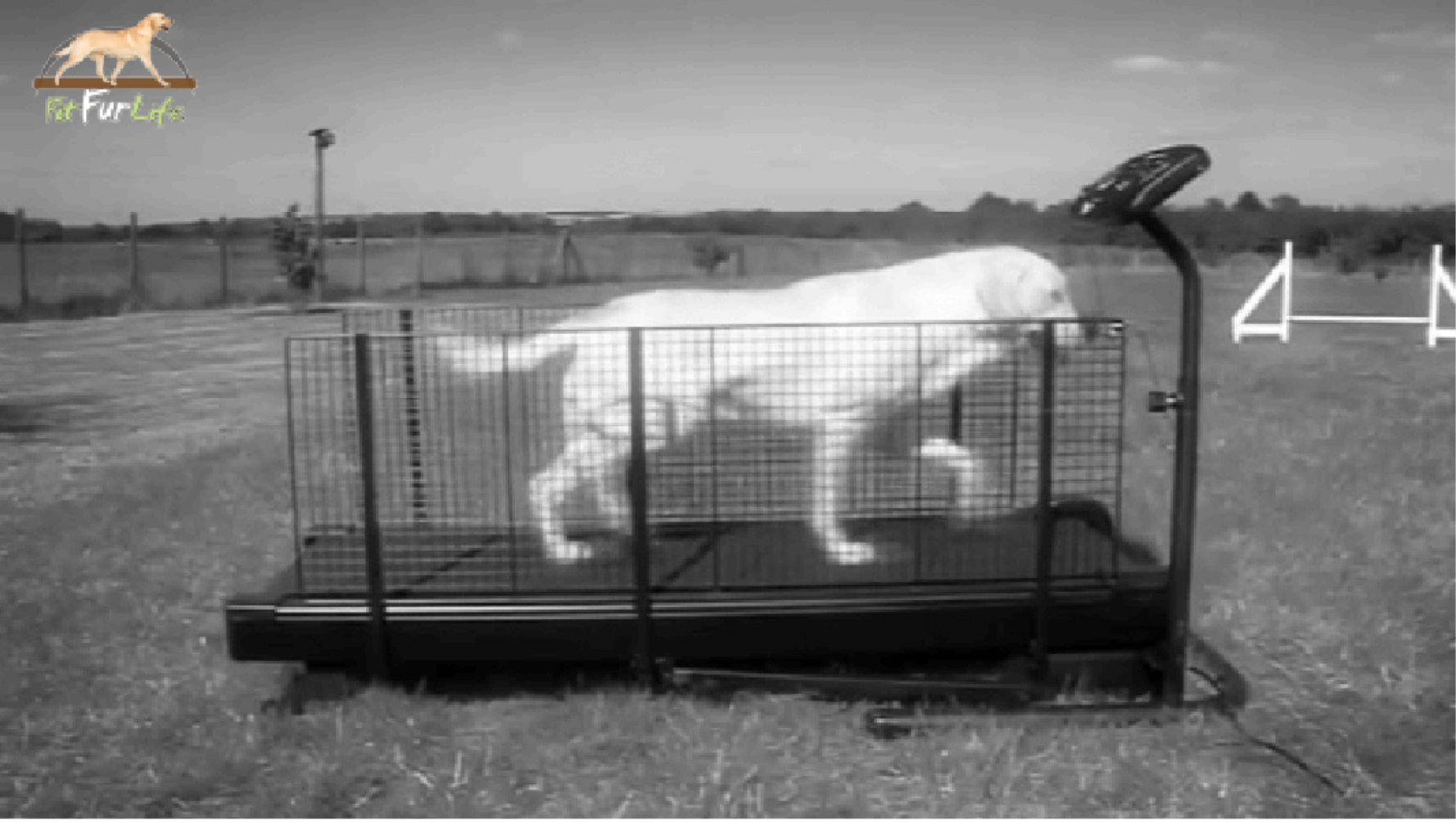}
      \label{fig:dogTest2}
    }
  \end{center}
  \caption{ 
    The last frame of the training video \subref{fig:dogTrain} is smoothly followed by the first frame \subref{fig:dogTest1} of the generated video. A subsequent generated frame can be seen in \subref{fig:dogTest2}.}
  \label{fig:dog}
\end{figure}

\subsection{Class Conditional Density Estimation \label{sec:classification}}

In this experiment we use the variational GP-LVM to build a
generative classifier for handwritten digit recognition.  We consider
the well known USPS digits dataset. This dataset consists of $16
\times 16$ images for all $10$ digits and it is divided into $7,291$
training examples and $2,007$ test examples.  We run $10$ variational
GP-LVMs, one for each digit, on the USPS data base. We used $10$
latent dimensions and $50$ inducing variables for each model. This
allowed us to build a probabilistic generative model for each digit so
that we can compute Bayesian class conditional densities in the test
data having the form $p(\dataMatrix_*|\dataMatrix, \text{digit})$. These class conditional
densities are approximated through the ratio of lower bounds in
eq. \eqref{eq:predictive1} as described in Section \ref{section:predictions}.
The whole approach allows us to classify new digits by determining the
class labels for test data based on the highest class conditional
density value and using a uniform prior over class labels. For comparison
we used a 1-vs-all logistic regression classification approach.
As shown in Table \ref{table:usps}, the variational GP-LVM outperforms this baseline.

\begin{table}[h]
  \caption{The test error made by the variational GP-LVM and 1-vs-all Logistic Regression
  classification in the whole set of $2,007$ test points.}
  \label{table:usps}
  \begin{center}
    \begin{tabular}{c||r|r|}
       & \# misclassified & error (\%) \\  \hline
      \hline
      variational GP-LVM & $\mathbf{95}$ & $\mathbf{4.73}$ \% \\  \hline
      Logistic Regression & $283$ & $14.10$ \%  \\
    \end{tabular}
  \end{center}
\end{table}


%% file: extensions.tex

\section{\label{section:extensions} Extensions for Different Kinds of Inputs}

So far we considered the typical dimensionality reduction scenario where, given high-dimensional output
data we seek to find a low-dimensional latent representation in a completely unsupervised manner.
For the dynamical variational GP-LVM we have additional temporal information, but the input space $\latentMatrix$
from where we wish to propagate the uncertainty is still treated as fully unobserved.
However, our framework for propagating the input uncertainty through the GP mapping is applicable
to the full spectrum of cases, ranging from fully unobserved to fully observed inputs with
known or unknown amount of uncertainty per input.
In this section we discuss these cases and, further, show how they give rise
to an auto-regressive model (Section \ref{uncertainInputs}) and a semi-supervised GP model (Section \ref{sec:semiSupervisedExtension}).


\subsection{Gaussian Process Inference with Uncertain Inputs\label{uncertainInputs}}
Gaussian processes have been used
extensively and with great success in a variety of regression tasks.
In the most common setting, we are given a dataset of observed
input-output pairs, denoted as $\bfZ \in \Re^{\numData \times
  \latentDim}$ and $\dataMatrix \in \Re^{\numData \times \dataDim}$
respectively, and we wish to infer the unknown outputs $\dataMatrix^*
\in \Re^{\numData^* \times \dataDim}$ corresponding to some novel
given inputs $\bfZ^* \in \Re^{\numData^* \times \latentDim}$. However,
in many real-world applications the inputs are uncertain, for example
when measurements come from noisy sensors.  In this case, the GP
methodology cannot be trivially extended to account for the variance
associated with the input space \citep{Girard:uncertain01,
  mchutchon:gaussian}.  The aforementioned problem is also closely
related to the field of heteroscedastic Gaussian process regression,
where the uncertainty in the noise levels is modelled in the output
space as a function of the inputs
\citep{Kersting:MLH07,Bishop:gps_nips97,gredilla:variationalheteroscedastic11}.

In this section we show that our variational framework can be used to explicitly model the input uncertainty
in the GP regression setting. 
The assumption made is that the observed inputs $\bfZ$ are 
obtained by the noise-free latent inputs $\latentMatrix$ by adding 
Gaussian noise, 
\begin{equation}
\label{uncertainInputsX}
 \bfzi_{i, :}  = \latentVector_{i, :}  + \noiseVector_x,
\end{equation}
where $\bfzi_{i, :}$ denotes the $i$-th observed input of the dataset
$\bfZ$ and $\noiseVector_z \sim
\gaussianSamp{\zerosVector}{\bfSigma_z}$, as in
\citep{mchutchon:gaussian}. 
Since $\bfZ$ is observed and $\latentMatrix$ unobserved the above equation essentially 
induces a Gaussian prior distribution over $\latentMatrix$ that has the form, 
\begin{equation}
 \label{uncertainInputsPX}
p(\latentMatrix | \bfZ) = \prod_{i=1}^\numData \gaussianDist{\latentVector_{i, :}}{\bfzi_{i, :}}{\bfSigma_z},
\end{equation}
where $\bfSigma_z$ is typically an unknown parameter. 
Given that  $\latentMatrix$ are really the inputs that eventually are passed through the GP latent function 
(to subsequently generate the outputs) the whole probabilistic model 
becomes a GP-LVM with the above special form for the prior distribution over the latent inputs,
making thus our variational framework easily applicable. 
More precisely, using the above prior, we can define a variational bound on $p(\dataMatrix)$ as well
as an associated approximation $q(\latentMatrix)$ to the true posterior $p(\latentMatrix|\dataMatrix, \bfZ)$. 
This variational distribution $q(\latentMatrix)$ can be used as a probability estimate of the noisy input locations $\latentMatrix$. During optimisation of the lower bound we can also learn the parameter 
$\bfSigma_z$. Furthermore, if we wish to reduce the number of parameters in the 
variational distribution $q(\latentMatrix) = \gaussianSamp{\mathcal{M}}{\mathcal{S}}$
a sensible choice would be to set $\mathcal{M} = \bfZ$, although such a choice 
may not be optimal. 

Having a method which implicitly models the
uncertainty in the inputs also allows for doing predictions in an autoregressive manner while
propagating the uncertainty through the predictive sequence \citep{Girard:uncertain01}.
To demonstrate this in the context of our framework, we will take the simple case where the
process of interest is a multivariate time-series given as pairs of 
time points $\bft = \{t\}_{i=1}^\numData$ and
corresponding output locations $\dataMatrix = \{\dataVector_{i, :}\}_{i=1}^\numData$, $\dataVector_{i, :} \in \Re^\dataDim$.
Here, we take the time locations to be
deterministic and equally spaced, so that they can be simply denoted by the subscript of 
the output points $\dataVector_{i, :}$; we thus simply denote with $\dataVector_k$ the output point 
$\dataVector_{k, :}$ which corresponds to $t_k$. 

We can now reformat the given data $\dataMatrix$ into input-output pairs 
$\hat{\bfZ}$ and $\hat{\dataMatrix}$,
where:
\begin{align*}
[\hat{\bfzi}_1, \hat{\bfzi}_2, ..., \hat{\bfzi}_{\numData-\tau}] &= 
\left[ \left[\dataVector_1, \dataVector_2, ..., \dataVector_\tau \right], 
	\left[\dataVector_2, \dataVector_3, ..., \dataVector_{\tau+1}\right], ..., 
	\left[\dataVector_{\numData-\tau}, \dataVector_{\numData-\tau+1}, ..., \dataVector_{\numData-1}\right]	\right], \\ 
[\hat{\dataVector}_1, \hat{\dataVector}_2, ..., \hat{\dataVector}_{\numData-\tau}] &= 
	[\dataVector_{\tau+1}, \dataVector_{\tau+2}, ..., \dataVector_\numData] 
\end{align*}
and $\tau$ is the size of the dynamics' ``memory''.
In other words, we define a window of size $\tau$ which shifts in time so that the output in time $t$ becomes an input
in time $t+1$. Therefore, 
the uncertain inputs method described earlier in this section
can be applied to the new dataset $[\hat{\bfZ}, \hat{\dataMatrix}]$. In particular, although the \emph{training} inputs
$\hat{\bfZ}$ are not necessarily uncertain in this case, the aforementioned way of performing inference
is particularly advantageous when the task is extrapolation. 

In more detail, consider the simplest case described in this section where the posterior
$q(\latentMatrix)$ is centered in the given noisy inputs and we allow for variable noise around the
centers. To perform extrapolation one firstly needs to train the model on the dataset
$[\hat{\bfZ}, \hat{\dataMatrix}]$. Then, we can perform iterative $k-$step ahead prediction in order to
find a future sequence $\left[\dataVector_{\numData+1}, \dataVector_{\numData+2}, ... \right]$ where,
similarly to the approach taken by \cite{Girard:uncertain01}, the predictive variance in each
step is accounted for and propagated in the subsequent predictions.
For example, if $k=1$ the algorithm will make iterative 1-step predictions in the future;
in the beginning, the output $\dataVector_{\numData+1}$ will be predicted given the training set. In the next step,
the training set will be augmented to include the previously predicted $\dataVector_{\numData+1}$ as part
of the input set, where the predictive variance is now encoded as the uncertainty of this point.

The advantage of the above method, which resembles a state-space model, 
is that the future predictions do not almost immediately revert to
the mean, as in standard stationary GP regression, neither do they underestimate the uncertainty,
as would happen if the predictive variance was not propagated through the inputs in a principled way.

%
%

\subsubsection{Demonstration: iterative $k-$step ahead forecasting\label{sec:experimentsKstep}}
Here we demonstrate our framework in the simulation of a state space model, as was
described previously. More specifically, we consider the 
Mackey-Glass chaotic time series, a standard benchmark which was also considered in
\citep{Girard:uncertain01}. The data is one-dimensional so that the timeseries can be represented as pairs of values $\{\bfy, \bft\}, t = 1,2, \cdots, \numData$ and simulates:
$$
\frac{\intd \zeta(t)}{dt} = 
-b\zeta(t) \alpha \frac{\zeta(t-T)}{1+\zeta(t-T)^{10}}, 
\;\; \text{with} \;\;  \alpha = 0.2, b=0.1, T=17.
$$
As can be seen, the generating process is very non-linear, something which makes this dataset particularly challenging.

The model trained on this dataset was the one described previously, where the modified dataset 
$\{ \hat{\bfy}, \hat{\bfz} \}$ was created with $\tau = 16$ and we 
used the first $96$ points to train the model and predicted the subsequent $180$ points in the future. The comparison was made firstly with a standard GP model (which we refer to as $\mathcal{GP}_{\bft, \bfy}$), where the input - output pairs were given in the standard form, that is, $\bft$ and $\bfy$ respectively and the predictions were made in the standard way, that is, given $\bft_*$. Further, we compared with a standard GP model where the input - output pairs were given by the modified dataset $\{\hat{\bfz}, \hat{\bfy}\}$ that was mentioned previously; this model is here referred to as 
$\mathcal{GP}_{\hat{\bfz}, \hat{\bfy}}$. For the latter model, the predictions are made in the $k-$step ahead manner, according to which the predicted values for iteration $k$ are added to the training set. However, this standard GP model has no straight forward way of propagating the uncertainty, and therefore the input uncertainty is zero for every
step of the iterative predictions.

\begin{figure}[ht]
\begin{center}
\includegraphics[width=0.6\textwidth]{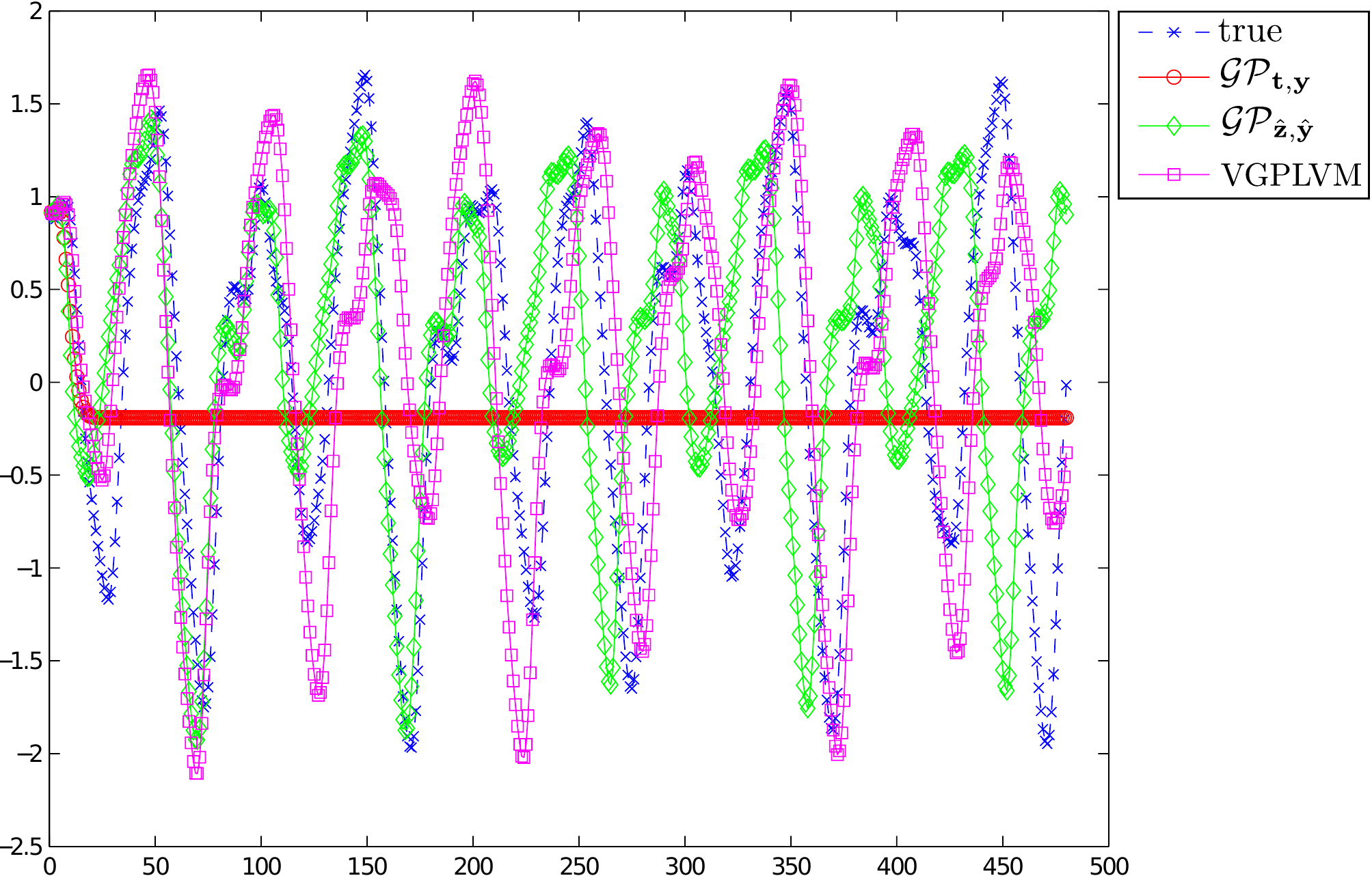}
\end{center}
\caption{\small{
Iterative $1-$step ahead prediction for a chaotic timeseries. 
Comparing a standard GP approach ($\mathcal{GP}_{\bft,\bfy}$),
an ``autoregressive'' GP approach which does not propagate uncertainties 
($\mathcal{GP}_{\hat{\bfz},\hat{\bfy}}$) and the variational GP-LVM (VGPLVM) in an
``autoregressive'' setting.
}
}
\label{fig:uncInputsExtrapolation}
\end{figure}

The predictions obtained can be seen in Figure \ref{fig:uncInputsExtrapolation}.
As can be seen, the variational GP-LVM is more robust in handling the uncertainty throughout the predictions something which results in lower predictive error. In particular, notice that in the first few predictions all methods give the same answer. However, the standard GP regression model, $\mathcal{GP}_{\bft, \bfy}$, very quickly reverts to the mean, as expected, when the test inputs are too far from the training ones and the uncertainty is very large. On the other hand, $\mathcal{GP}_{\hat{\bfz}, \hat{\bfy}}$ has the opposite problem; every predictive step results in a prediction which underestimates the uncertainty; therefore, although the initial predictions are reasonable, once they diverge a little by the true values the error is carried on and amplified.




\subsection{Semi-supervised GP Regression and Data Imputation\label{sec:semiSupervisedExtension}}

In this section, we describe how our proposed model can be used in a data imputation / semi-supervised regression problem where part of the training inputs are missing. This scenario is obviously a special case of the uncertain input modeling discussed above. Although a more general setting can be defined, here we consider the case where we have a fully and a partially observed set of inputs, \ie $\bfZ = (\bfZ^\observedSet, \bfZ^\unobservedSet)$, where $\observedSet$ and $\unobservedSet$ denote set of rows of $(\bfZ, \dataMatrix)$ that contain fully and partially observed inputs respectively\footnote{
In section \ref{section:predictions}, the superscript $\unobservedSet$ denoted the set of missing \emph{columns from test outputs}. Here it refers to \emph{rows of training inputs} that are \emph{partially} observed, \ie the union of $\observedSet$ and $\unobservedSet$ is now
$\{ 1, \cdots, \numData \}$.
}.
This is a realistic scenario; it is often the case that certain input features are more difficult to obtain (\eg human specified tags) than others, but we would nevertheless wish to model all available information within the same model. The features missing in $\bfZ^\unobservedSet$ can be different in number / location for each individual point $\bfz^\unobservedSet_{\dataIndex,:}$.

A standard GP regression model cannot straightforwardly model jointly $\bfZ^\observedSet$ and $\bfZ^\unobservedSet$. In contrast, in our framework the inputs are replaced by distributions $q(\latentMatrix^\observedSet)$ and $q(\latentMatrix^\unobservedSet)$, so that $\bfZ^\unobservedSet$ can be taken into account naturally by simply initialising the uncertainty of $q(\latentMatrix^\unobservedSet)$ in the missing locations to 1 (assuming normalized inputs) and the mean to the empirical mean and then, optionally, optimising $q(\latentMatrix^\unobservedSet)$.
In our experiments we use a slightly more sophisticated approach which resulted in better results. Specifically, we can use the fully observed data subset $(\bfZ^\observedSet, \dataMatrix^\observedSet)$ to train an initial model for which we fix $q(\latentMatrix^\observedSet) = \gaussianDist{\latentMatrix^\observedSet}{\bfZ^\observedSet}{\bfzero}$. Given this model, we can then use $\dataMatrix^\unobservedSet$ to estimate the predictive posterior $q(\latentMatrix^\unobservedSet$) in the missing locations of $\bfZ^\unobservedSet$ (for the observed locations we match the mean with the observations, as for $\bfZ^\observedSet$). After initializing $q(\latentMatrix)=q(\latentMatrix^\observedSet, \latentMatrix^\unobservedSet)$ in this way, we can proceed by training our model on the full (extended) training set
$\left( \left( \bfZ^\observedSet, \bfZ^\unobservedSet \right),
        \left( \dataMatrix^\observedSet, \dataMatrix^\unobservedSet \right)
\right)$,
which contains fully and partially observed inputs. During this training phase, the variational distribution $q(\latentMatrix)$ is held fixed in the locations corresponding to observed values and is optimised in the locations of missing inputs. 

Given the above formulation, we can define a \emph{semi-supervised GP} model which naturally incorporates fully and partially observed examples by communicating the uncertainty throughout the relevant parts of the model in a principled way. In specific, the predictive uncertainty obtained by the initial model trained on the fully observed data can be incorporated as input uncertainty via $q(\latentMatrix^\unobservedSet)$ in the model trained on the extended dataset, similarly to how extrapolation was achieved for our auto-regressive approach in Section \ref{uncertainInputs}. In extreme cases resulting in very non-confident predictions, for example presence of outliers, the corresponding locations will simply be ignored automatically due to the large uncertainty. This mechanism, together with the subsequent optimisation of $q(\latentMatrix^\unobservedSet)$, guards against reinforcing bad predictions when imputing missing values based on a smaller training set. 
In particular, in the limit of having no observed values the semi-supervised GP is equivalent to the GP-LVM and when there are no missing values (or when all missing locations have uncertainty $1$) it is equivalent to GP regression. Details of the algorithm for this approach are given in Appendix \ref{app:semiSupervised}.

The algorithm defined above can be seen as a particular instance of semi-supervised learning
which uses self-training for initialisation. Traditionally, semi-supervised settings are encountered in classification problems where only part of the training data are associated with known class labels. A simple approach to exploiting the unlabelled examples is to use self-training \citep{Rosenberg:semiSupervisedSelfTraining}, according to which an initial model is trained on the labelled examples and then used to incorporate the unlabelled examples in the manner dictated by the specific self-training methodology followed. In a \emph{bootstrap}-based self-training approach this incoroporation is achieved by predicting the missing labels using the initial model and, subsequently, augmenting the training set using only the confident predictions subset. Recently, \cite{Kingma14:semiSupervisedDeepGenerative} demonstrated the applicability of generative models in semi-supervised learning. Their method defines a latent space $\latentMatrix$ and estimates an approximate and factorised with respect to data points posterior $q(\latentMatrix | \bfZ)$ using labelled and unlabelled examples. Subsequently, the algorithm builds a classifier from the latent to the label space by sampling from areas of the approximate posterior that correspond to labelled instances.

While our framework can be adapted to tackle the aforementioned classification scenario, this is redirected to future work. Instead, here we focused on a regression problem where the missing values appear in the \emph{inputs}. However, there exist some similarities with the work referenced in the previous paragraph. In specific, our generative method treats the semi-supervised task as a data imputation problem, similarly to \citep{Kingma14:semiSupervisedDeepGenerative}. One of the differences with their work is that we do not use a latent space representation for the inputs but, instead, we directly associate the input space with uncertainty. Concerning relations with other methods which use self-training, our algorithm also trains an initial model on the fully observed portion of the data and predicts the missing values. However, these predictions only constitute initialisations which are later optimised along with model parameters and, hence, we refer to this step as \emph{partial} self-training. Further, in our framework the predictive uncertainty is not used as a hard measure of discarding unconfident predictions but, instead, we allow all values to contribute according to an optimised uncertainty measure. Therefore, the way in which uncertainty is handled makes the self-training part of our algorithm principled compared to many bootstrap-based approaches.

\subsubsection{Demonstration\label{sec:semiSupervisedExperiments}}
\begin{figure}[ht]
\begin{center}
    \includegraphics[width=1\textwidth]{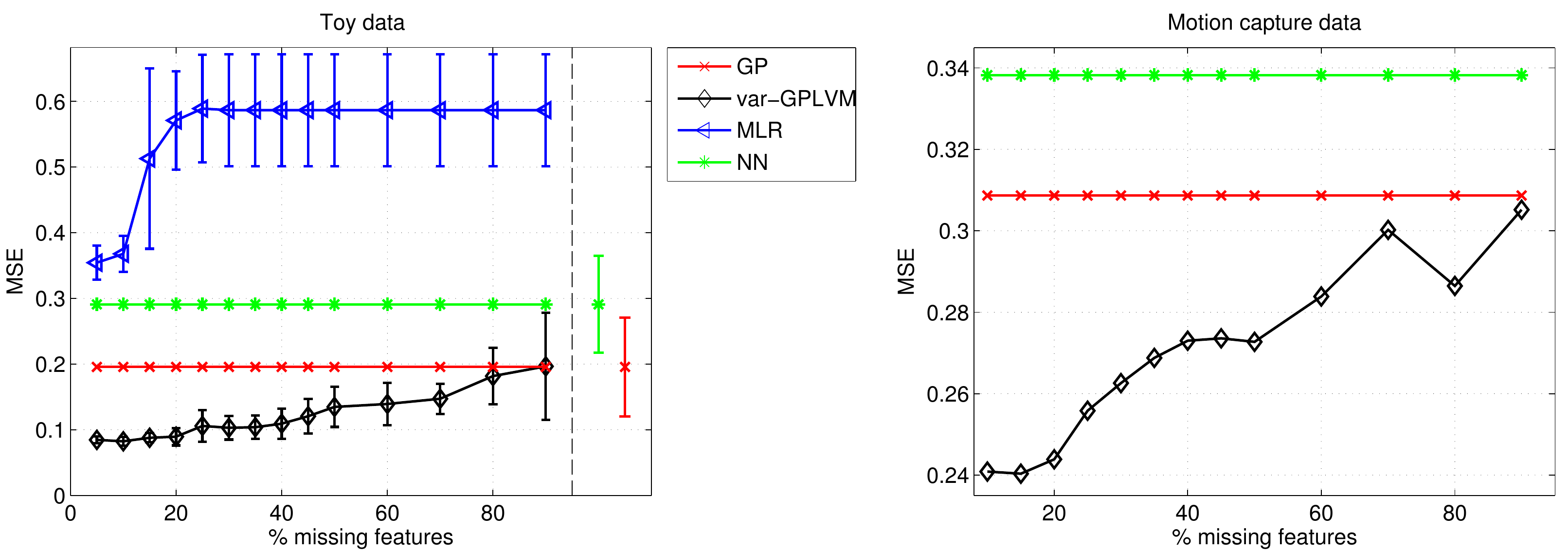}
\end{center}
\caption{\small{
Mean Squared Error for predictions obtained by different methods in simulated (left) and motion capture data (right). Flat line errors correspond to methods that cannot take into account partially observed inputs. The results for simulated data are obtained from 4 trials and, hence, errorbars are also plotted. For GP and NN, errorbars do not change with $x$-axis and, for clarity, they are plotted separately on the right of the dashed vertical line (for nonsensical $x$ values). Methods that resulted in very high MSE compared to the rest are not shown, for clearer plots; specifically, MSE for predicting with the data mean was $0.82$. For the motion capture data, MLR performed very badly, with average MSE $0.72$.
}
}
\label{fig:semiSupervised}
\end{figure}

In this section we consider simulated and real-world data to demonstrate our semi-supervised GP algorithm, which was discussed in Section \ref{sec:semiSupervisedExtension}. The simulated data were created by sampling inputs $\bfZ$ from an unknown to the competing models GP and gave this as input to another (again, unknown) GP to obtain the corresponding outputs $\dataMatrix$. For the real-world data demonstration we 
considered a subset of the same motion capture dataset discussed in Section \ref{sec:experimentsMocap}, which corresponds to a walking motion of a human body represented as a set of $59$ joint locations. We formulated a regression problem where the first $20$ dimensions of the original data are used as targets and the rest $39$ as inputs. In other words, given a partial joint representation of the human body, the task is to infer the rest of the representation. For both datasets, simulated and motion capture, we selected a portion of the training inputs, denoted as $\bfZ^\unobservedSet$, to have randomly missing features. 
The extended dataset $\left( \left(\bfZ^\observedSet, \bfZ^\unobservedSet \right),\left(\dataMatrix^\observedSet, \dataMatrix^\unobservedSet \right) \right)$ was used to train our method as well as multiple linear regression (MLR). Using only the observed data $\left(\bfZ^\observedSet, \dataMatrix^\observedSet \right)$ we trained a standard GP and nearest neighbour (NN), both of which cannot handle missing inputs straightforwardly. 
The goal was to reconstruct test outputs $\dataMatrix_*$ given fully observed test inputs $\bfZ_*$. For the simulated data we used the following sizes: $| \bfZ^\observedSet | = 40$, $| \bfZ^\unobservedSet | = 60$ and $| \bfZ_* | = 100 $. The dimensionality of the inputs is $15$ and of the outputs is $5$. For the motion capture data we used $| \bfZ^\observedSet | = 50$, $| \bfZ^\unobservedSet | = 80$ and $| \bfZ_* | = 200 $. In Figure \ref{fig:semiSupervised} we plot the MSE obtained by the competing methods for a varying percentage of missing features in $\bfZ^\unobservedSet$. For the simulated data experiment, each of the points in the plot is an average of 4 runs which considered different random seeds.
As can be seen, the semi-supervised GP is able to handle the extra data and make better predictions, even if a very large portion is missing. Indeed, its performance starts to converge to that of a standard GP when there are 90\% missing values in $\bfZ^\unobservedSet$ and performs identically to the standard GP when 100\% of the values are missing.

%% file: conclusions.tex

\section{\label{section:conclusion} Conclusion}

We have introduced an approximation to the marginal likelihood of the Gaussian process latent variable model
in the form of a variational lower bound. This provides a Bayesian training procedure which is robust
to overfitting and allows for the appropriate dimensionality of the latent space to be automatically determined.
Our framework is extended for the case where the observed data constitute multivariate timeseries and,
therefore, we obtain a very generic method for dynamical systems modelling able to capture complex, non-linear
correlations. We demonstrated the advantages of the rigorous lower bound defined in our framework on a range of
disparate real world data sets. This also emphasised the ability of the model to handle vast dimensionalities.

\par Our approach was easily extended to be applied to training Gaussian processes with uncertain inputs where these
inputs have Gaussian prior densities. This gave rise to an auto-regressive and a semi-supervised GP variant of our model.
 For future research, we envisage several other extensions that become computationally
feasible using the same set of methodologies we espouse. 
%
In particular, propagation of uncertain inputs through the Gaussian process allows Bayes filtering 
\citep{GP-Based,deisenroth2012robust,frigola2014identification} applications to be carried out through variational bounds. Bayes filters are non-linear dynamical systems where time is discrete and the observed data $\dataVector_t$ at time point $t$, is non-linearly related to some unobserved latent state, $\latentVector_t$,
\[
\dataVector_t = f(\latentVector_t)
\] 
which itself has a non-linear autoregressive relationship with past latent states:
\[
\latentVector_t = g(\latentVector_{t-1})
\]
where both $g(\cdot)$ and $f(\cdot)$ are assumed to be Gaussian processes. Propagation of the uncertainty through both processes can be achieved through our variational lower bound allowing fast efficient approximations to Gaussian process dynamical models.

The bound also allows for a promising new direction of research, that of \emph{deep Gaussian processes}. In a deep Gaussian process \citep{Lawrence:hgplvm07,Damianou:deepgp13} the idea of placing a temporal prior over the inputs to a GP is further extended by hierarchical application. This formalism leads to a powerful class of models where Gaussian process priors are placed over function compositions, for example in a five layer model we have
\[
f(\latentMatrix) = g_5(g_4(g_3(g_2(g_1(\latentMatrix)))))
\]
where each $g_i(\cdot)$ is a draw from a Gaussian process. By combining such models with structure learning \citep{Damianou:manifold12} we can develop the potential to learn very complex non linear interactions between data. In contrast to other deep models \emph{all} the uncertainty in parameters and latent variables is marginalised out.


%% file: supplementary.tex

\appendix
\section{Further Details About the Variational Bound \label{appendix:bound}}
This appendix contains supplementary details for deriving some mathematical formulae
related to the calculation of the final expression of the variational lower bound for the training phase.

Since many derivations require completing the square to recognize a Gaussian, 
we will use the following notation throughout the Appendix:
$$
\mathcal{Z} = \text{the collection of all constants for the specific line in equation},
$$
where the definition of a constant depends on the derivation at hand.

\subsection{Calculation of: $\left\langle \log p(\dataVector_{:, \outputIndex} | \mappingFunctionVector_{:, \outputIndex}) 
    \right\rangle_{p(\mappingFunctionVector_{:, \outputIndex} | \inducingVector_{:, \outputIndex}, \latentMatrix)} 
$}
First, we show in detail how to obtain the r.h.s of equation \eqref{eq:expectYgivenF} for
the following quantity: \\
$\left\langle \log p(\dataVector_{:, \outputIndex} | \mappingFunctionVector_{:, \outputIndex}) 
    \right\rangle_{p(\mappingFunctionVector_{:, \outputIndex} | \inducingVector_{:, \outputIndex}, \latentMatrix)} 
$ which appears in the variational bound of equation \eqref{Fd}. Notice that this term itself lower bounds
the logarithm of the likelihood:
\begin{align}
\log p(\dataVector_{:, \outputIndex} | \latentMatrix) 
&= 
\log \int
p(\dataVector_{:, \outputIndex} | \mappingFunctionVector_{:, \outputIndex})
p(\mappingFunctionVector_{:, \outputIndex} | \inducingVector_{:, \outputIndex}, \latentMatrix) 
  \intd \mappingFunctionVector_{:, \outputIndex} \intd \inducingVector_{:, \outputIndex} \nonumber \\
  & \ge 
   \int
p(\mappingFunctionVector_{:, \outputIndex} | \inducingVector_{:, \outputIndex}, \latentMatrix) 
  \log
p(\dataVector_{:, \outputIndex} | \mappingFunctionVector_{:, \outputIndex})
  \intd \mappingFunctionVector_{:, \outputIndex} \intd \inducingVector_{:, \outputIndex},
\end{align}
where the last line follows from Jensen's inequality. 
In the following, we compute the above quantity analytically while temporarily using the notation $ \la \cdot \ra =
\la \cdot \ra_{p(\mappingFunctionVector_{:,\outputIndex} | \inducingVector_{:,\outputIndex}, \latentMatrix)}$ :
\begin{align}
\la \log p(\dataVector_{:,\outputIndex} | \mappingFunctionVector_{:,\outputIndex}) \ra \stackrel{\text{eq. } \eqref{eq:yGivenF}}{=} {}&
  \left\langle \log \gaussianDist{\dataVector_{:,\outputIndex}}{\mappingFunctionVector_{:,\outputIndex}}{\dataStd^2 \eye_\dataDim} \right\rangle \nonumber \\
={}&  - \frac{\numData}{2} \log(2 \pi) - \frac{1}{2} \log \vert \dataStd^2 \eye_\dataDim \vert 
 - \frac{1}{2}
  \tr{ \dataStd^{-2} \eye_\dataDim \left( \dataVector_{:,\outputIndex} \dataVector_{:,\outputIndex}^\T - 
    2\dataVector_{:,\outputIndex} \la \mappingFunctionVector_{:,\outputIndex}^\T \ra + \la \mappingFunctionVector_{:,\outputIndex} \mappingFunctionVector_{:,\outputIndex}^\T \ra \right) } \nonumber \\
\stackrel{\text{eq. } \eqref{priorF2}}{=}{}& 
\mathcal{Z}
- \frac{1}{2} \tr{ \dataStd^{-2} \eye_\dataDim \left( \dataVector_{:,\outputIndex} \dataVector_{:,\outputIndex}^\T - 2 \dataVector_{:,\outputIndex} \bfa_\outputIndex^\T 
    + \bfa_\outputIndex \bfa_\outputIndex^\T + \bfSigma_f \right) } \nonumber . 
\end{align}

\noindent By completing the square we find:

\begin{align}
\la \log p(\dataVector_{:,\outputIndex} | \mappingFunctionVector_{:,\outputIndex}) \ra_{p(\mappingFunctionVector_{:,\outputIndex} | \inducingVector_{:,\outputIndex}, \latentMatrix)} 
={}& \log \gaussianDist{\dataVector_{:,\outputIndex}}{\bfa_\outputIndex}{\dataStd^2 \eye_\dataDim } 
    - \frac{1}{2} \tr{ \dataStd^{-2} \bfSigma_f } \nonumber \\
\stackrel{\text{eq. } \eqref{eq:conditionalGPmeanCovar}}{=}{}& \log \gaussianDist{\dataVector_{:,\outputIndex}}{\bfa_\outputIndex}{\dataStd^2 \eye_\dataDim }
  - \frac{1}{2\dataStd^{2}} \tr{ \Kff - \Kfu \Kuu^{-1} \Kuf } . \label{boundFAnalytically3}
\end{align}

\subsection{Calculating the Explicit Form of $q(\inducingVector_{:,j})$\label{app:qu}}
From equation \eqref{qu}, we have:
\begin{equation}
\label{eq:log_q_u}
\log q(\inducingVector_{:, \outputIndex}) \propto \la \log \gaussianDist{\dataVector_{:, \outputIndex}}{\bfa_\outputIndex}{ \dataStd^2 \eye_\dataDim} \ra_{q(\latentMatrix)} + \log p(\inducingVector_{:, \outputIndex}) .
\end{equation}
All the involved distributions are Gaussian and, hence, we only need to compute the r.h.s of the above equation and
complete the square in order to get the posterior Gaussian distribution for $q(\inducingVector_{:,j})$.
The expectation appearing in the above equation is easily computed as:
\begin{align}
\la \log \gaussianDist{\dataVector_{:,\outputIndex}}{\bfa_\outputIndex}{\dataStd^2 \eye_\dataDim} \ra_{q(\latentMatrix)} = & 
\mathcal{Z}
- \frac{1}{2 \dataStd^2} \tr{  \dataVector_{:,\outputIndex} \dataVector_{:,\outputIndex}^\T - 
2 \dataVector_{:,\outputIndex} \la \bfa_\outputIndex^\T \ra_{q(\latentMatrix)} + \la \bfa_\outputIndex \bfa_\outputIndex^\T \ra_{q(\latentMatrix)}  } \nonumber \\
\stackrel{\text{eq. } \eqref{eq:conditionalGPmeanCovar}}{=} &
\mathcal{Z}
- \frac{1}{2 \dataStd^2} \text{tr} \Big(
   \dataVector_{:,\outputIndex} \dataVector_{:,\outputIndex}^\T - 2 \dataVector_{:,\outputIndex} \inducingVector_{:,\outputIndex}^\T \Kuu^{-1} \la \Kfu^\T \ra_{q(\latentMatrix)} \nonumber \\
& \quad \; \; \qquad + \inducingVector_{:,\outputIndex}^\T \Kuu^{-1} \la \Kfu^\T \Kfu \ra_{q(\latentMatrix)} \Kuu^{-1} \inducingVector_{:,\outputIndex}  \Big) \nonumber \\
\stackrel{\text{eq. } \eqref{psis}}{=} &
\mathcal{Z}
- \frac{1}{2 \dataStd^2} \text{tr} \Big(
   \dataVector_{:,\outputIndex} \dataVector_{:,\outputIndex}^\T - 2 \dataVector_{:,\outputIndex} \inducingVector_{:,\outputIndex}^\T \Kuu^{-1} \bfPsi_1^\T \nonumber \\
& \quad \; \; \qquad   +  \inducingVector_{:,\outputIndex}^\T \Kuu^{-1} \bfPsi_2 \Kuu^{-1} \inducingVector_{:,\outputIndex}  \Big) \label{eq:q_u_part1} .
\end{align}

We can now easily find equation \eqref{eq:log_q_u} by combining equations \eqref{eq:q_u_part1} and \eqref{pfu}:
\begin{align}
\log q(\inducingVector_{:, \outputIndex}) 
& \propto \la \log \gaussianDist{\dataVector_{:, \outputIndex}}{\bfa_\outputIndex}{ \dataStd^2 \eye_\dataDim} \ra_{q(\latentMatrix)} + \log p(\inducingVector_{:, \outputIndex}) \nonumber \\
& = 
\mathcal{Z}
- \frac{1}{2 \dataStd^2} \text{tr} \Big(
   \dataVector_{:,\outputIndex} \dataVector_{:,\outputIndex}^\T - 2 \dataVector_{:,\outputIndex} \inducingVector_{:,\outputIndex}^\T \Kuu^{-1} \bfPsi_1^\T + 
  \inducingVector_{:,\outputIndex}^\T \Kuu^{-1} \bfPsi_2 \Kuu^{-1} \inducingVector_{:,\outputIndex}  \Big) \nonumber \\
& \qquad \; \; \; \qquad -  \frac{1}{2} \text{tr} \Big( \Kuu^{-1} \inducingVector_{:,\outputIndex} \inducingVector_{:,\outputIndex}^\T \Big) \nonumber \\
& = 
\mathcal{Z}
 - \frac{1}{2} \text{tr} \Big( \inducingVector_{:,\outputIndex}^\T \left( \dataStd^{-2} \Kuu^{-1} \bfPsi_2 \Kuu^{-1} + \Kuu^{-1} \right) \inducingVector_{:,\outputIndex}
 + \dataStd^{-2} \dataVector_{:,\outputIndex} \dataVector_{:,\outputIndex}^\T 
  - 2 \dataStd^{-2} \Kuu^{-1} \bfPsi_1^\T \dataVector_{:,\outputIndex} \inducingVector_{:,\outputIndex}^\T \Big) . \label{eq:boundFIntegral2}
\end{align}

\noindent We can now complete the square again and recognize that $q(\inducingVector_{:,\outputIndex}) = \gaussianDist{\inducingVector_{:,\outputIndex}}{\bfmu_\inducingScalar}{\bfSigma_\inducingScalar}$, where:
\begin{align*}
\bfSigma_\inducingScalar = {}& \left( \dataStd^{-2} \Kuu^{-1} \bfPsi_2 \Kuu^{-1} + \Kuu^{-1} \right)^{-1} \mbox{\;\;\;\; and} \\
\bfmu_\inducingScalar = {}& \dataStd^{-2} \bfSigma_\inducingScalar \Kuu^{-1} \bfPsi_1^\T \dataVector_{:,\outputIndex} .
\end{align*}

\noindent By ``pulling'' the $\Kuu$ matrices out of the inverse and after simple manipulations 
we get the final form of $q(\inducingVector_{:,\outputIndex})$:
\begin{equation}
\begin{aligned}
\label{eq:qu}
 q(\inducingVector_{:,\outputIndex}) = {}& \gaussianDist{\inducingVector_{:,\outputIndex}}{\bfmu_\inducingScalar}{\bfSigma_\inducingScalar} \mbox{\;\; where}  \\
\bfmu_\inducingScalar = {}& \Kuu \left( \dataStd^2 \Kuu + \bfPsi_2 \right)^{-1} \bfPsi_1^\T \dataVector_{:,\outputIndex} \\
\bfSigma_\inducingScalar = {}& \dataStd^2 \Kuu \left( \dataStd^2 \Kuu + \bfPsi_2 \right)^{-1} \Kuu .
\end{aligned}
\end{equation}

\subsection{Detailed Derivation of $\hat{\F}_\outputIndex(q(\latentMatrix))$ \label{appendix:bound3}}
The quantity $\hat{\F}_\outputIndex(q(\latentMatrix))$ appears in equation \eqref{boundFAnalyticallyFinalIntegral}.
Based on the derivations of the previous section, we can rewrite equation \eqref{eq:boundFIntegral2} as a function of the optimal 
$q(\inducingVector_{:,\outputIndex})$ found in equation \eqref{eq:qu} by completing the constant terms:
\begin{align}
\la \log \gaussianDist{\dataVector_{:, \outputIndex}}{\bfa_\outputIndex}{ \dataStd^2 \eye_\dataDim} \ra_{q(\latentMatrix)} + \log p(\inducingVector_{:, \outputIndex}) 
&= \mathcal{B} + \log \gaussianDist{\bfu_d}{\bfmu_\inducingScalar}{\bfSigma_\inducingScalar} \label{eq:boundFIntegral3}
\end{align}
where we have defined:
\begin{equation}
\label{eq:boundBterm}
\mathcal{B}=- \frac{\numData}{2} \log (2 \pi) - \frac{1}{2} \log \vert \dataStd^2 \eye_\dataDim \vert
  - \frac{1}{2} \log \vert \Kuu \vert 
 - \frac{1}{2\dataStd^2} \dataVector_{:,\outputIndex}^\T \dataVector_{:,\outputIndex} + \frac{1}{2} \bfmu_\inducingScalar^\T \bfSigma_\inducingScalar^{-1} \bfmu_\inducingScalar + \frac{1}{2} \log \vert \bfSigma_\inducingScalar \vert .
\end{equation}

We can now obtain the final expression for \eqref{boundFAnalyticallyFinalIntegral}
 by simply putting the quantity of \eqref{eq:boundFIntegral3} on the exponent and integrating. By doing so, we get:

\begin{align}
& \int e^{\la \log \gaussianDist{\dataVector_{:,\outputIndex}}{\bfa_d}{\dataStd^2 I_d} \ra_{q(\latentMatrix)}}
    p(\bfu_d) \intd \bfu_d = 
  \int e^{\mathcal{B}} e^{\log \gaussianDist{\bfu_d}{\bfmu_\inducingScalar}{\bfSigma_\inducingScalar}}  \intd\bfu_d = e^{\mathcal{B}}\nonumber \\
\stackrel{\text{eq. } \eqref{eq:boundBterm}}{=} {}& \;\;\;\;
  (2 \pi)^{-\frac{N}{2}} \dataStd^{-\numData} \vert \Kuu \vert^{-\frac{1}{2}}
  e^{- \frac{1}{2\dataStd^2} \dataVector_{:,\outputIndex}^\T \dataVector_{:,\outputIndex}} \vert \bfSigma_\inducingScalar \vert^{\frac{1}{2}} e^{\frac{1}{2}\bfmu_\inducingScalar^\T \bfSigma_\inducingScalar^{-1} \bfmu_\inducingScalar} \label{eq:boundIntegral4} .
\end{align}

By using equation \eqref{eq:qu} and some straightforward algebraic manipulations, we can replace in the above
 $\bfmu_\inducingScalar^\T \bfSigma_\inducingScalar^{-1} \bfmu_\inducingScalar$ with:
\begin{equation}
\label{eq:mTSm}
\bfmu_\inducingScalar^\T \bfSigma_\inducingScalar^{-1} \bfmu_\inducingScalar = \dataVector_{:,\outputIndex}^\T \underbrace{\dataStd^{-4} \bfPsi_1 (\dataStd^{-2} \bfPsi_2 + \Kuu)^{-1} \bfPsi_1^\T}_{\bfW'} \dataVector_{:,\outputIndex}.
\end{equation}

Finally, using equation \eqref{eq:qu} to replace $\bfSigma_\inducingScalar$ with its equal, as well as equation \eqref{eq:mTSm},
we can write the integral of equation \eqref{eq:boundIntegral4} as:
\begin{equation}
\label{eq:boundIntegralFinal}
\int e^{\la \log \gaussianDist{\dataVector_{:,\outputIndex}}{\bfa_d}{\dataStd^2 I_d} \ra_{q(\latentMatrix)}}
    p(\bfu_d) \intd \bfu_d = 
    \frac{\dataStd^{-\numData} \vert \Kuu \vert^{-\frac{1}{2}} \vert \Kuu \vert e^{-\frac{1}{2\dataStd^2} \dataVector_{:,\outputIndex}^\T \dataVector_{:,\outputIndex}}}
   {(2 \pi)^{N/2}  \vert \dataStd^{-2} \bfPsi_2 + \Kuu \vert^{\frac{1}{2}} }
   e^{\frac{1}{2} \dataVector_{:,\outputIndex}^\T \bfW' \dataVector_{:,\outputIndex}} .
\end{equation}

We can now obtain the final form for the variational bound by replacing
equation \eqref{eq:boundIntegralFinal} in equation \eqref{boundFAnalyticallyFinalIntegral}, 
as well as replacing the term $\mathcal{A}$ with its equal and
defining $\bfW = \dataStd^{-2} \eye_\numData - \bfW'$.
By doing the above, we get exactly the final form of the bound of equation \eqref{FdFinal}.

\section{Calculating the $\Psi$ Quantities \label{PsiQuantities}}
Here we explain how one can compute the $\Psi$ quantities
(introduced in Section \ref{sec:lowerBoundWithAuxVars})
for two standard choices for the GP prior covariance. For completeness, we start
by rewriting the equations \eqref{eq:psi0}, \eqref{eq:psi1} and \eqref{eq:psi2}:

\begin{equation}
\psi_0 = \sum_{i=1}^\numData \psi_0^i, \; \; \text{with} \; \; 
\psi_0^i = \int k(\latentVector_{i,:},\latentVector_{i,:}) \mathcal{N}(\latentVector_{i,:} |\bfmu_{\dataIndex,:} , \bfS_i) 
  \intd \latentVector_{i,:}.
\end{equation}
\begin{equation}
  (\bfPsi_1)_{i,k} = \int k \left( \latentVector_{i,:},(\latentVector_u)_{k,:} \right) 
  \mathcal{N}(\latentVector_{i,:}|\bfmu_{\dataIndex,:}, \bfS_i) \intd
  \latentVector_{i,:}.
\end{equation}

\begin{equation}
\bfPsi_2 = \sum_{i=1}^\numData \bfPsi_2^i  \; \; \text{where } \; \;
(\Psi^i_2)_{k, k'} = \int k(\latentVector_{i,:},(\latentVector_u)_{k,:})
  k((\latentVector_u)_{k',:},\latentVector_{i,:}) 
  \mathcal{N}(\latentVector_{i,:}|\bfmu_{\dataIndex,:}, \bfS_i) \intd \latentVector_{i,:}.
\end{equation}
The above computations involve convolutions of the covariance function
with a Gaussian density. For some standard kernels such the ARD
exponentiated quadratic (RBF) covariance and the linear covariance function
these statistics are obtained analytically. In particular for the ARD
exponentiated quadratic kernel of equation \eqref{ard} we have:
\begin{align} 
\psi_0 & = \numData \sigma_f^2 \\
(\bfPsi_1)_{i,k} & = \sigma^2_f \prod_{j=1}^\latentDim
\frac{ \exp \left( - \frac{1}{2} \frac{ w_j (\mu_{i,j}  -
       (\latentScalar_u)_{k,j})^2}{w_j S_{i,j} + 1} \right)
    }
    {( w_j S_{i,j} + 1)^{\frac{1}{2}}
    } \\
(\Psi^i_2)_{k, k'} & = \sigma_f^4 
\prod_{j=1}^\latentDim 
    \frac{
         \exp \left( - \frac{w_j ((\latentScalar_u)_{k,j} -
    	(\latentScalar_u)_{k',j})^2}{4} - \frac{w_j \left(\mu_{i,j} -
 		\bar{x}_{:,j} \right)^2}{2 w_j S_{i,j} + 1} \right)
     }
     {(2 w_j S_{i,j} + 1)^{\frac{1}{2}}
     },
\end{align}
where $\bar{x}_{:,j} = \frac{((\latentScalar_u)_{k,j} + (\latentScalar_u)_{k',j})}{2}$. This gives us all
the components we need to compute the variational lower bound for the
ARD exponentiated quadratic kernel. 

For the linear covariance function (with ARD) the integrals
are also tractable, such that 
\begin{align}
  \psi_0^i & = \tr{\bfC (\bfmu_{\dataIndex,:} \bfmu_{\dataIndex,:}^\T + \bfS_i) } \\
  (\bfPsi_1)_{i,k} & = \bfmu_{\dataIndex,:}^\T \bfC (\latentVector_u)_{k,:} \\
  (\bfPsi_2^i)_{k,k'} & = 
  (\latentVector_u)_{k,:}^\top \bfC \left( \bfmu_{\dataIndex,:}\bfmu_{\dataIndex,:}^\T + \bfS_i \right)
   \bfC (\latentVector_u)_{k',:}.
\end{align}

\section{Derivatives of the Variational Bound for the Dynamical Version}
Before giving the expressions for the derivatives of the variational bound \eqref{jensens1},
it should be recalled that the variational parameters $\bfmu_j$ and $\bfS_j$ (for all $q$s) have been
reparametrised as 
\begin{equation*}
\bfS_\outputIndex = \left( \Kx^{-1} + \text{diag}(\boldsymbol \lambda_\outputIndex) \right)^{-1}  \text{ and }   
\bfmu_{:,\outputIndex} = \Kx \bar{\bfmu}_{:,\outputIndex}, 
\end{equation*}
where the function $\text{diag}(\cdot)$ transforms a vector into a square diagonal matrix and vice versa.
Given the above, the set of the parameters to be optimised is 
$( \bftheta_f, \bftheta_x, \{ \bar{\bfmu}_{:,\outputIndex}, \boldsymbol \lambda_\outputIndex \}_{\outputIndex=1}^\latentDim,
\tilde{\bfX})$. The gradient w.r.t the inducing points $\tilde{\bfX}$,
however, has exactly the same form as for $\bftheta_f$ and, therefore, is not
presented here.

\textbf{Some more notation:} 
\begin{enumerate}
\item $\lambda_j$ is a scalar, an element of the vector $\boldsymbol \lambda_j$ which, in turn, is
the main diagonal of the diagonal matrix $\bfLambda_j$. 
\item $(S_\latentIndex)_{k, l} \triangleq S_{\latentIndex;kl}$ the element of $\bfS_\latentIndex$ found in the $k$-th row and $l$-th column.
\item $\mathbf{s}_\latentIndex \triangleq \lbrace (S_{\latentIndex})_{i,i} \rbrace_{i=1}^\numData$, \ie it is a vector with the
diagonal of $\bfS_\latentIndex$.
\end{enumerate}

\subsection{Derivatives w.r.t the Variational Parameters}
\begin{equation}
    \label{derivVarParamSuppl}
\frac{\vartheta \mathcal{F}}{\vartheta \bar{\bfmu}_j} 
=  \Kx \left( \frac{\vartheta \hat{\mathcal{F}}}{\vartheta \bfmu_{:,j}} - 
    \bar{\bfmu}_{:,j} \right)
\text{ and }
 \frac{\vartheta \mathcal{F}}{\vartheta \boldsymbol \lambda_j}
= - ( \bfS_j \circ \bfS_j) \left( \frac{\vv \hat{\mathcal{F}}}{\vv \mathbf{s}_j} + 
	\frac{1}{2} \boldsymbol \lambda_j \right).
\end{equation}

where for each single dimensional element we have:

\begin{align}
 \frac{\hat{\mathcal{F}}}{\vartheta \mu_j}
{}& = - \frac{\dataDim}{\dataStd^2 2} \frac{\vartheta \psi_0}{\vartheta \mu_j}
    + \dataStd^{-2} \tr {\frac{\vartheta \bfPsi_1^\T}{\vartheta \mu_j} \bfY \bfY^\T \bfPsi_1 \bfA^{-1} } \nonumber \\
{}& + \frac{1}{2 \dataStd^2} \tr{ \frac{\vartheta \bfPsi_2}{\vartheta \mu_j}
       \left(
	  \dataDim \Kuu^{-1} - \dataStd^2 \dataDim \bfA^{-1} - \bfA^{-1} \bfPsi_1^\T \bfY \bfY^\T \bfPsi_1 \bfA^{-1}
       \right) } \label{derivFTildeEfficientComputationMu}
\end{align}

\begin{align}
 \frac{\vv \hat{\mathcal{F}}}{\vartheta (S_\latentIndex)_{k, l}}
{}& = - \frac{\dataDim}{2 \dataStd^2} \frac{\vartheta \Psi_0}{\vartheta (S_\latentIndex)_{k, l}}
    + \dataStd^{-2} \tr  { \frac{\vartheta \bfPsi_1^\T}
			    {\vartheta (S_\latentIndex)_{k, l}} \bfY \bfY^\T \bfPsi_1 \bfA^{-1}  } \nonumber \\
{}& + \frac{1}{2 \dataStd^2} \tr{\frac{\vartheta \bfPsi_2}{\vartheta (S_\latentIndex)_{k, l}}
       \left(
	  \dataDim \Kuu^{-1} - \dataStd^2 \dataDim \bfA^{-1} - \bfA^{-1} \bfPsi_1^\T \bfY \bfY^\T \bfPsi_1 \bfA^{-1}
       \right) } \label{derivFTildeEfficientComputationS}
\end{align}

with $\bfA=\dataStd^2 \Kuu + \bfPsi_2$.


\subsection{Derivatives w.r.t $\bftheta = (\bftheta_f, \bftheta_x)$ and $\beta = \dataStd^{-2}$}
In our implementation, we prefer to parametrise the software with 
the data precision $\beta$, rather than the data variance, $\dataStd^2$.
Therefore, here we will give directly the derivatives for the precision. Obviously, through the use
of the chain rule and the relationship $\dataStd^2 = \beta^{-1}$ one can obtain the derivatives for the variance.
Further, when it comes to model parameters, we will write the gradients with respect to each single element $\theta_f$
or $\theta_x$.

Given that the KL term involves only the temporal prior, its gradient w.r.t the parameters $\bftheta_f$ is zero. Therefore:
\begin{equation}
   \label{DerivativeOfFComplete}
      \frac{\vartheta \mathcal{F}}{\vartheta \theta_f} = 
      \frac{\vartheta \hat{\mathcal{F}}}{\vartheta \theta_f}
\end{equation}

  with:

\begin{align}
\frac{\vartheta \hat{\mathcal{F}}}{\vartheta \theta_f} {}& = \text{const} - 
\frac{\beta \dataDim}{2} \frac{\vartheta \psi_0}{\vartheta \theta_f}
 + \beta \tr{\frac{\vartheta \bfPsi_1^\T}{\vartheta \theta_f} \bfY \bfY^\T
   \bfPsi_1 \bfA^{-1} } \nonumber \\
{}& + \frac{1}{2} \tr{ \frac{\vartheta \Kuu}{\vartheta \theta_f}
        \left(
	   \dataDim \Kuu^{-1} - \dataStd^2 \dataDim \bfA^{-1} - \bfA^{-1} \bfPsi_1^\T \bfY \bfY^\T
	   \bfPsi_1 \bfA^{-1} - \beta \dataDim \Kuu^{-1} \bfPsi_2 \Kuu^{-1} 
         \right) } \nonumber \\
{}& + \frac{\beta}{2} \tr{ \frac{\vartheta \bfPsi_2}{\vartheta \theta_f} \;\;\;\;
       \left(
	  \dataDim \Kuu^{-1} - \dataStd^2 \dataDim \bfA^{-1} - 
	  \bfA^{-1} \bfPsi_1^\T \bfY \bfY^\T \bfPsi_1 \bfA^{-1}
       \right) } \label{DerivativeOfFtildeComplete}
\end{align}

The expression above is identical for the derivatives w.r.t the inducing points.
For the gradients w.r.t the $\beta$ term, we have a similar expression:

\begin{align}
\frac{\vartheta \hat{\mathcal{F}}}{\vartheta \beta} ={}&
  \frac{1}{2} \Big[ 
      \dataDim \left( \tr{\Kuu^{-1} \bfPsi_2} + (\numData-\numInducing)\dataStd^2 - \psi_0 \right) - \tr{\bfY \bfY^\T}
	  + \tr{\bfA^{-1} \bfPsi_1^\T \bfY \bfY^\T \bfPsi_1} \nonumber \\
   +{}& \beta^{-2} \dataDim \; \tr{ \Kuu \bfA^{-1} } + 
   \dataStd^2 \tr{ \Kuu^{-1} \bfA^{-1} \bfPsi_1^\T \bfY \bfY^\T 
	  \bfPsi_1 \bfA^{-1} } \Big] .
\label{derivb2}
\end{align}

In contrast to the above, the term $\hat{\mathcal{F}}$ does involve parameters $\bftheta_x$,
because it involves the variational parameters that are now reparametrised with $\Kx$,
which in turn depends on $\bftheta_x$. 
To demonstrate that, we will forget for a moment the reparametrisation of $\bfS_j$ and we will express the bound as $\mathcal{F}(\bftheta_x, \mu_j (\bftheta_x))$ (where $\mu_j (\bftheta_x) = K_t \bar{\bfmu}_{:,j}$) so as to show explicitly the dependency on the variational mean which is now a function of $\bftheta_x$. Our calculations must now take into account the term
$
\left( \frac{\vartheta \hat{\mathcal{F}}(\bfmu_{:,j})}{\vartheta \bfmu_{:,j}} \right)^\T
       \frac{\vartheta \mu_j (\bftheta_x)}{\vartheta \bftheta_x}
$
that is what we ``miss'' when we consider $\mu_j(\bftheta_x) = \bfmu_{:,j}$:
\begin{align}
\frac{\vartheta \mathcal{F}(\bftheta_x, \mu_j(\bftheta_x))}{\vartheta \theta_x} = {}&
	\frac{\vartheta \mathcal{F}(\bftheta_x, \bfmu_{:,j})}{\vartheta \theta_x} 
  +  \left( \frac{\vartheta \hat{\mathcal{F}}(\bfmu_{:,j})}{\vartheta \bfmu_{:,j}} \right)^\T
            \frac{\vartheta \mu_j(\bftheta_x)}{\vartheta \theta_x} \nonumber \\
= {}&
 \cancel{
    \frac{\vartheta \hat{\mathcal{F}}(\bfmu_{:,j})}{\vartheta \theta_x}
  } +
  \frac{\vv (-\text{KL})(\bftheta_x, \mu_j(\bftheta_x))}{\vartheta \theta_x}
+  \left( \frac{\vartheta \hat{\mathcal{F}}(\bfmu_{:,j})}{\vartheta \bfmu_{:,j}} \right)^\T
            \frac{\vartheta \mu_j(\bftheta_x)}{\vartheta \theta_x} .
\label{meanReparamDerivFTheta}
\end{align}

We do the same for $\bfS_j$ and then we can take the resulting equations and replace 
$\bfmu_j$ and $\bfS_j$ with their equals so as to take the final expression which only
contains $\bar{\bfmu}_{:,j}$ and $\boldsymbol \lambda_j$:

\begin{align}
\frac{\vartheta \mathcal{F}(\bftheta_x, \mu_j(\bftheta_x), 
	\bfS_j(\bftheta_x))}{\vartheta \theta_x}
={}& \text{tr} \bigg[
\Big[ - \frac{1}{2} \left( \hat{\bfB}_j \Kx \hat{\bfB}_j + \bar{\bfmu}_{:,j} \bar{\bfmu}_{:,j}^\T \right) \nonumber \\
+{}& \left( \bfI - \hat{\bfB}_j \Kx \right)
 diag \left(  \frac{\vv \hat{\mathcal{F}}}{\vv \mathbf{s}_j} \right)
			 \left( \bfI - \hat{\bfB}_j \Kx \right)^\T \Big]
			  \frac{\vv \Kx}{\vv \theta_x} \bigg] 	\nonumber \\	
+{}&  \left( \frac{\vartheta \hat{\mathcal{F}}( \bfmu_{:,j})}{\vartheta \bfmu_{:,j}} \right)^\T
					\frac{\vv \Kx}{\vv \theta_x} \bar{\bfmu}_{:,j} 
\label{CompleteBoundDerivThetatB}
\end{align}
where $\hat{\bfB}_j = \bfLambda_j^{\frac{1}{2}} \widetilde{\bfB}_j^{-1} \bfLambda_j^{\frac{1}{2}}$.
and $\tilde{\bfB}_j = \bfI + \bfLambda_j^{\frac{1}{2}} \Kx \bfLambda_j^{\frac{1}{2}}$. 
Note that by using this $\tilde{\bfB}_j$ matrix (which has eigenvalues bounded below by one)
we have an expression which, when implemented, leads to more numerically stable computations,
as explained in \cite{Rasmussen:book06} page 45-46.

\section{Variational Lower Bound for Partially Observed Test Data \label{app:partialtest}}

This section provides some more details related to the task of doing predictions based
on partially observed test data $\dataMatrix^\unobservedSet_*$. 
Specifically, section \ref{app:partialtestBound} explains in more detail
the form of the variational lower bound for the aforementioned prediction scenario
and illustrates how this gives rise to certain computational differences for the standard and the dynamical GP-LVM.
Section \ref{app:partialtestPosterior} gives some more details for the mathematical formulae
associated with the above prediction task.

\subsection{The Variational Bound in the Test Phase and Computational Issues\label{app:partialtestBound}}
As discussed in Section \ref{sec:predStandardGPLVM}, when doing predictions based
on partially observed outputs with the variational GP-LVM, one needs to construct
a variational lower bound as for the training phase. However, this now needs to be associated
with the full set of observations $(\dataMatrix, \dataMatrix^\unobservedSet_*)$.
Specifically, we need to lower bound the marginal likelihood given in equation \eqref{eq:marginalPredictions2}
with a variational bound that takes the form:
\begin{equation}
\label{eq:predictiveMissing1a}
\log p(\dataMatrix_*^\observedSet, \dataMatrix) \geq \int q(\latentMatrix_*, \latentMatrix) \log \frac{ p(\dataMatrix^\unobservedSet | \latentMatrix) 
    p(\dataMatrix_*^\observedSet, \dataMatrix^\observedSet|\latentMatrix_*, \latentMatrix) p(\latentMatrix_*,\latentMatrix)}{ q(\latentMatrix_*, \latentMatrix)} 
    \intd  \latentMatrix_* \intd \latentMatrix .
\end{equation}
For the standard variational GP-LVM, we can further expand the above equation by noticing that
the distributions $q(\latentMatrix,\latentMatrix_*)$ and $p(\latentMatrix,\latentMatrix_*)$ are fully factorised
as $q(\latentMatrix, \latentMatrix_*) = 
      \prod_{\dataIndex=1}^\numData q(\latentVector_{\dataIndex, :})
      \prod_{\dataIndex=1}^{\numData_*} q(\latentVector_{\dataIndex, *})$.
Therefore, equation \eqref{eq:predictiveMissing1a} can be written as:
\begin{align}
\log p(\dataMatrix_*^\observedSet, \dataMatrix) & \geq  \int q(\latentMatrix) \log p(\dataMatrix^\unobservedSet | \latentMatrix) \intd  \latentMatrix 
    +  \int q(\latentMatrix_*,\latentMatrix) \log p(\dataMatrix_*^\observedSet, \dataMatrix^\observedSet|\latentMatrix_*, \latentMatrix)
    \intd  \latentMatrix_* \intd  \latentMatrix  \nonumber \\
& - \KL{q(\latentMatrix)}{p(\latentMatrix)} - \KL{q(\latentMatrix_*)}{p(\latentMatrix_*)}. \label{eq:predictiveMissing1b}
\end{align}  
Recalling equation \eqref{approxF1}, we see the first term above 
can be obtained as the sum $\sum_{j \in \unobservedSet} \hat{\F}_j \left( q(\latentMatrix) \right)$
where each of the involved terms is given by equation \eqref{FdFinal} and is already computed during the training
phase and, therefore, can be held fixed during test time. Similarly, 
the third term of equation \eqref{eq:predictiveMissing1b} is also held fixed during test time.
As for the second and fourth term, they can be optimised exactly
as the bound computed for the training phase with the difference that
now the data are augmented with test observations and only the observed
dimensions are accounted for.

In contrast, the dynamical version of our model requires the full set of latent variables
($\latentMatrix, \latentMatrix_*$) to be fully coupled in the variational distribution
$q(\latentMatrix, \latentMatrix_*)$, as they together form a timeseries. 
Consequently, the expansion of equation \eqref{eq:predictiveMissing1b} cannot be applied here,
meaning that in this case no precomputations can be used from the training phase.
However, one could apply the approximation $q(\latentMatrix, \latentMatrix_*) = q(\latentMatrix)q(\latentMatrix_*)$
to speed up the test phase. In this case, each set of latent variables is still correlated, but
the two sets are not. However, this approximation was not used in our implementation
as it is only expected to speed up the predictions phase if the training set is very big,
which is not the case for our experiments.

\subsection{Calculation of the Posterior $q(\mappingFunctionMatrix_*^\unobservedSet|\latentMatrix)$ \label{app:partialtestPosterior}}
Optimisation based on the variational bound constructed for the test phase with partially observed outputs,
as explained in Section \ref{sec:predStandardGPLVM}, gives rise to the posterior 
$q(\mappingFunctionMatrix_*^\unobservedSet, \inducingMatrix, \latentMatrix_*)$, as exactly happens in the
training phase. Therefore, according to equation \eqref{varDistr} we can write:
\begin{equation}
q(\mappingFunctionMatrix_*^\unobservedSet, \inducingMatrix,\latentMatrix_*) = 
 \left( \prod_{j=1}^\dataDim p(\mappingFunctionVector_{*, \outputIndex}^\unobservedSet | \inducingVector_{:, \outputIndex}, \latentMatrix_* )q(\inducingVector_{:, \outputIndex}) \right) q(\latentMatrix_*) .
\end{equation}
The marginal $q(\mappingFunctionMatrix_*^\unobservedSet|\latentMatrix_*)$ 
(of equation \eqref{eq:testPosteriorF})
is then simply found as
$\prod_{j \in \unobservedSet} \int p(\mappingFunctionVector_{*, \outputIndex}^\unobservedSet | \inducingVector_{:, \outputIndex}, \latentMatrix_* )q(\inducingVector_{:, \outputIndex}) \intd \inducingVector_{:,\outputIndex}$.
The integrals inside the product are easy to compute since both types of densities appearing there are Gaussian,
according to equations \eqref{priorF2} and \eqref{eq:qu}. In fact, each factor takes the form of a
projected process predictive distribution from sparse GPs \citep{Csato:sparse02,Seeger:fast03,Rasmussen:book06}.

We will show the analytic derivation for the general case where we do not distinguish between training or test variables and all dimensions are observed. In specific, we want to compute:
\begin{align*}
p(\mappingFunctionVector_{:,\outputIndex}|\latentMatrix) 
 &=  \int p(\mappingFunctionVector_{:,\outputIndex} | \inducingVector_{:,\outputIndex}, \latentMatrix) 
          q(\inducingVector_{:,\outputIndex}) \intd \inducingVector_{:,\outputIndex}  .
\end{align*}
For this calculation we simply use the following identity for Gaussians:
\begin{align*}
 \int \gaussianDist{\mappingFunctionVector_{:,\outputIndex}}
                      {\bfM \inducingVector_{:,\outputIndex} + \bfm}{\bfSigma_f} 
         \gaussianDist{\inducingVector_{:,\outputIndex}}
                      {\bfmu_u}{\bfSigma_u}  \intd \inducingVector_{:, \outputIndex}
 &=   \gaussianDist{\mappingFunctionVector_{:,\outputIndex}}
                   {\bfM \bfmu_u + \bfm}{\bfSigma_f + \bfM \bfSigma_u \bfM^\T}.
\end{align*}
From equations \eqref{eq:conditionalGPmeanCovar} and \eqref{eq:qu} we recognise:
\newline

\begin{tabular}{ l l }
 $\bfM = \Kfu \Kuu^{-1}$ , $\bfm = \bfzero$ & $\bfmu_u = \Kuu (\dataStd^2 \Kuu + \bfPsi_2)^{-1} \bfPsi_1^\T \dataVector_{:,\outputIndex}$ \\
   $\bfSigma_f = \Kfu - \Kfu \Kuu^{-1} \Kuf$     & $\bfSigma_u = \dataStd^2 \Kuu (\dataStd^2 \Kuu + \bfPsi_2)^{-1} \Kuu$
\end{tabular}
\newline

\noindent from where we easily find:
\begin{equation*}
p(\mappingFunctionVector_{:,\outputIndex} | \latentMatrix) = \gaussianDist{\mappingFunctionVector_{:,\outputIndex}}
              {\Kfu \bfB}
              {\Kff - \Kfu \left( \Kuu^{-1} + \left( \Kuu + \dataStd^{-2} \bfPsi_2 \right)^{-1} \Kuf \right)}
\end{equation*}
with $\bfB = \dataStd^{-2}(\Kuu+\dataStd^{-2}\bfPsi_2)^{-1} \bfPsi_1^\T \dataVector_{:,\outputIndex}$.

\section{Algorithm for Semi-supervised Gaussian Processes \label{app:semiSupervised}}
Consider a fully and a partially observed set of inputs, \ie $\bfZ = (\bfZ^\observedSet, \bfZ^\unobservedSet)$, where $\observedSet$ and $\unobservedSet$ denote set of rows of $(\bfZ, \dataMatrix)$ that contain fully and partially observed inputs respectively.
The features missing in $\bfZ^\unobservedSet$ can be different in number / location for each individual point $\bfz^\unobservedSet_{\dataIndex,:}$.
We can train the model in all of these observations jointly, by replacing the inputs $\bfZ^\observedSet$ and $\bfZ^\unobservedSet$ with distributions $q(\latentMatrix^\observedSet)$ and $q(\latentMatrix^\unobservedSet)$ respectively, and using Algorithm \ref{algorithm:semiSupervised}. Since the posterior distribution is factorised, the algorithm constrains it to be close to a delta function in regions where we have observations, \ie in areas corresponding to $\bfZ^\observedSet$ and in areas corresponding to non-missing locations of $\bfZ^\unobservedSet$. The rest of the posterior area's parameters (means and variances of Gaussian marginals) are initialised according to a prediction model $\mathcal{M}^\observedSet$ and are subsequently optimised (along with model parameters) in an augmented model $\mathcal{M}^{\observedSet,\unobservedSet}$. Notice that the initial model $\mathcal{M}^\observedSet$ is obtained by training a variational GP-LVM model with a posterior $q(\latentMatrix^\observedSet)$ whose mean is fully constrained to match the observations $\bfZ^\observedSet$ with very small uncertainty and, thus, the model $\mathcal{M}^\observedSet$ behaves almost as a standard GP regression model.
\begin{algorithm}
\caption{Semi-supervised Gaussian Processes: Training and predictions}\label{algorithm:semiSupervised}
\begin{algorithmic}[1]
\SSTATE \emph{Given}: fully observed data $(\bfZ^\observedSet, \dataMatrix^\observedSet)$ and partially observed data $(\bfZ^\unobservedSet, \dataMatrix^\unobservedSet)$
\SSTATE Define a small value, \eg $\varepsilon = 10^{-9}$
\SSTATE Initialize $q(\latentMatrix^\observedSet) = \prod_{\dataIndex=1}^\numData
      \gaussianDist{\latentVector^\observedSet_{\dataIndex,:}}
                    {\bfz^\observedSet_{\dataIndex,:}}
                    {\varepsilon \eye}$
\SSTATE Fix $q(\latentMatrix^\observedSet)$ in the optimiser \COMMENT{\emph{(\ie will not be optimised)}}
\SSTATE Train a variational GP-LVM model $\mathcal{M}^\observedSet$ given the above $q(\latentMatrix^\observedSet)$ and $\dataMatrix^\observedSet$

\FOR{$\dataIndex = 1, \cdots, |\dataMatrix^\unobservedSet|$}
    \STATE Predict $p(\hat{\latentVector}_{\dataIndex, :}^\unobservedSet |  \dataVector^\unobservedSet_{\dataIndex}, \mathcal{M}^\observedSet) 
    \approx q(\hat{\latentVector}_{\dataIndex, :}^\unobservedSet) = 
    \gaussianDist{\hat{\latentVector}_{\dataIndex, :}^\unobservedSet}{\hat{\bfmu}_{\dataIndex,:}^\unobservedSet}{\hat{\bfS}_{\dataIndex}^\unobservedSet}$
    \STATE Initialize $q(\latentVector_{\dataIndex, :}^\unobservedSet) = 
    \gaussianDist{\latentVector_{\dataIndex, :}^\unobservedSet}
                 {\bfmu^\unobservedSet_{\dataIndex,:}}
                 {\bfS^\unobservedSet_\dataIndex}$ as follows:
      \FOR{$\latentIndex = 1, \cdots, \latentDim$}
      \IF{$z^\unobservedSet_{\dataIndex, \latentIndex}$ is observed}
          \STATE $\mu^\unobservedSet_{\dataIndex, \latentIndex}=z^\unobservedSet_{\dataIndex, \latentIndex}$
            and  $(S^\unobservedSet_\dataIndex)_{\latentIndex, \latentIndex} = \varepsilon$
            \COMMENT{\emph{$(S^\unobservedSet_\dataIndex)_{\latentIndex, \latentIndex}$ denotes the $\latentIndex$-th diagonal element of $\bfS_\dataIndex^\unobservedSet$}}
          \STATE Fix $\mu^\unobservedSet_{\dataIndex, \latentIndex}, (S^\unobservedSet_\dataIndex)_{\latentIndex, \latentIndex}$ in the optimiser
           \COMMENT{\emph{(\ie will not be optimised)}}
      \ELSE
            \STATE $\mu^\unobservedSet_{\dataIndex, \latentIndex} = \hat{\mu}^\unobservedSet_{\dataIndex, \latentIndex}$
              and  $(S^\unobservedSet_\dataIndex)_{\latentIndex, \latentIndex} = (\hat{S}^\unobservedSet_\dataIndex)_{\latentIndex, \latentIndex}$
      \ENDIF
      \ENDFOR
\ENDFOR

\SSTATE Train a variational GP-LVM model $\mathcal{M}^{\observedSet, \unobservedSet}$ using the initial
      $q(\latentMatrix^\observedSet)$ and $q(\latentMatrix^\unobservedSet)$ defined above and
      data $\dataMatrix^\observedSet, \dataMatrix^\unobservedSet$ (the locations that were fixed for the variational distributions will not be optimised). 
\SSTATE All subsequent predictions can be made using model $\mathcal{M}^{\observedSet, \unobservedSet}$.
\end{algorithmic}
\end{algorithm}

\section{Additional Results from the Experiments}
\begin{figure}[ht]
\begin{center}
\subfigure[]{
	\includegraphics[width=0.4\textwidth]{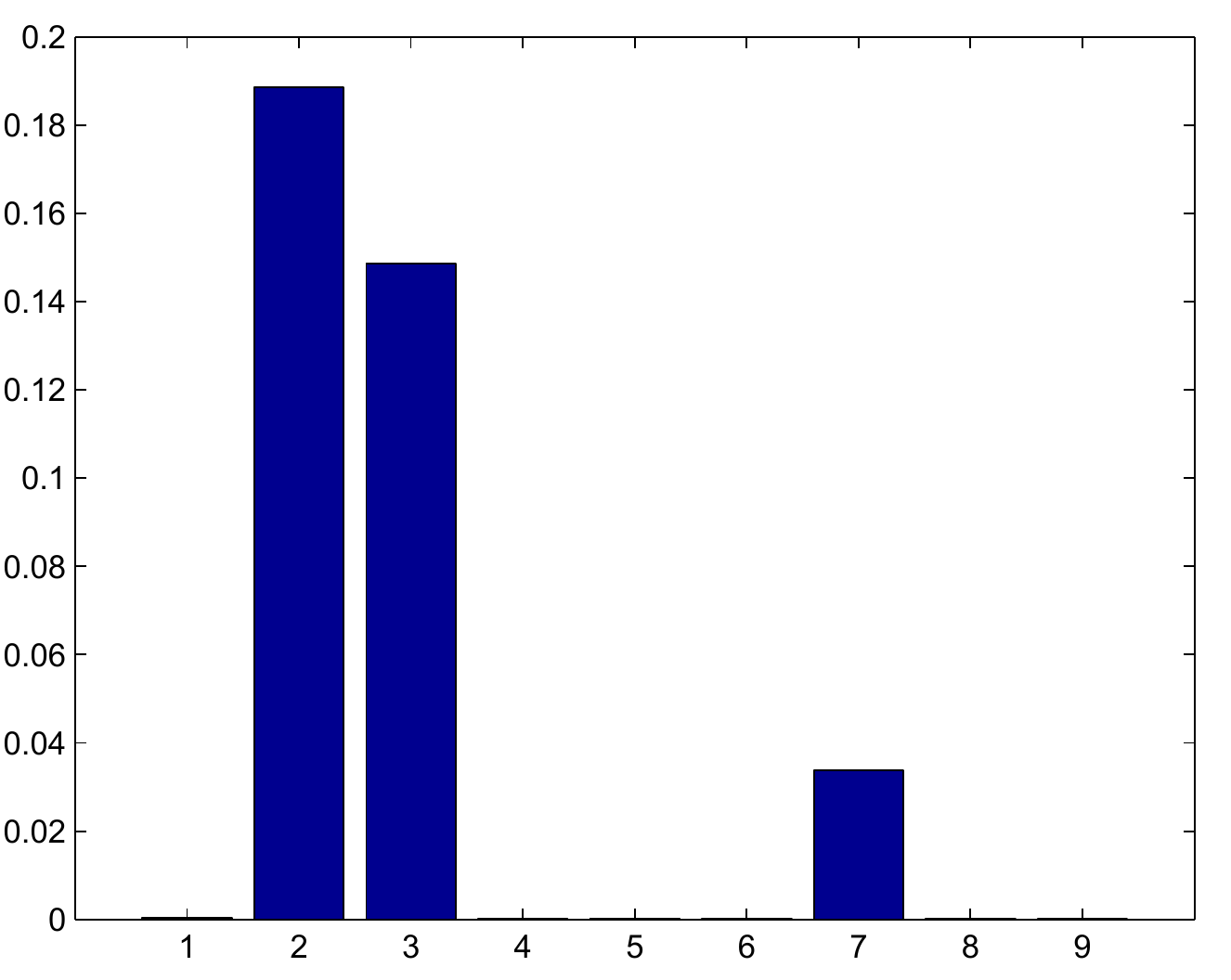}
	\label{fig:suppMocap1}
}
\subfigure[]{
	\includegraphics[width=0.4\textwidth]{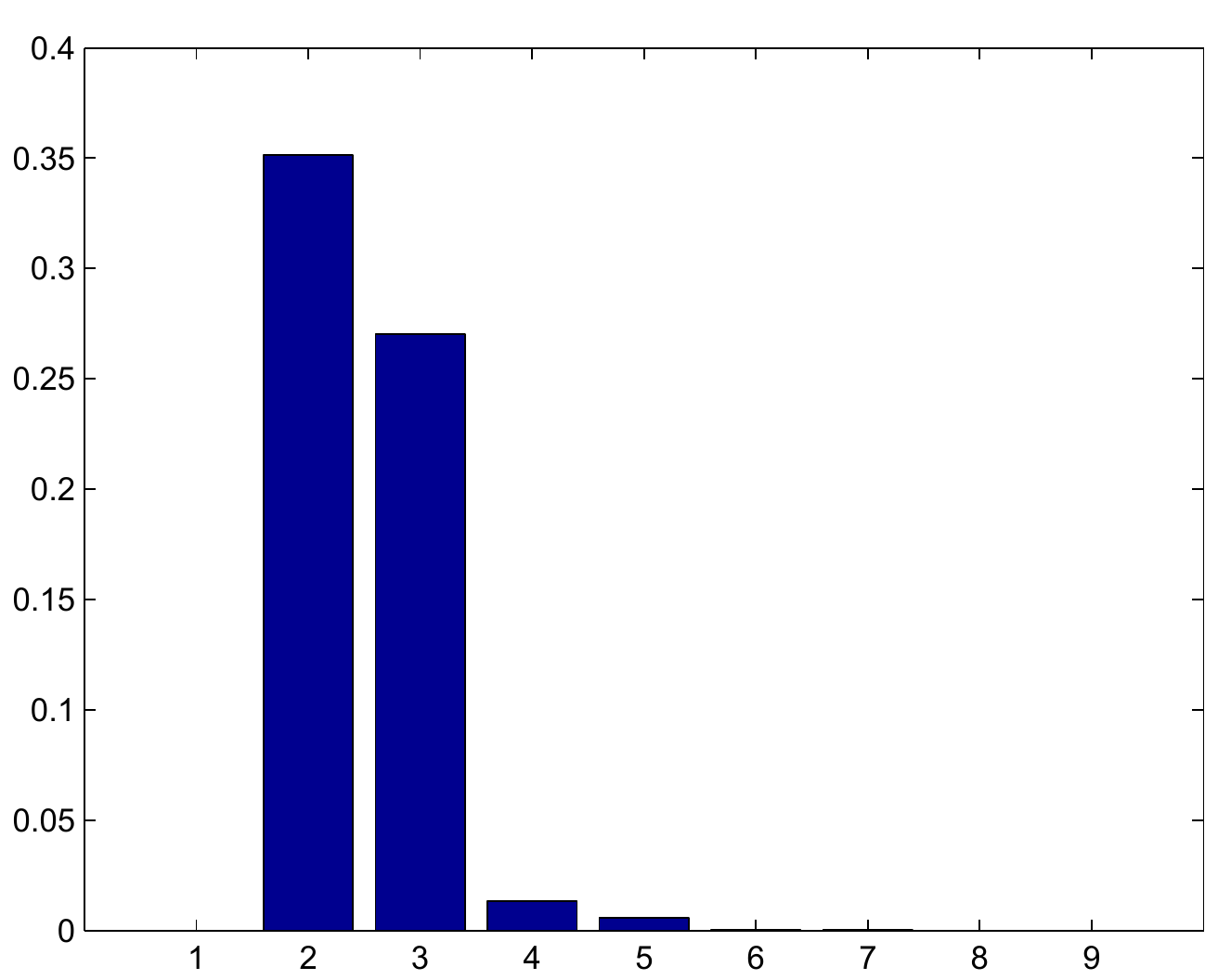}
	\label{fig:suppMocap2}
}
\end{center}
\caption{\small{
The values of the scales of the ARD kernel after training on the motion capture dataset using the exponentiated quadratic (fig: \subref{fig:suppMocap1}) and the Mat\'ern (fig: \subref{fig:suppMocap2}) kernel to model the dynamics for the dynamical variational GP-LVM. The scales that have zero value ``switch off'' the corresponding dimension of the latent space. The latent space is, therefore, 3-D for \subref{fig:suppMocap1} and 4-D for \subref{fig:suppMocap2}. Note that the scales were initialized with very similar values (\eg a vector of ones with added random noise).
}
}
\label{fig:supplMocap1}
\end{figure}

\begin{figure}[ht]
\begin{center}
\subfigure[]{
	\includegraphics[width=0.48\textwidth]{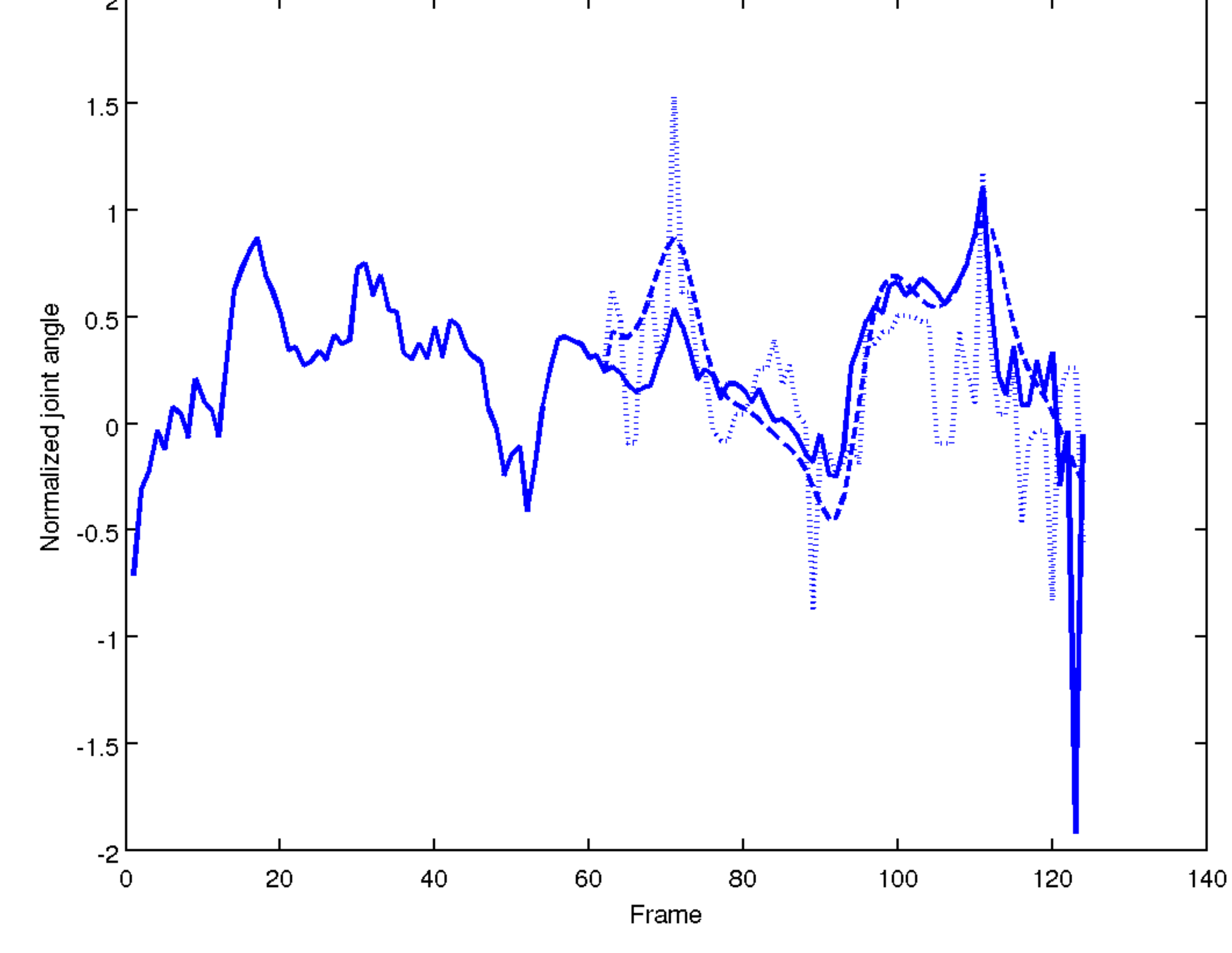}
	\label{fig:suppMocap3}
}
\subfigure[]{
	\includegraphics[width=0.48\textwidth]{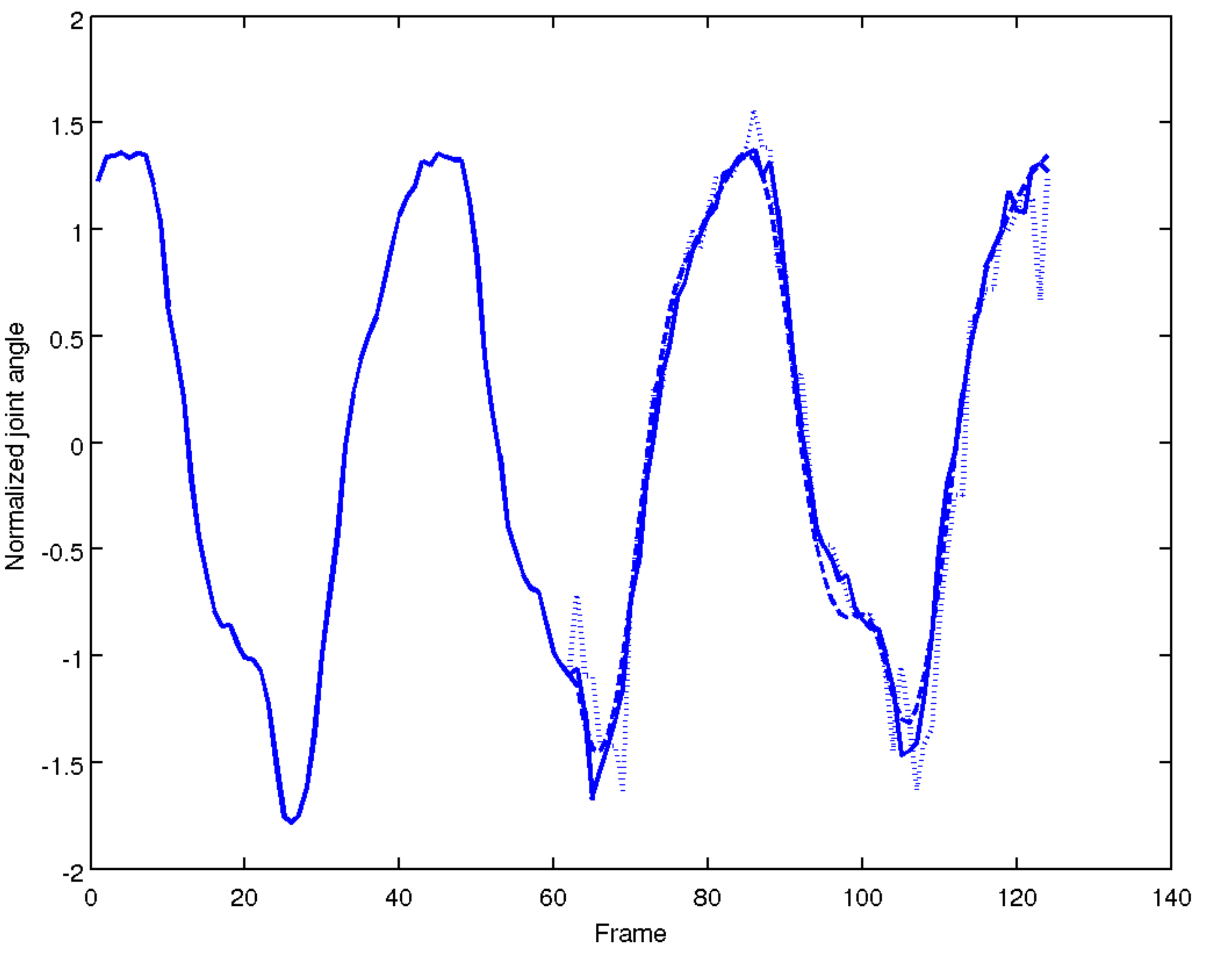}
	\label{fig:suppMocap4}
}
\end{center}
\caption{\small{
The prediction for two of the test angles for the body (fig: \ref{fig:suppMocap3}) and for the legs part (fig: \ref{fig:suppMocap4}). Continuous line is the original test data, dotted line is nearest neighbour in scaled space, dashed line is dynamical variational GP-LVM (using the exponentiated quadratic kernel for the body reconstruction and the Mat\'ern for the legs).
}
}
\label{fig:supplMocap2}
\end{figure}

In this section we present additional figures obtained from the experiment
described in Section \ref{sec:experimentsMocap} using motion capture data.
Figure \ref{fig:supplMocap1} depicts the optimised ARD weights for each of the
dynamical models employed in the experiment. Figure \ref{fig:supplMocap2} illustrates
examples of the predictive performance of the models by plotting the true and 
predicted curves in the angle space.

\begin{figure}[ht]
\begin{center}
\subfigure[]{
  \includegraphics[width=0.23\textwidth]{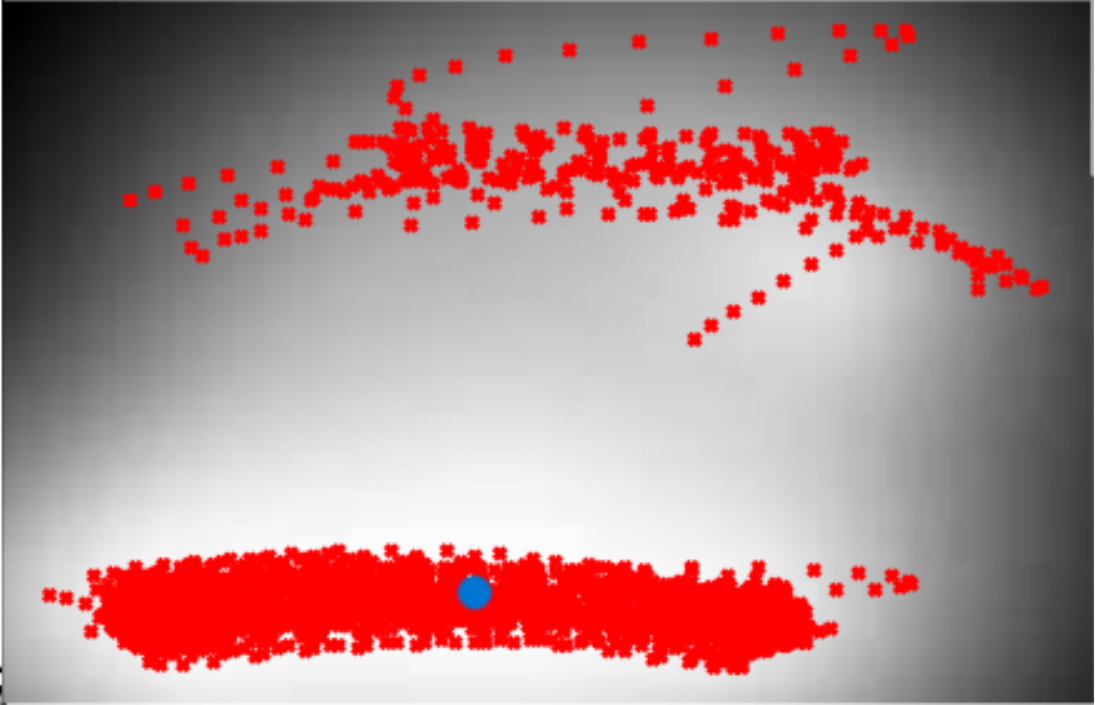}
  \label{fig:walkRegime}
}
\subfigure[]{
  \includegraphics[width=0.23\textwidth]{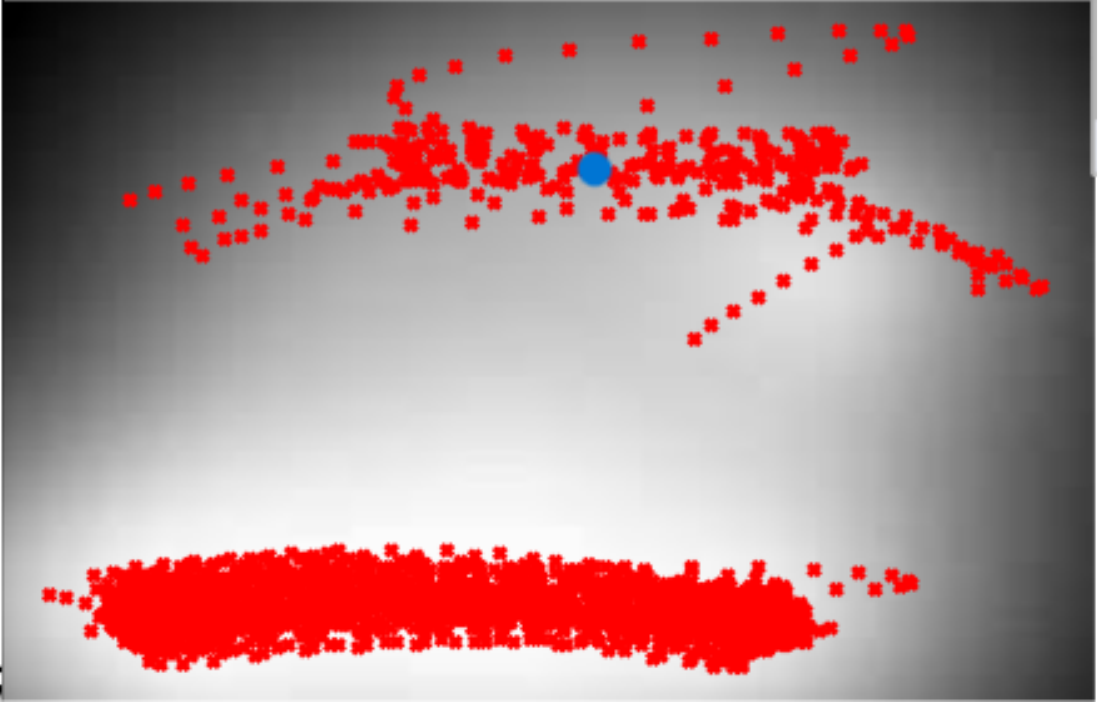}
  \label{fig:runRegime}
}
\subfigure[]{
  \includegraphics[width=0.23\textwidth]{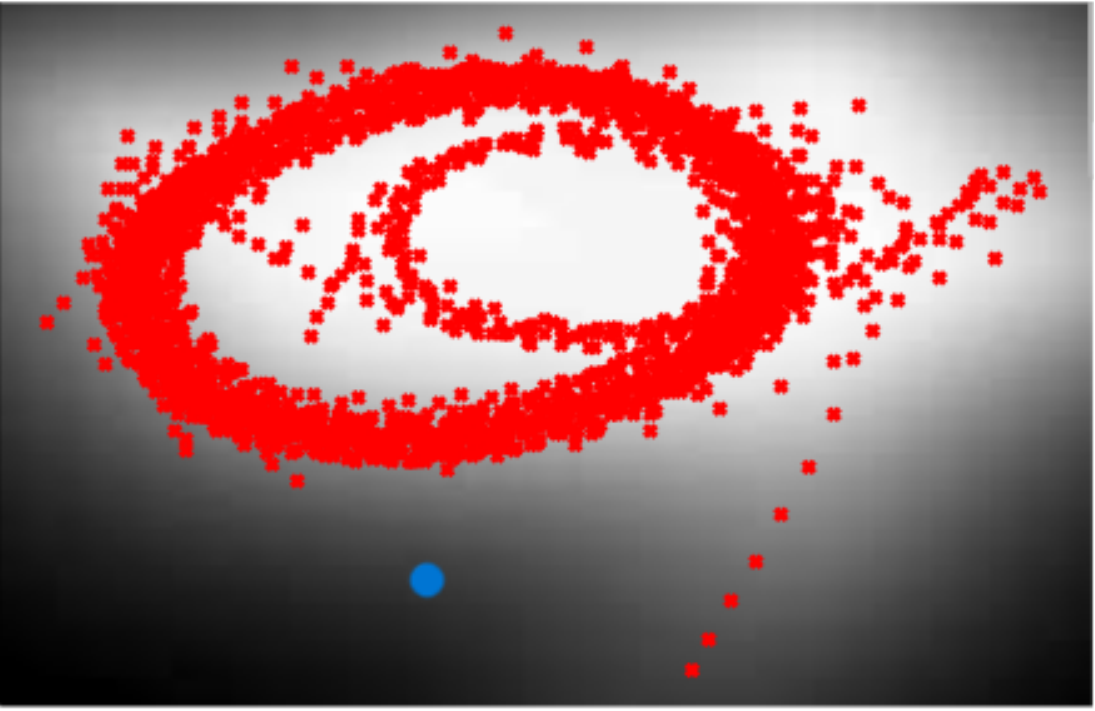}
  \label{fig:unnatural_lat}
}
\subfigure[]{
  \includegraphics[width=0.10\textwidth]{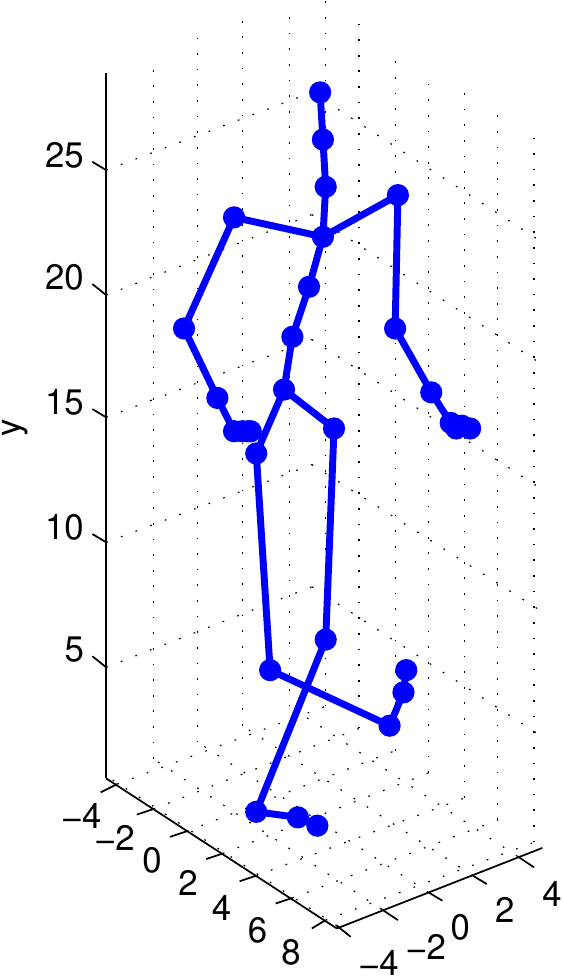}
  \label{fig:unnatural_mocap_run}
}
\end{center}
\caption{\small{Plots \subref{fig:walkRegime} and \subref{fig:runRegime} depict the projection of
the latent space on dimensions $2$ and $4$, with the blue dot corresponding to the value on which
these dimensions were fixed for the sampled latent points and red crosses represent latent points corresponding
to training outputs. The intensity of the grayscale background represents the posterior uncertainty at each region
(white corresponds to low predictive variance). 
Plot \subref{fig:unnatural_lat} depicts
a latent space projection on dimensions $2$ and $3$, with the fixed latent positions corresponding to the
generated output depicted in plot \subref{fig:unnatural_mocap_run}.}
}
\label{fig:supplMocapSamples2}
\end{figure}

\begin{figure}[ht]
\begin{center}
\subfigure{
  \includegraphics[width=0.23\textwidth]{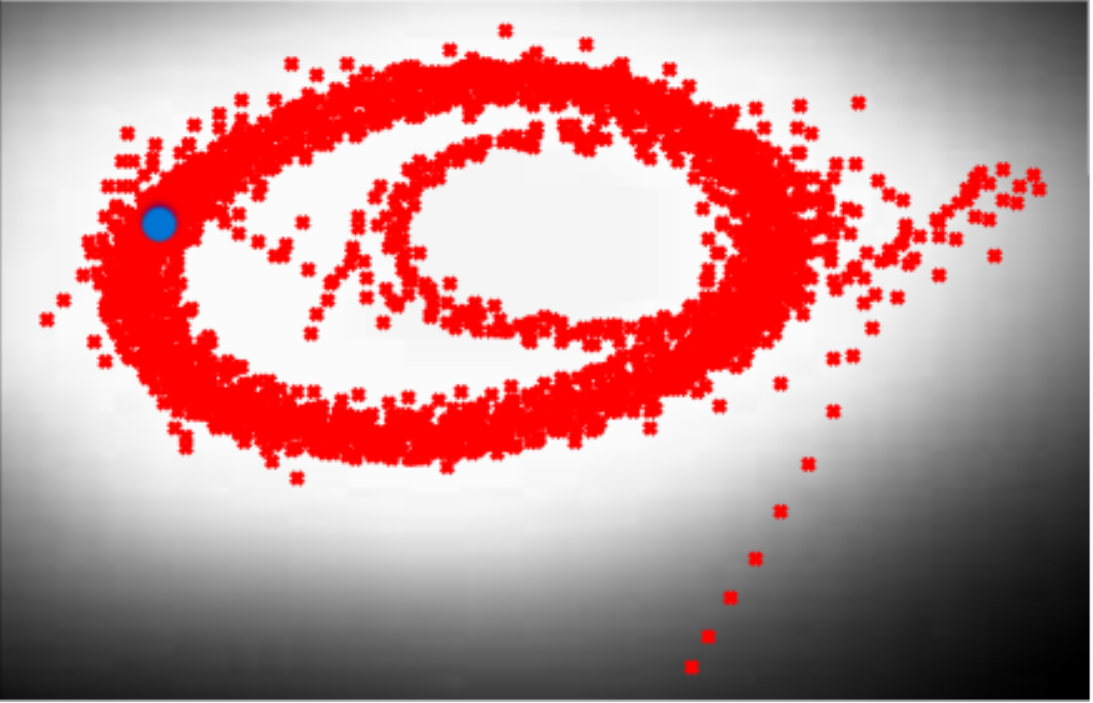}
  \label{fig:1lat}
}
\subfigure{
  \includegraphics[width=0.23\textwidth]{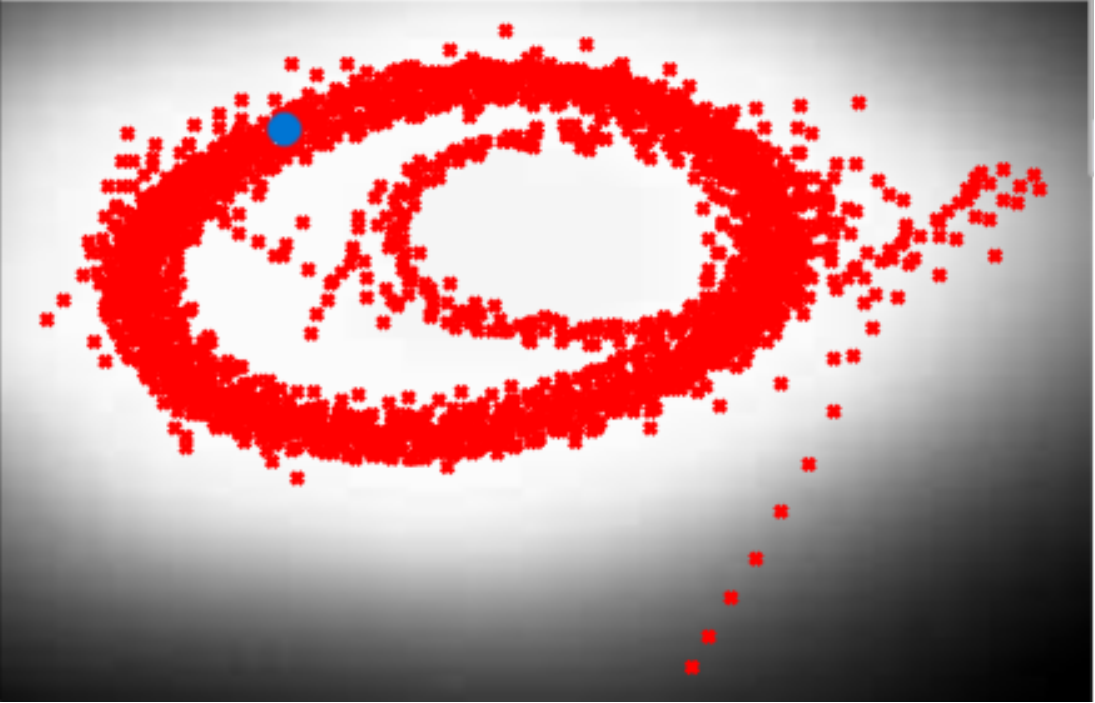}
  \label{fig:2lat}
}
\subfigure{
  \includegraphics[width=0.23\textwidth]{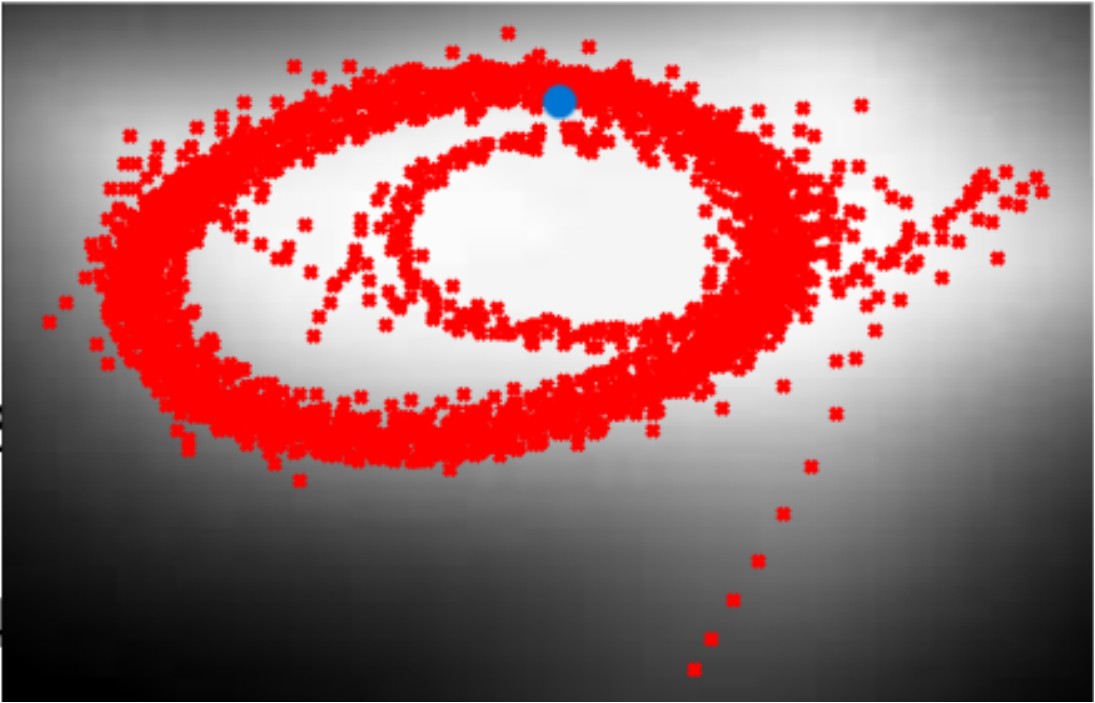}
  \label{fig:3lat}
}
\subfigure{
  \includegraphics[width=0.23\textwidth]{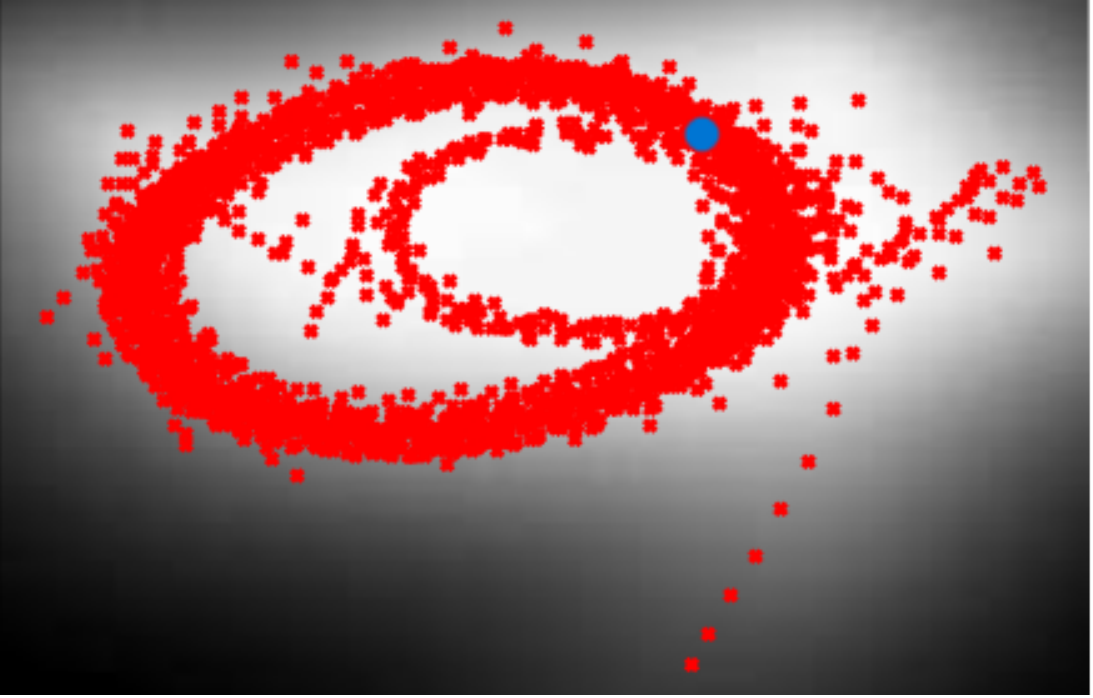}
  \label{fig:4lat}
}
\subfigure{
  \includegraphics[width=0.12\textwidth]{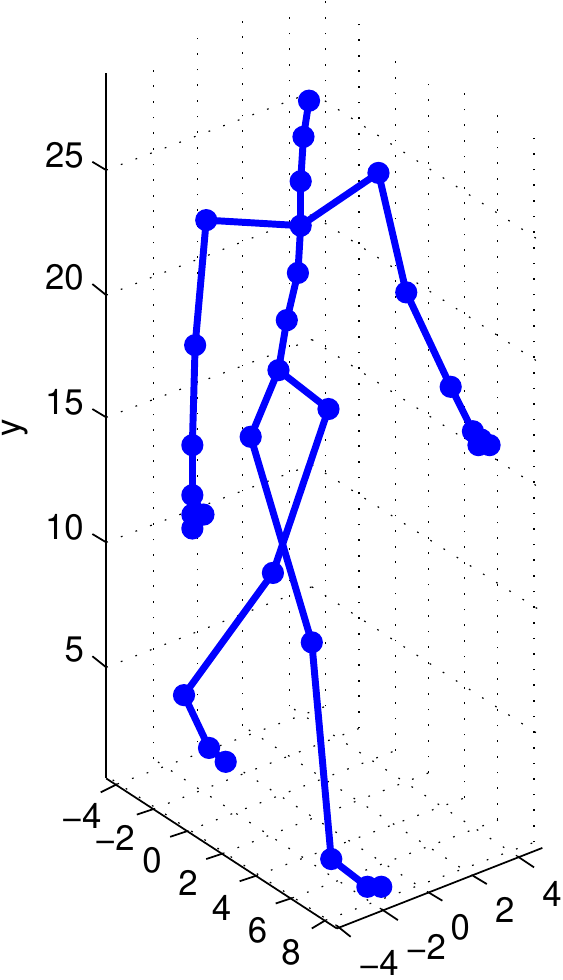}
  \label{fig:1mocap_walk}
}
\hfill
\subfigure{
  \includegraphics[width=0.12\textwidth]{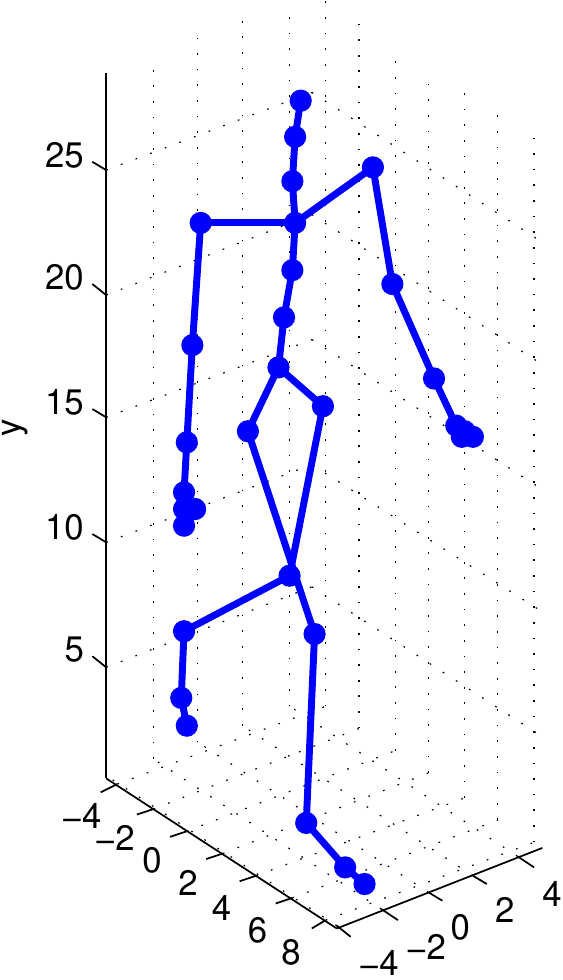}
  \label{fig:2mocap_walk}
}
\hfill
\subfigure{
  \includegraphics[width=0.12\textwidth]{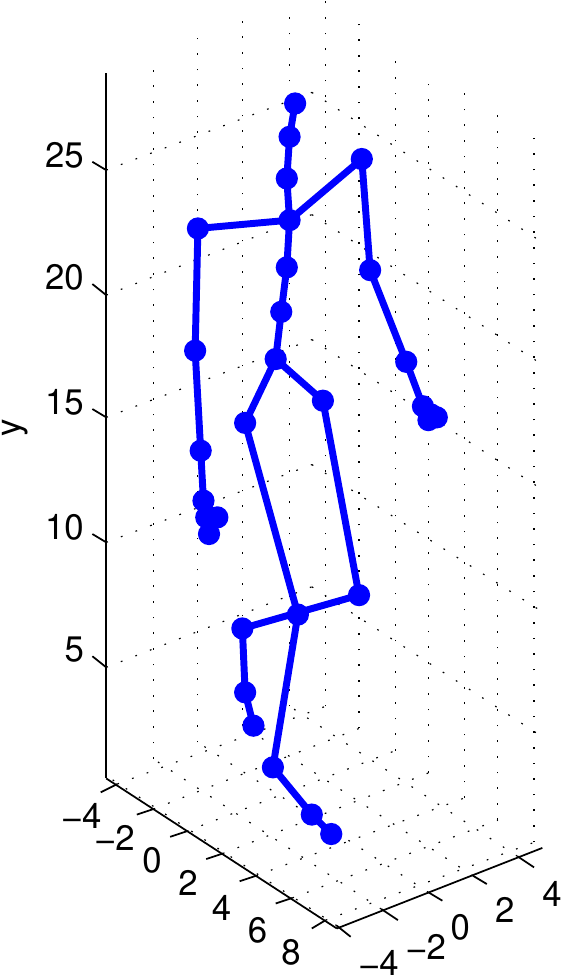}
  \label{fig:3mocap_walk}
}
\hfill
\subfigure{
  \includegraphics[width=0.12\textwidth]{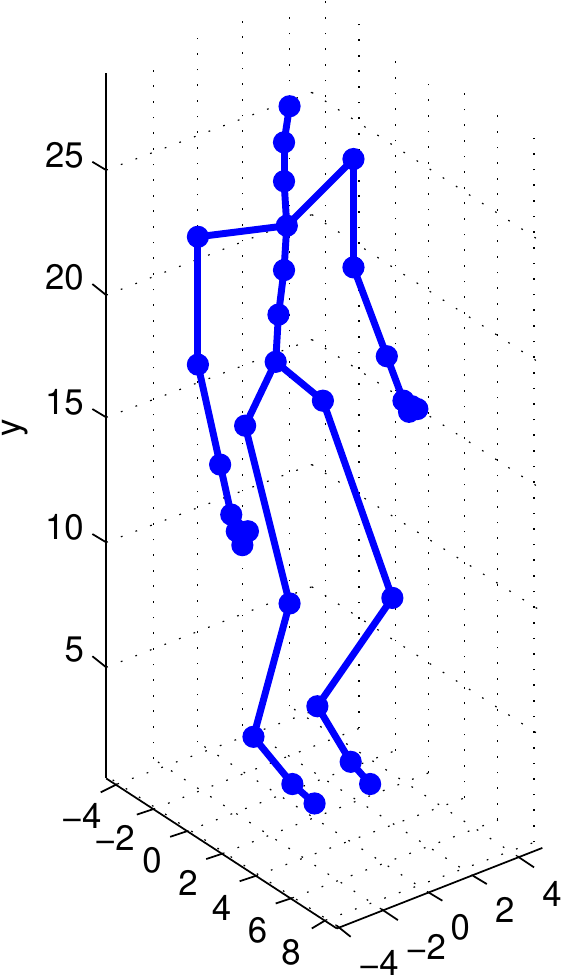}
  \label{fig:4mocap_walk}
}
\\
\subfigure{
  \includegraphics[width=0.12\textwidth]{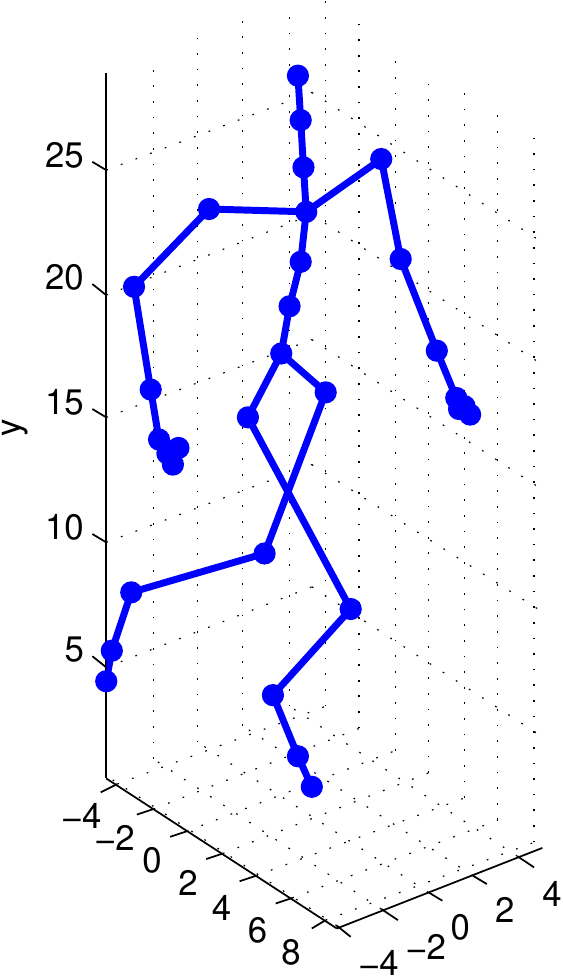}
  \label{fig:1mocap_run}
}
\hfill
\subfigure{
  \includegraphics[width=0.12\textwidth]{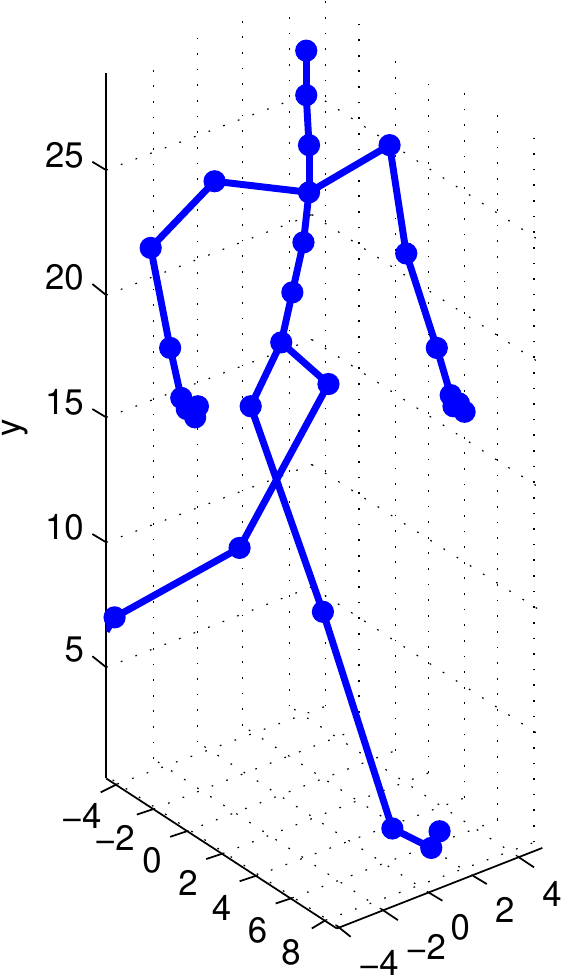}
  \label{fig:2mocap_run}
}
\hfill
\subfigure{
  \includegraphics[width=0.12\textwidth]{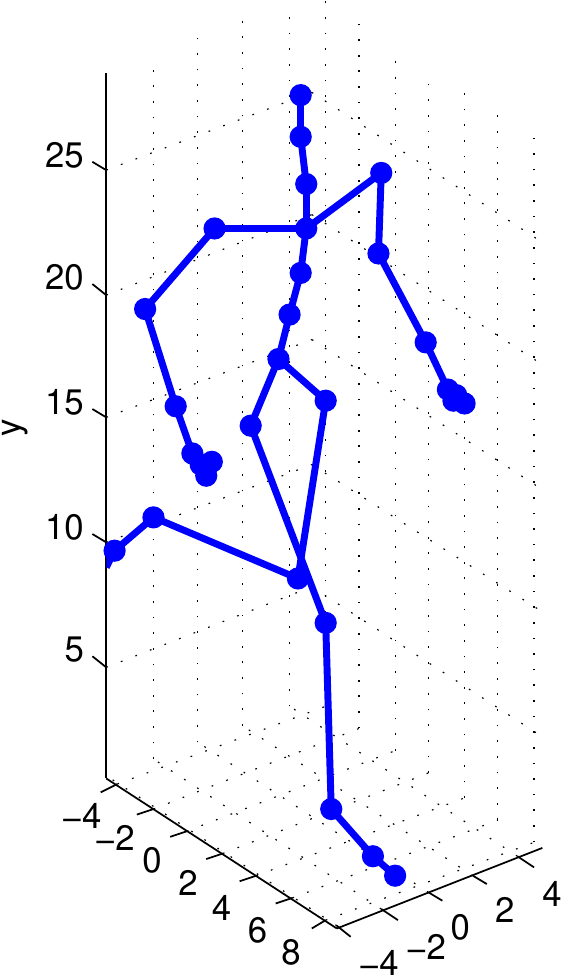}
  \label{fig:3mocap_run}
}
\hfill
\subfigure{
  \includegraphics[width=0.12\textwidth]{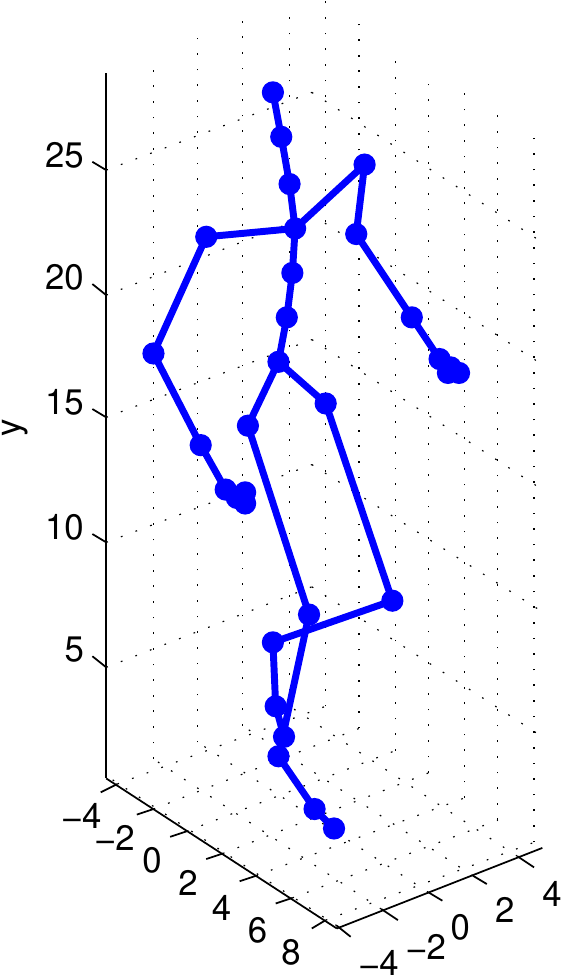}
  \label{fig:4mocap_run}
}
\end{center}
\caption{\small{The first row depicts a projection of the latent space on dimensions $2$ and $3$ with the
blue dot showing the value at which these dimensions were fixed for the sampled latent points. 
The corresponding outputs are depicted in the second row (for the walk regime) and third row (for the run regime).}
}
\label{fig:supplMocapSamples}
\end{figure}

As was explained in Section \ref{sec:experimentsMocap}, all employed models
encode the ``walk'' and ``run'' regime as two separate subspaces in the latent space.
To illustrate this more clearly we sampled points from the learned latent space $\latentMatrix$ of a 
trained dynamical variational GP-LVM model and generated the corresponding outputs, so as to
investigate the kind of information that is encoded in each subspace of $\latentMatrix$.
Specifically, we considered the model that employed a Mat\'ern $\frac{3}{2}$ covariance function to constrain the latent space and, based on the ARD weights of Figure \ref{fig:suppMocap2}, we projected the latent space on 
dimensions $(2,3)$ and $(2,4)$. Interacting with the model revealed that dimension $4$ separates the
``walk'' from the ``run'' regime. In particular, we first fixed dimension $4$ on a value belonging to the region encoding the walk, as can be seen in Figure \ref{fig:walkRegime}, and then sampled multiple latent points by varying the other two dominant dimensions, namely $2$ and $3$, as can be seen in the top row of Figure \ref{fig:supplMocapSamples}. The corresponding outputs are shown in the second row of Figure \ref{fig:supplMocapSamples}. When dimension $4$ was fixed on a value belonging to the region encoding the run (Figure \ref{fig:runRegime}) the outputs obtained by varying dimensions $2$ and $3$ as before produced a smooth running motion, as can be seen in the third row of Figure \ref{fig:supplMocapSamples}. 
Finally, Figure \ref{fig:unnatural_mocap_run} illustrates a motion which clearly is very different from the training set and was obtained by sampling a latent position far from the training data, as can be seen in Figure \ref{fig:unnatural_lat}. This is indicative of a generative model's ability of producing novel data.

